\documentclass[12pt,a4paper]{article}

\usepackage[
  a4paper,
  left=2.5cm,
  right=2.5cm,
  top=2.3cm,
  bottom=2.3cm,
  headheight=12pt,
  headsep=6pt,
  footskip=1.2cm
]{geometry}

\usepackage{amsmath,amssymb,amsthm}
\usepackage{graphicx,psfrag,epsf}
\usepackage{enumerate}
\usepackage[authoryear,round,sort&compress]{natbib}

\usepackage{booktabs}
\usepackage{cleveref}
\usepackage{algorithm}
\usepackage{algpseudocode}
\usepackage{enumitem}
\usepackage{url}
\usepackage{fancyhdr}
\usepackage{bm}

\newtheorem{theorem}{Theorem}
\newtheorem{lemma}{Lemma}

\theoremstyle{definition}

\theoremstyle{remark}

\fancypagestyle{plain}{
  \fancyhf{}
  \fancyfoot[C]{\footnotesize \thepage}

}

\usepackage{graphicx}

\newcommand{\panelimage}[1]{%
  \includegraphics[
    width=1.35cm,
    height=1.35cm,
    keepaspectratio
  ]{fig/#1.png}%
}

\newcommand{\shapepanel}[1]{%
  \begingroup

  \setlength{\tabcolsep}{1.5pt}
  \renewcommand{\arraystretch}{0.95}

  \begin{tabular}{@{}ccccc@{\hspace{3pt}}c@{}}
    \scriptsize True
    &
    \scriptsize Normal
    &
    \scriptsize $t_3$
    &
    \scriptsize LogN
    &
    \scriptsize Cont.
    &
    {}
    \\[2pt]
    
    \panelimage{r5#1}
    &
    \panelimage{ex1shape#1dim5n50noise0errnormdselectRERcov_typeenvelopelambda0}
    &
    \panelimage{ex1shape#1dim5n50noise0errt3dselectRERcov_typeenvelopelambda0}
    &
    \panelimage{ex1shape#1dim5n50noise0errlnormdselectRERcov_typeenvelopelambda0}
    &
    \panelimage{ex1shape#1dim5n50noise0errcontaminateddselectRERcov_typeenvelopelambda0}
    &
    \scriptsize $n=50$
    \\

    &
    \panelimage{ex1shape#1dim5n100noise0errnormdselectRERcov_typeenvelopelambda0}
    &
    \panelimage{ex1shape#1dim5n100noise0errt3dselectRERcov_typeenvelopelambda0}
    &
    \panelimage{ex1shape#1dim5n100noise0errlnormdselectRERcov_typeenvelopelambda0}
    &
    \panelimage{ex1shape#1dim5n100noise0errcontaminateddselectRERcov_typeenvelopelambda0}
    &
    \scriptsize $n=100$
    \\

    &
    \panelimage{ex1shape#1dim5n200noise0errnormdselectRERcov_typeenvelopelambda0}
    &
    \panelimage{ex1shape#1dim5n200noise0errt3dselectRERcov_typeenvelopelambda0}
    &
    \panelimage{ex1shape#1dim5n200noise0errlnormdselectRERcov_typeenvelopelambda0}
    &
    \panelimage{ex1shape#1dim5n200noise0errcontaminateddselectRERcov_typeenvelopelambda0}
    &
    \scriptsize $n=200$
    \\

    &
    \panelimage{ex1shape#1dim5n400noise0errnormdselectRERcov_typeenvelopelambda0}
    &
    \panelimage{ex1shape#1dim5n400noise0errt3dselectRERcov_typeenvelopelambda0}
    &
    \panelimage{ex1shape#1dim5n400noise0errlnormdselectRERcov_typeenvelopelambda0}
    &
    \panelimage{ex1shape#1dim5n400noise0errcontaminateddselectRERcov_typeenvelopelambda0}
    &
    \scriptsize $n=400$
    \\

  \end{tabular}
  \endgroup
}

\newcommand{\shapebox}[1]{%
  \begin{minipage}[t]{0.480\textwidth}
    \centering

    \resizebox{\linewidth}{!}{%
      \shapepanel{#1}%
    }

  \end{minipage}%
}

\begin{document}
\def\spacingset#1{\renewcommand{\baselinestretch}%
{#1}\small\normalsize} \spacingset{1}

\title{  Partial Modal Tensor  Regression }
\author{ 
   Jintao Han ,  Li Zhang  \\
  Soochow University, School of Mathematical Sciences, \\
  %Suzhou, 215006, Jiangsu, China
  \and
  Xuejun Ma \thanks{Corresponding author.
  Email: xuejunma@ruc.edu.cn}\\
  Renmin University of China, School of Smart Governance, \\
  %Suzhou, 215006, Jiangsu, China
}
\maketitle

\bigskip
\begin{abstract}
With the rapid development of modern scientific research, tensor data have become increasingly prevalent across a wide range of fields. How to effectively characterize the relationship between tensor predictors and a response while preserving the intrinsic structure of tensor data has therefore become an important problem in high-dimensional statistical analysis. Existing tensor regression methods mainly focus on conditional means or conditional quantiles. When the response distribution is asymmetric, heavy-tailed, or contaminated by outlying observations, the conditional modal provides a robust and interpretable alternative characterization. In this paper, we propose a partial modal tensor regression (PMTR) method. The proposed method constructs the partial modal tensor covariance, which incorporates information on the conditional modal to guide tensor dimension reduction, and subsequently establishes a modal tensor regression model in the reduced tensor space. This strategy alleviates the dimensionality burden caused by direct vectorization while preserving the intrinsic structural information of tensor  predictors. Under suitable regularity conditions, we establish the consistency and asymptotic properties of the proposed estimators and employ a data-driven dimension-selection procedure to determine the effective dimension along each tensor mode. Extensive simulation studies further demonstrate that PMTR can stably recover the underlying tensor coefficient structure and exhibits satisfactory finite-sample performance across different sample sizes, coefficient structures, error distributions, and noise levels. Finally, we apply PMTR to the Beijing air-quality data. The empirical results show that the proposed method can effectively extract low-dimensional structure and reveal interpretable associations between environmental variables at different times of the day and the conditional modal of daily average PM\(_{2.5}\) concentrations.
\end{abstract}

\vfill

\newpage
\spacingset{1.45} 

% 正文
\section{Introduction}
Tensor data, as a multidimensional or N-way array, have become increasingly prevalent in modern scientific research, arising naturally in a wide range of fields such as neuroimaging, genomics, chemometrics, and signal processing \citep{kolda2009tensor, auddy2025tensors}.
Unlike conventional vector-valued data, tensor data naturally preserve the intrinsic multiway structure and dependence across different modes of observation.
A straightforward vectorization of tensor predictors, however, is often inadequate, as it ignores their intrinsic multiway structure and leads to a regression parameter space whose dimension grows multiplicatively with the mode dimensions \citep{zhou2013tensor, zhong2015tensor}.
Consequently, statistical modeling of tensor predictors requires methods that can effectively accommodate their high dimensionality while preserving the underlying multilinear structure.
These challenges have stimulated substantial development of regression and dimension-reduction methods tailored to tensor predictors.

Dimension reduction has played a central role in the development of tensor regression. A widely used strategy is to incorporate low-rank tensor decompositions into regression modeling.
In particular, \citet{zhou2013tensor} developed a generalized linear tensor regression framework based on the CANDECOMP/PARAFAC (CP) decomposition, substantially reducing the dimensionality of the regression model through a low-rank representation of the coefficient tensor.
\citet{lock2018tensor} proposed a tensor-on-tensor regression framework with a reduced CP-rank coefficient tensor, allowing tensor predictors to be associated with tensor  responses.
Beyond low-rank parameterization, another important line of research focuses on directly extracting low-dimensional information from tensor predictors that is most relevant to the response.
\citet{zhong2015tensor} developed a tensor sufficient dimension reduction framework to identify low-dimensional subspaces that retain the regression information contained in tensor  predictors. 
Along a related direction, \citet{ZhangandLi2017TEPLS} proposed tensor envelope partial least-squares regression, which integrates tensor dimension reduction with partial least-squares modeling to extract response-relevant information more efficiently.
More recently, this supervised dimension-reduction perspective has been extended beyond conditional-mean regression. \citet{Sunetal.2025pqtr} introduced partial quantile tensor regression by combining tensor partial least-squares dimension reduction with quantile regression, thereby enabling the modeling of heterogeneous effects across different parts of the conditional response distribution.
Nevertheless, existing tensor regression and dimension-reduction methods have primarily focused on conditional means or conditional quantiles, whereas regression methods targeting the conditional modal  remain much less explored in the presence of tensor  predictors.

Beyond conditional means and quantiles, the conditional modal provides a complementary perspective on the conditional distribution by characterizing the most probable response given the predictors. 
Such a perspective can be particularly informative when the conditional distribution is asymmetric, heavy-tailed, or contaminated. 
Early work on modal regression dates back to \citet{lee1989mode}, followed by the semiparametric formulation of \citet{kemp2012regression}. 
\citet{yaoandli2014modal} developed modal linear regression based on kernel density estimation, together with an EM-type estimation procedure and corresponding asymptotic theory. 
Subsequent studies extended modal regression to nonparametric and regularized high-dimensional settings \citep{chen2016nonparametric,wang2017regularized}.
However, these methods are primarily developed for vector predictors.
For the tensor covariates, direct vectorization can lead to severe dimensionality and loss of intrinsic multiway structure, motivating the development of a structure-preserving and response-guided framework for modal regression with tensor predictors.

The main contributions of this work are summarized as follows. First, we propose a new partial modal tensor regression (PMTR) framework for characterizing the conditional modal relationship between a continuous response and high-dimensional tensor predictors.
Second, we develop a theoretical framework for PMTR and under suitable regularity conditions, we establish consistency and asymptotic properties of the proposed estimators, thereby providing a theoretical foundation for PMTR. A data-driven dimension-selection procedure is further employed to determine the effective dimension along each tensor modal.
Finally, extensive simulation studies are conducted under a variety of sample sizes, coefficient structures, error distributions, and noise levels to evaluate the finite-sample performance of PMTR. We further apply the proposed method to the Beijing air-quality data. The empirical analysis demonstrates that PMTR can effectively uncover low-dimensional structure and identify interpretable hour-specific associations between environmental variables and the conditional modal of daily average PM\(_{2.5}\) concentrations.

This paper adopts the tensor-related definitions and operations from 
\citet{ZhangandLi2017TEPLS}. To distinguish between matrices and higher-order tensors, bold uppercase letters $\mathbf{A}, \mathbf{B}, \mathbf{C}, \dots$ are used to denote matrices or vectors, while calligraphic uppercase letters $\mathcal{A}, \mathcal{B}, \mathcal{C}, \dots$ are used to denote tensors.

\section{Partial Modal Tensor Regression}

Let $Y \in \mathbb{R}$ be a response variable, $\mathcal{X} \in \mathbb{R}^{p_1 \times \cdots \times p_m}$ be an $m$-th order tensor covariate, and $\mathbf{Z} \in \mathbb{R}^{p_z}$ be a $p_z$-dimensional vector covariate. 
Given the tensor covariate $\mathcal{X} = x$ and the vector covariate $\mathbf{Z} = z$, the conditional modal of $Y$ is defined as
\[
\mathrm{Modal}(Y \mid \mathcal{X} = x, \mathbf{Z} = z)
= \arg\max_{y} f_{Y \mid \mathcal{X}, \mathbf{Z}}(y \mid x, z),
\]
where $f_{Y \mid \mathcal{X}, \mathbf{Z}}(\cdot \mid x, z)$ denotes the conditional density function of $Y$ given $\mathcal{X} = x$ and $\mathbf{Z} = z$.

Motivated by the modal regression framework in  \citet{yaoandli2014modal}, we assume that the conditional modal of the response variable $\mathbf{Y}$, given the covariates, follows a linear structure:
\begin{equation}\label{eeq:1}
\mathrm{Modal}(\mathbf{Y} \mid \mathcal{X}, \mathbf{Z})
= \alpha + \gamma^\top \mathbf{Z} + \langle \mathcal{B}, \mathcal{X} \rangle,
\end{equation}
where $\alpha \in \mathbb{R}$ is the intercept, $\gamma \in \mathbb{R}^{p_z}$ is the coefficient vector associated with $\mathbf{Z}$, and $\mathcal{B} \in \mathbb{R}^{p_1 \times \cdots \times p_m}$ is the coefficient tensor corresponding to the tensor covariate $\mathcal{X}$.

Equivalently, the model can be written as
\[
\mathbf{Y}  = \alpha  + \gamma^\top \mathbf{Z}+ \langle  \mathcal{B} , \mathcal{X} \rangle  + \varepsilon,
\]
where the error term $\varepsilon$ satisfies
\[
\mathrm{Modal}(\varepsilon \mid \mathcal{X} , \mathbf{Z} ) = 0
\]

Without loss of generality, we assume that the tensor covariate $\mathcal{X}$ and the vector covariate $\mathbf{Z}$ are centered, i.e.,
$
\mathbb{E}(\mathcal{X}) = 0, \quad \mathbb{E}(\mathbf{Z}) = 0.
$
For notational convenience, we define the augmented covariate vector and parameter vector as
$
\tilde{\mathbf{Z}} = (1, \mathbf{Z}^\top)^\top, \quad
\tilde{\boldsymbol{\gamma}} = (\alpha, \gamma^\top)^\top.
$
Then, the model \ref{eeq:1} can be rewritten as
$
\mathrm{Modal}(Y \mid \mathcal{X}, \mathbf{Z})
= (\tilde{\boldsymbol{\gamma}})^\top \tilde{\mathbf{Z}} + \langle \mathcal{B}, \mathcal{X} \rangle.
$

Let $\{(\mathcal{X}_i , \mathbf{Z}_i , \mathbf{Y}_i   ) \}_{i=1}^n$  denote an i.i.d. sample of size $n$.
For the estimation of the model in \ref{eeq:1}, we consider maximizing a kernel-based objective function:
$$
Q_h( \alpha , \gamma , \mathcal{B} ) \equiv \frac{1}{n} \sum_{i=1}^n \phi_h(\mathbf{Y}_i - 
\alpha  - \gamma^\top \mathbf{Z}_i -  \langle  \mathcal{B} , \mathcal{X}_i \rangle  ),
$$
where $\phi_h(t) = \phi(t/h)/h$, $\phi(\cdot)$ is a symmetric kernel density function (e.g., the Gaussian kernel), and $h$ is the bandwidth parameter. 
Following \citet{yaoandli2014modal}, an optimal bandwidth $\hat{h}_{opt}$ can be selected. Under suitable regularity conditions, as $n \to \infty$ and $h \to 0$, the resulting modal regression estimator achieves the convergence rate of order $O_p\big((nh^3)^{-1/2}\big)$.

However, in practice, even when the mode dimensions $p_i \ (i=1,2,\dots,m)$ of the tensor $\mathcal{X}$ are moderate, the total number of unknown parameters, $\prod_{i=1}^m p_i + p_z + 1$, can still be much larger than the sample size $n$, potentially leading to non-identifiability or overfitting, and rendering existing modal regression estimation methods infeasible.

\section{ Modal Tensor PLS Algorithm   }

To address the curse of dimensionality in tensor modal regression, we employ a Tucker decomposition of $\mathcal{X}$ guided by the partial least squares (PLS) criterion for effective dimension reduction. 
This framework was originally developed for tensor linear regression(TEPLS) by  \citet{ZhangandLi2017TEPLS} and 
later adopted by \citet{Sunetal.2025pqtr} to implement quantile tensor regression .

TEPLS assumes that the tensor $\mathcal{X}$ admits a separable Kronecker covariance structure. Specifically, the covariance of its vectorized form satisfies
\[
\mathrm{Cov}(\mathrm{vec}(\mathcal{X})) = \Sigma_{\mathcal{X}} = \Sigma_m \otimes \cdots \otimes \Sigma_1,
\]
where $\Sigma_k \in \mathbb{R}^{p_k \times p_k}$ is the mode-$k$ covariance matrix.

This assumption implies that the dependence structure of $\mathcal{X}$ is separable across different modes, which substantially reduces the number of free parameters in the covariance estimation. It has been widely utilized in the existing literature. \citep{hoff2011separable,TsiligkaridisandHero2013,ZhangandLi2017TEPLS,Sunetal.2025pqtr} 
This paper proposes the partial modal tensor covariance of the response $\mathbf{Y}$ and the tensor covariate $\mathcal{X}$ given the vector covariate $\mathbf{Z}$, for the purpose of dimensionality reduction of $\mathcal{X}$. 
Specifically, the partial modal tensor covariance is given by
$$
\operatorname{pmtcov}( \mathbf{Y} , \mathcal{X} \mid \mathbf{Z} ) = \mathbb{E} (\mathrm{R}(\mathbf{Y} - \tilde{\gamma}^\top \tilde{\mathbf{Z}} )  \mathcal{X}  )  ,
$$where $\mathrm{R}(t) = -\frac{t}{h^2} \phi_h(t)  $ and $\tilde{\gamma} = \arg\max_{\tilde{\gamma}} \frac{1}{n} \sum_{i=1}^n \phi_h(\mathbf{Y} - \tilde{\gamma}^\top \tilde{\mathbf{Z}} )$.
Here, \( \mathrm{R}(t) \) may be viewed as the modal score of \( \mathbf{Y} \) after adjusting for the marginal modal effect of \( \mathbf{Z} \).

Let
$\beta_0=\operatorname{vec}(\mathcal{B}_0)$
and
$d_h(\mathcal{X},\boldsymbol{Z})
=\mathbb{E}[\mathrm{R}(\varepsilon)
\mid\mathcal{X},\boldsymbol{Z}]$.
For
$a=(\widetilde{\boldsymbol{\gamma}}_{Y\mid Z}^{0})^\top
\widetilde{\boldsymbol{Z}}$
and
$b=\widetilde{\boldsymbol{\gamma}}_0^\top
\widetilde{\boldsymbol{Z}}
+\langle\mathcal{B}_0,\mathcal{X}\rangle$,
define
$
v(\boldsymbol{Z},\mathcal{X})
=
-\int_0^1
\mathbb{E}[
\mathrm{R}'\{Y-a-u(b-a)\}
\mid\mathcal{X},\boldsymbol{Z}
]\mathrm{d}u.
$
Provided that
$\mathbb{E}[v(\boldsymbol{Z},\mathcal{X})
\widetilde{\boldsymbol{Z}}
\widetilde{\boldsymbol{Z}}^\top]$
is nonsingular, define
$
A^\top
=
\mathbb{E}[
v(\boldsymbol{Z},\mathcal{X})
\operatorname{vec}(\mathcal{X})
\widetilde{\boldsymbol{Z}}^\top
]
{
\mathbb{E}[
v(\boldsymbol{Z},\mathcal{X})
\widetilde{\boldsymbol{Z}}
\widetilde{\boldsymbol{Z}}^\top
]
}^{-1},\\
V
=
\mathbb{E}[
v(\boldsymbol{Z},\mathcal{X})
\{
\operatorname{vec}(\mathcal{X})
-A^\top\widetilde{\boldsymbol{Z}}
\}
\{
\operatorname{vec}(\mathcal{X})
-A^\top\widetilde{\boldsymbol{Z}}
\}^\top
].
$
Furthermore, let
$
\boldsymbol{r}_h
=
\mathbb{E} [
\{
\operatorname{vec}(\mathcal{X})
-A^\top\widetilde{\boldsymbol{Z}}
\}
d_h(\mathcal{X},\boldsymbol{Z})
].
$

\begin{lemma}\label{lem:1}
Suppose that all the above expectations exist and
$\mathbb{E}[v(\boldsymbol{Z},\mathcal{X})
\widetilde{\boldsymbol{Z}}
\widetilde{\boldsymbol{Z}}^\top]$
is nonsingular. Then
\[
\operatorname{vec}\left\{
\operatorname{pmtcov}
(Y,\mathcal{X}\mid\boldsymbol{Z})
\right\}
=
-V\operatorname{vec}(\mathcal{B}_0)
+\boldsymbol{r}_h.
\]
If $d_h(\mathcal{X},\boldsymbol{Z})=0$ almost surely,
then
$\operatorname{vec}\{
\operatorname{pmtcov}(Y,\mathcal{X}\mid\boldsymbol{Z})
\}=-V\operatorname{vec}(\mathcal{B}_0)$.
Consequently, if $V$ is nonsingular,
\[
\operatorname{pmtcov}
(Y,\mathcal{X}\mid\boldsymbol{Z})=0
\quad\Longleftrightarrow\quad
\mathcal{B}_0=0.
\]
Moreover, suppose that
$g(e\mid\mathcal{X},\boldsymbol{Z})$
is three times continuously differentiable
in a neighborhood of zero, with
$g^{(1)}(0\mid\mathcal{X},\boldsymbol{Z})=0$,
and that suitable domination and integrability
conditions hold. Then
$\boldsymbol{r}_h=O(h^2)$, and hence
\[
\operatorname{vec}\left\{
\operatorname{pmtcov}
(Y,\mathcal{X}\mid\boldsymbol{Z})
\right\}
=
-V\operatorname{vec}(\mathcal{B}_0)
+O(h^2).
\]
\end{lemma}

Lemma~\ref{lem:1} shows that, after vectorization, the partial modal
tensor covariance admits a residualized pseudo-linear representation of
the coefficient tensor, together with a smoothing remainder. Under the
conditional zero-score condition and the nonsingularity of \(V\), this
representation is exact, and the partial modal tensor covariance is zero
if and only if all elements of \(\mathcal{B}_0 \) are zero. For asymmetric
conditional errors, the remainder is generally nonzero at a fixed positive bandwidth but is of order $O(h^2)$ under the stated smoothness
and integrability conditions. Based on this representation and the
separable Kronecker covariance structure, we develop the following modal
tensor PLS algorithm.

\begin{algorithm}[htbp]
\caption{Modal Tensor PLS Algorithm}
\label{alg:MTpls_c}
\begin{algorithmic}[1]
\State Initialize the partial modal tensor covariance $\mathcal{C}_{M} = \operatorname{pmtcov}(\mathbf{Y},  \mathcal{X}|\mathbf{Z})  \in \mathbb{R}^{p_1 \times \cdots \times p_m} $.
\Statex \textbf{for} $k = 1$ \textbf{to} $m$ \textbf{do}
    \State \quad Let $\mathcal{C}_{M(k)} $ be the  mode-$k$ cross covariance matrix . Standardize it as 
    \[
    \mathcal{C}_{0k}  = \mathcal{C}_{M(k)} \left( 
    \boldsymbol{\Sigma}_m^{-1/2} \otimes \dots \otimes \boldsymbol{\Sigma}_{k+1}^{-1/2} \otimes \boldsymbol{\Sigma}_{k-1}^{-1/2} \otimes \dots \otimes \boldsymbol{\Sigma}_1^{-1/2} \right) \in \mathbb{R}^{p_k \times (\prod_{j \neq k} p_j)}
    \]
    \Statex \quad \textbf{for} $s = 1$ \textbf{to} $d_k$ \textbf{do}
        \State \quad \quad Find $\boldsymbol{w}_{sk} \in \mathbb{R}^{p_k}$ that maximizes $\boldsymbol{w}^\top 
        \boldsymbol{M}_{(s-1)k} 
        \boldsymbol{w}$ subject to $\boldsymbol{w}^\top \boldsymbol{w} = 1$,where $ \boldsymbol{M}_{(s-1)k}=\mathcal{C}_{(s-1)k} \mathcal{C}_{(s-1)k}^\top $
        \State \quad \quad Update the cross covariance $\mathcal{C}_{sk} = \boldsymbol{Q}_{sk} \mathcal{C}_{0k}$, with the  projection onto the orthogonal subspace of $\operatorname{span}(\boldsymbol{\Sigma}_k \boldsymbol{W}_{sk})$, 
        \[
        \boldsymbol{Q}_{sk} = \boldsymbol{I}_{p_k \times p_k} - \boldsymbol{\Sigma}_k \boldsymbol{W}_{sk} \left( \boldsymbol{W}_{sk}^\top \boldsymbol{\Sigma}_k \boldsymbol{W}_{sk} \right)^{-1} \boldsymbol{W}_{sk}^\top \in \mathbb{R}^{p_k \times p_k}
        \]
        \quad \quad where
        $\boldsymbol{W}_{sk} = (\boldsymbol{w}_{1k}, \dots, \boldsymbol{w}_{sk} )\in \mathbb{R}^{p_k \times s}$.
    \Statex \quad \textbf{end for}
    \Statex \quad After extracting all $d_k$
  directions, collect them as $\boldsymbol{W}_k = \boldsymbol{W}_{d_k k} \in \mathbb{R}^{p_k \times d_k}$
\Statex \textbf{end for}   
\State Reduce $\mathcal{X} \in \mathbb{R}^{p_1 \times \dots \times p_m}$ to a tensor $\mathcal{T} = \left[\left[ \mathcal{X}; \boldsymbol{W}_1^\top, \dots, \boldsymbol{W}_m^\top \right]\right] \in \mathbb{R}^{d_1 \times \dots \times d_m}$ .
\State Perform the modal regression for $\mathbf{Y}$ with respect to $\mathbf{Z}$ and $\mathcal{T}$, and obtain the estimation for coefficients $\alpha$, $\boldsymbol{\gamma}$ and $\mathcal{D} \in \mathbb{R}^{d_1 \times d_2 \times \dots \times d_m}$.
\State Obtain $\mathcal{B} = \left[\left[ \mathcal{D}; \boldsymbol{W}_1, \dots, \boldsymbol{W}_m \right]\right] \in \mathbb{R}^{p_1 \times \dots \times p_m}$.
\end{algorithmic}
\end{algorithm}

Since, conditional on $\mathbb{E}(\mathrm{R}(\mathbf{Y} - \tilde{\gamma}^\top \tilde{\mathbf{Z}} ))=0 $ and $\mathbb{E}(\mathcal{X})=0$ , we have $\operatorname{pmtcov}( \mathbf{Y} , \mathcal{X} \mid \mathbf{Z} ) = \mathbb{E} (\mathrm{R}(\mathbf{Y} - \tilde{\gamma}^\top \tilde{\mathbf{Z}} )  \mathcal{X}  )  = \operatorname{cov}(\mathrm{R}(\mathbf{Y} - \tilde{\gamma}^\top \tilde{\mathbf{Z}} ) , \mathcal{X})$ the quantity $\mathrm{R}(\mathbf{Y} - \tilde{\gamma}^\top \tilde{\mathbf{Z}} )$ in our algorithm can be regarded as a variant of $\mathbf{Y}$ in the TEPLS algorithm.
In Step (2) of the algorithm, compared with $\boldsymbol{\Sigma}_\mathbf{Y}^{-1/2}$ in TEPLS, the transformed $\boldsymbol{\Sigma}_{\mathrm{R}(\mathbf{Y} - \tilde{\gamma}^\top \tilde{\mathbf{Z}})}^{-1/2}$ differs only by a multiplicative constant. In Step (3), however, $\boldsymbol{w}_{sk}$ can be obtained as the eigenvector corresponding to the largest eigenvalue of $\tilde{\mathcal{C}}_{(s-1)k} \tilde{\mathcal{C}}_{(s-1)k}^\top $ . Since multiplication by a constant does not affect the corresponding eigenvectors, this constant factor can be omitted from our algorithm.
Since eigenvectors are invariant to an overall scaling of the matrix, Steps (2)–(4) of the algorithm are scale-invariant. Specifically, multiplying the input tensor $\mathcal{X}$ by any nonzero constant leaves the resulting $\boldsymbol{W}_{sk}$  unchanged.
The treatment of a symmetric tensor $\mathcal{X}$  follows the same strategy as in Algorithm PQTR. For a symmetrically structured input tensor \(\mathcal{X}\), the algorithm preserves the symmetry and consequently produces a symmetric tensor \(\mathcal{T}\). In Step (6), it is sufficient to regress the response \(\mathbf{Y}\) on the nonredundant elements of \(\mathcal{T}\).

For the model in \eqref{eeq:1}, this section describes the implementation of Algorithm \ref{alg:MTpls_c} on the observed data. To estimate the partial modal tensor covariance $\mathcal{C}_{M}$ , we first estimate $\tilde{\gamma}$ by solving the optimization problem $\tilde{\gamma} = \arg\max_{\tilde{\gamma}} \frac{1}{n} \sum_{i=1}^n \phi_h(\mathbf{Y} - \tilde{\gamma}^\top \tilde{\mathbf{Z}} )$.
This optimization problem can be solved using the modal EM (MEM) algorithm proposed by \citet{yaoandli2014modal}. The bandwidth \(h\) in $\mathrm{R}(t) = -\frac{t}{h^2} \phi_h(t)  $  is chosen according to the optimal bandwidth used in the MEM algorithm. Consequently, the partial modal tensor covariance can be estimated by
$
\hat{\mathcal{C}_M}=
\frac{1}{n}
\Sigma_{i=1}^n
 (\mathrm{R}(\mathbf{Y}_i - \tilde{\gamma}^\top \tilde{\mathbf{Z}}_i )  \mathcal{X}_i  )  .$
Following \citet{ZhangandLi2017TEPLS}, for a set of independent and identically distributed observations, $\Sigma_k$  can be estimated by
$ \hat{\Sigma}_k = \left( n \prod_{j \neq k} p_j \right)^{-1} \sum_{i=1}^{n} \mathcal{X}_{i(k)} \mathcal{X}_{i(k)}^{\top}$.
Under the condition $n \prod_{j \neq k} p_j > p_k$, this estimator is invertible. Moreover, if the tensor \(\mathcal{X}\) follows a tensor normal distribution, there exists an invertible maximum likelihood estimator given by
$\hat{\boldsymbol{\Sigma}}_k = \frac{1}{n \prod_{j \neq k} p_j} \sum_{i=1}^n \mathcal{X}_{i(k)} \left\{ \left( \hat{\boldsymbol{\Sigma}}_m \right)^{-1} \otimes \cdots \otimes \left( \hat{\boldsymbol{\Sigma}}_{k+1} \right)^{-1} \otimes \left( \hat{\boldsymbol{\Sigma}}_{k-1} \right)^{-1} \otimes \cdots \otimes \left( \hat{\boldsymbol{\Sigma}}_1 \right)^{-1} \right\} \mathcal{X}_{i(k)}^\top$.

A key tuning parameter in the proposed algorithm is the dimension \(d_k\) selected for each mode \(k\), namely, the column rank of \(\boldsymbol{w}_k\). In this paper, we determine \(d_k\) using the ridge-type eigenvalue ratio (RER) criterion. Compared with cross-validation, the RER criterion substantially reduces the computational burden and, at the same time, avoids the instability of the conventional eigenvalue ratio (ER) criterion caused by an excessively small denominator.
For each mode $k = 1,\dots,m$ , we have  $\boldsymbol{M}_{0k} = \boldsymbol{C}_{0k}\boldsymbol{C}_{0k}^\top,\quad \widehat{\boldsymbol{M}}_{0k} = \widehat{\boldsymbol{C}}_{0k}\widehat{\boldsymbol{C}}_{0k}^\top.$
Let the ordered eigenvalues of  $\widehat{\boldsymbol{M}}_{0k}$  be
$
\widehat{\lambda}_{k1} \ge \widehat{\lambda}_{k2} \ge \dots \ge \widehat{\lambda}_{kp_k} \ge 0.
$
Given an upper bound \(l_{k,\max}\) for the candidate dimension satisfying $ r_k \le l_{k,\max} < p_k, $
define the ridge-type eigenvalue ratio as
$ \mathrm{RER}_{kl} = \frac{\widehat{\lambda}_{kl} + \rho_n}{\widehat{\lambda}_{k,l+1} + \rho_n},\quad l = 1,\dots,l_{k,\max}, $
where the ridge sequence satisfies
$
\rho_n\to0,
\qquad
n^{-4/7}=o(\rho_n).
$
The dimension estimator for the k-th  mode is then defined as
$
\hat{d}_k = \mathop{\arg\max}_{1\le l \le l_{k,\max}} \mathrm{RER}_{kl}.
$
If the maximum is attained at more than one index, we take the smallest one by convention \citep{xia2015consistently,zhu2017adaptive}

\section{Theoretical Properties}
Let $\{(Y_i, \boldsymbol{Z}_i, \mathcal{X}_i): i = 1, \dots, n\}$ be an independent and identically distributed sample, where  $Y_i \in \mathbb{R}$, $\boldsymbol{Z}_i \in \mathbb{R}^{p_z}$ and $\mathcal{X}_i \in \mathbb{R}^{p_1 \times \cdots \times p_m}$. Let$\widetilde{\boldsymbol{Z}}_i = (1 , \boldsymbol{Z}_i^\top )$.
Suppose that the conditional modal regression model is given by
$
\operatorname{Modal}(Y_i \mid \mathcal{X}_i, \boldsymbol{Z}_i) = \widetilde{\boldsymbol{\gamma}}_0^{\top} \widetilde{\boldsymbol{Z}}_i + \langle \mathcal{B}_0, \mathcal{X}_i \rangle.
$
Define the error term as
$
\varepsilon_i = Y_i - \alpha_0 - \boldsymbol{\gamma}_0^\top \boldsymbol{Z}_i - \langle \mathcal{B}_0, \mathcal{X}_i \rangle.
$
Let $g(e \mid \mathcal{X}, \boldsymbol{Z})$ denote the conditional density of $\varepsilon$ given $(\mathcal{X}, \boldsymbol{Z})$. The conditional modal assumption implies that
$
g^{(1)}(0 \mid \mathcal{X}, \boldsymbol{Z}) = 0, \quad g^{(2)}(0 \mid \mathcal{X}, \boldsymbol{Z}) < 0
$
almost surely.

Suppose that the true tensor coefficient admits the Tucker decomposition\\
$
\mathcal{B}_0 = \left[\!\left[ \mathcal{D}_0; \boldsymbol{W}_{01}, \dots, \boldsymbol{W}_{0m} \right]\!\right],
$
where $\boldsymbol{W}_{0k} $ satisfies
$\boldsymbol{W}_{0k} \in \mathbb{R}^{p_k \times r_k}$
 and 
$ r_k = \operatorname{rank} (C_{0k} ).$
Accordingly, the true dimension-reduced tensor is given by
$
\mathcal{T}_i^0 = \mathcal{X}_i \times_1 \boldsymbol{W}_{01}^\top \times_2 \cdots \times_m \boldsymbol{W}_{0m}^\top.
$
By the multilinearity of the tensor inner product, we have
$
\langle \mathcal{B}_0, \mathcal{X}_i \rangle = \langle \mathcal{D}_0, \mathcal{T}_i^0 \rangle.
$
Let
$
r = \prod_{k=1}^m r_k, \quad q = 1 + p_z + r.
$
Define the low-dimensional design vector and its associated parameter as
$
\boldsymbol{U}_i^0 =
\begin{pmatrix}
1 ,
{\boldsymbol{Z}_i }^{\top} ,
{\operatorname{vec}(\mathcal{T}_i^0)}^{\top}
\end{pmatrix}^{\top},
$
$
\boldsymbol{\eta}_0 =
\begin{pmatrix}
\alpha_0 ,
{\boldsymbol{\gamma}_0 }^{\top},
{\operatorname{vec}(\mathcal{D}_0)}^{\top}
\end{pmatrix}^{\top}.
$
Then,
$
\operatorname{Modal}(Y_i \mid \mathcal{X}_i, \boldsymbol{Z}_i) = (\boldsymbol{U}_i^0)^\top \boldsymbol{\eta}_0.
$
The corresponding estimator of the design vector is given by
$
\widehat{\boldsymbol{U}}_i = \begin{pmatrix}
\widetilde{\boldsymbol{Z}}_i^\top ,
\mathrm{vec}(\widehat{\mathcal{T}}_i)^\top
\end{pmatrix}^\top ,\quad
\widehat{\mathcal{T}}_i = \mathcal{X}_i \times_1 \widehat{\boldsymbol{W}}_1^\top \times_2 \cdots \times_m \widehat{\boldsymbol{W}}_m^\top.
$
The resulting objective function is
$
Q_{n,h}(\boldsymbol{\eta}) = \frac{1}{n}\sum_{i=1}^n \phi_h\left(Y_i - \widehat{\boldsymbol{U}}_i^\top \boldsymbol{\eta}\right).
$
Let $\widehat{\boldsymbol{\eta}}$ denote a local maximizer of this objective function within a neighborhood of $\boldsymbol{\eta}_0$ . Rearranging its last $\prod_{k=1}^m r_k$  components into a tensor yields  $\widehat{\mathcal{D}}$.
Define
$
\widehat{\mathcal{B}} = \left[\!\left[   \widehat{\mathcal{D}}; \widehat{\boldsymbol{W}}_1, \dots, \widehat{\boldsymbol{W}}_m \right]\!\right].
$
Further, define
$
\boldsymbol{K} = \mathbb{E}\left[g^{(3)}(0 \mid \mathcal{X}, \boldsymbol{Z}) \boldsymbol{U}^0\right],
$
$
\boldsymbol{L} = \mathbb{E}\left[g(0 \mid \mathcal{X}, \boldsymbol{Z}) \boldsymbol{U}^0 (\boldsymbol{U}^0)^\top\right],
$
and
$
\nu_2 = \int_{\mathbb{R}} t^2 \phi^2(t) dt.
$
The latter is obtained by performing modal regression of $Y$ on $\widetilde{\boldsymbol{Z}}$ alone.

Define the population marginal modal parameter as
$
\widetilde{\boldsymbol{\gamma}}_{Y|Z}^0 = \mathop{\arg\max}_{\widetilde{\boldsymbol{\gamma}} \in \Gamma} \mathbb{E} \left[ \phi_{h_Z} \left( Y - \widetilde{\boldsymbol{\gamma}}^{\top} \widetilde{\boldsymbol{Z}} \right) \right],
$
where $\phi_{h_Z}(u) = h_Z^{-1} \phi(u/h_Z)$ and $\phi$ denotes the standard normal density. Its sample estimator is defined by
$
\widehat{\widetilde{\boldsymbol{\gamma}}}_{Y|Z} = \mathop{\arg\max}_{\widetilde{\boldsymbol{\gamma}} \in \Gamma} \frac{1}{n} \sum_{i=1}^n \phi_{h_Z} \left( Y_i - \widetilde{\boldsymbol{\gamma}}^{\top} \widetilde{\boldsymbol{Z}}_i \right).
$
Define the population and sample marginal modal residuals, respectively, as
$
e_i^0 = Y_i - \left( \widetilde{\boldsymbol{\gamma}}_{Y|Z}^0 \right)^{\top} \widetilde{\boldsymbol{Z}}_i, \quad
\widehat{e}_i = Y_i - \widehat{\widetilde{\boldsymbol{\gamma}}}_{Y|Z}^{\top} \widetilde{\boldsymbol{Z}}_i.
$
Further, let
$
\boldsymbol{\Omega}_{-k} = \boldsymbol{\Sigma}_m^{-1/2} \otimes \cdots \otimes \boldsymbol{\Sigma}_{k+1}^{-1/2} \otimes \boldsymbol{\Sigma}_{k-1}^{-1/2} \otimes \cdots \otimes \boldsymbol{\Sigma}_1^{-1/2},
$

Then, for each mode $k=1,\dots,m$,
$
\mathcal{C}_{0k} = \mathcal{C}_{\mathrm{M}(k)} \boldsymbol{\Omega}_{-k}.
$
Let $\boldsymbol{w}_{1k}^0,\dots,\boldsymbol{w}_{r_kk}^0$  denote, in order, the true parameter values corresponding to $\boldsymbol{w}_{1k},\dots,\boldsymbol{w}_{r_kk}$, and write
$
\boldsymbol{W}_{sk}^0 = \big(\boldsymbol{w}_{1k}^0,\dots,\boldsymbol{w}_{sk}^0\big).
$
Accordingly, the \(s-th\) deflated matrices are given by
$
\boldsymbol{Q}_{sk}^0 = \boldsymbol{I}_{p_k} - \boldsymbol{\Sigma}_k \boldsymbol{W}_{sk}^0 \big[\big(\boldsymbol{W}_{sk}^0\big)^\top \boldsymbol{\Sigma}_k \boldsymbol{W}_{sk}^0\big]^{-1} \big(\boldsymbol{W}_{sk}^0\big)^\top.
$
and
$
\mathcal{C}_{sk} = \boldsymbol{Q}_{sk}^0 \mathcal{C}_{0k},  s=1,\dots,r_k-1.
$
where, by convention, $\boldsymbol{Q}_{0k}^0 = \boldsymbol{I}_{p_k}$, and hence $\mathcal{C}_{0k} = \boldsymbol{Q}_{0k}^0 \mathcal{C}_{0k}$.
The corresponding sample estimators are denoted by $\widehat{\mathcal{C}}_{0k}$,$\widehat{\mathcal{C}}_{sk}$,$\widehat{\boldsymbol{M}}_{sk}$,$\widehat{\boldsymbol{Q}}_{sk}$ and $\widehat{\boldsymbol{w}}_{s+1,k}$ respectively.

\begin{enumerate}[label=(A\arabic*),leftmargin=2.8em]
\item \label{assum:A1} The observations $\{(Y_i, \boldsymbol{Z}_i, \mathcal{X}_i)\}_{i=1}^n$ are i.i.d.. The dimensions $p_z,p_1,\dots,p_m,r_1,\dots,r_m$ are fixed constants.

\item \label{assum:A2} The parameter space \(\Gamma\) and the final low-dimensional parameter space are compact, and the true parameters lie in the interiors of their respective parameter spaces.

\item \label{assum:A4} The marginal modal regression estimator satisfies
$
\left\| \widehat{\widetilde{\boldsymbol{\gamma}}}_{Y|Z} - \widetilde{\boldsymbol{\gamma}}_{Y|Z}^0 \right\|_2 = O_p(a_n), \quad a_n = h_Z^2 + (n h_Z^3)^{-1/2}.
$
and
$
a_n^2 = o(n^{-1/2}).
$
In particular, when $h_Z \asymp n^{-1/7}$,
$a_n = O(n^{-2/7})$
and
$a_n^2 = O(n^{-4/7}) = o(n^{-1/2}).$

\item \label{assum:A5} The partial modal tensor covariance satisfies the following orthogonality condition with respect to the marginal modal parameter:
$
\mathbb{E}\left[ \mathrm{R}'(e^0) \mathrm{vec} (\mathcal{X}) \widetilde{\boldsymbol{Z}}^\top \right] = \boldsymbol{0}.
$

\item \label{assum:A6} 
$
\mathbb{E}\|\mathcal{X}\|_F^4 < \infty,\quad \mathbb{E}\|\widetilde{\boldsymbol{Z}}\|_2^4 < \infty,\quad \mathbb{E}\left( \|\mathcal{X}\|_F^2 \|\widetilde{\boldsymbol{Z}}\|_2^4 \right) < \infty.
$

\item \label{assum:A7} For each mode, the covariance matrix $\boldsymbol{\Sigma}_k$ is positive definite, and its smallest and largest eigenvalues are uniformly bounded away from zero and infinity, respectively.

\item \label{assum:A8} For each \(k\) and \(s=0,\ldots,r_k-1\), the largest eigenvalue of \(\boldsymbol{M}_{sk}\) is simple. Moreover, there exists a constant \(\Delta>0\) such that
$
\lambda_1(\boldsymbol{M}_{sk})-\lambda_2(\boldsymbol{M}_{sk})\ge \Delta.
$

\item \label{assum:A9} For each mode \(k\), the smallest eigenvalue of
$
(\boldsymbol{W}_{sk}^0)^\top \boldsymbol{\Sigma}_k \boldsymbol{W}_{sk}^0
$
is uniformly bounded away from zero. The sign of the sample eigenvector is identified by imposing
$
\widehat{\boldsymbol{w}}_{sk}^\top \boldsymbol{w}_{sk}^0 \ge 0.
$

\item \label{assum:A10} The conditional error density \(g(e\mid \mathcal{X},\boldsymbol{Z})\) in the modal regression model has four continuous derivatives in a neighborhood of zero, and these derivatives are dominated by an integrable envelope. Moreover,
$
g^{(1)}(0\mid \mathcal{X},\boldsymbol{Z})=0
$
almost surely, and
$
\boldsymbol{J}
=
\mathbb{E}\!\left[
g^{(2)}(0\mid \mathcal{X},\boldsymbol{Z})
\boldsymbol{U}^0(\boldsymbol{U}^0)^\top
\right]
$
satisfies that \(-\boldsymbol{J}\) is positive definite.

\item \label{assum:A11}  $h$ satisfies
$
h \to 0,\quad n h^5 \to \infty.
$

\item \label{assum:A12}
Let $f_n(e\mid\mathcal X,\boldsymbol Z)$ denote the conditional
density of the population marginal modal residual $e^0$
given $(\mathcal X,\boldsymbol Z)$, where the subscript $n$
reflects its dependence on $h_Z$.
This density is three times continuously differentiable in $e$.
Moreover, there exists a deterministic constant $M<\infty$
such that, almost surely,
\[
\sup_{n\geq1}\sup_{e\in\mathbb R}
\left|
f_n^{(j)}(e\mid\mathcal X,\boldsymbol Z)
\right|
\leq M,
\qquad j=0,1,2,3,
\]
where $f_n^{(0)}=f_n$.

\item \label{assum:A13}
The covariates satisfy
\[
\mathbb E\|\mathcal X\|_F^6
+
\mathbb E\|\widetilde{\boldsymbol Z}\|_2^6
<
\infty.
\]

\end{enumerate}

\begin{lemma}\label{lem:2}
Under Assumptions~\ref{assum:A1}, \ref{assum:A4},
\ref{assum:A6}, and~\ref{assum:A12},
if the deterministic bandwidth satisfies $h_Z\asymp n^{-1/7}$,
then
\[
\left\|
\widehat{\mathcal C}_{\mathrm M}
-
\mathcal C_{\mathrm M}
\right\|_F
=
O_p\!\left((nh_Z^3)^{-1/2}\right)
=
O_p(n^{-2/7}).
\]
\end{lemma}

\begin{lemma}\label{lem:3}
Under the conditions of Lemma~\ref{lem:2} and
Assumption~\ref{assum:A7}, for each mode $k$, let
$
c_{0k}=\prod_{\ell\neq k}c_\ell^{-1/2},
$
where $c_\ell>0$ is the proportionality constant satisfying
$\mathbb E(\widehat{\boldsymbol{\Sigma}}_\ell)
=c_\ell\boldsymbol{\Sigma}_\ell$
for the empirical covariance estimator defined above.

\[
\sqrt{nh_Z^3}
\left\|
\widehat{\mathcal C}_{0k}
-
c_{0k}\mathcal C_{0k}
\right\|_{\mathrm{op}}
=
O_p(1).
\]
\end{lemma}

In Lemma~\ref{lem:4}, $\widehat{\boldsymbol W}_k$
contains the first $r_k$ sample directions, with their signs
chosen as in Assumption~\ref{assum:A9}.
The dependence of the population directions and deflated
matrices on $h_Z$ is suppressed.
The lower bounds in Assumptions~\ref{assum:A8}
and~\ref{assum:A9} are understood to hold uniformly in $n$.

\begin{lemma}\label{lem:4}
Under the conditions of Lemma~\ref{lem:3} and
Assumptions~\ref{assum:A8}--\ref{assum:A9},
for each mode $k$,
\[
\sqrt{nh_Z^3}
\left\|
\widehat{\boldsymbol W}_k
-
\boldsymbol W_{0k}
\right\|_F
=
O_p(1).
\]
\end{lemma}

Define the oracle objective function by
\[
Q_{n,h}^0(\boldsymbol\eta)
=
\frac1n\sum_{i=1}^n
\phi_h\!\left(
Y_i-(\boldsymbol U_i^0)^\top\boldsymbol\eta
\right).
\]

\begin{lemma}\label{lem:5}
Under Assumptions~\ref{assum:A1}--\ref{assum:A13},
if $h_Z\asymp h\asymp n^{-1/7}$, then, for every
deterministic sequence $\delta_n=O(n^{-2/7})$,
\[
\sup_{\|\boldsymbol\eta-\boldsymbol\eta_0\|_2\leq\delta_n}
\left\|
\nabla Q_{n,h}(\boldsymbol\eta)
-
\nabla Q_{n,h}^0(\boldsymbol\eta)
\right\|_2
=
O_p(n^{-2/7}).
\]
\end{lemma}

Building on the above auxiliary lemmas, we now establish the main theoretical result. The following theorem characterizes the asymptotic behavior of the proposed estimator under the stated assumptions.
Following \citet{Sunetal.2025pqtr}, for any two parameter values $  \boldsymbol{\theta}_j = (\alpha_j, \boldsymbol{\gamma}_j, \mathcal{B}_j),  j = 1,2,  $   define the distance metric
$
d(\boldsymbol{\theta}_1, \boldsymbol{\theta}_2) = \Big\{ (\alpha_1 - \alpha_2)^2 + \|\boldsymbol{\gamma}_1 - \boldsymbol{\gamma}_2\|_2^2 + \|\mathcal{B}_1 - \mathcal{B}_2\|_F^2 \Big\}^{1/2}.
$

The dependence of the population quantities on $h_Z$ is
suppressed in the notation. All lower bounds in
Assumptions~\ref{assum:A7}--\ref{assum:A10} hold uniformly
in $n$. In particular, there exists a constant $c>0$ such that
$\lambda_{\min}(-\boldsymbol{J})\geq c$.
The true parameters remain a positive distance from the
boundaries of their parameter spaces.

\begin{theorem}\label{thm:1}
Under Assumptions~\ref{assum:A1}--\ref{assum:A13},
suppose that $d_k=r_k$ for $k=1,\ldots,m$ and
$h_Z\asymp h\asymp n^{-1/7}$.
Then, with probability tending to one, $Q_{n,h}$ admits
a local maximizer $\widehat{\boldsymbol{\eta}}$ satisfying
\[
\left\|
\widehat{\boldsymbol{\eta}}-\boldsymbol{\eta}_0
\right\|_2
=O_p(n^{-2/7}).
\]
The corresponding full parameter estimator satisfies
\[
d(\widehat{\boldsymbol{\theta}},\boldsymbol{\theta}_0)
=O_p(n^{-2/7}).
\]
\end{theorem}

The asymptotic results in Theorem \ref{thm:1} are derived under correct specification of the mode-wise reduction dimensions. This assumption is useful for characterizing the oracle behavior of the estimator, but the true dimensions $r_k$ are generally unknown in practice. We therefore consider a fully data-driven dimension selection procedure based on adjacent eigenvalue ratios of $\widehat{\boldsymbol{M}}_{0k}$. The next theorem establishes the selection consistency of this RER criterion and hence provides theoretical justification for replacing the unknown dimensions $r_k$ by their data-driven estimates.

For each mode $k$, the candidate upper bound satisfies
$1\leq r_k\leq l_{k,\max}<p_k$.
The population matrices may depend on $h_Z$, and
the signal condition below is understood uniformly in $n$.

\begin{theorem}\label{thm:2}
Under the conditions of Lemma~\ref{lem:3}, suppose that,
for each $k=1,\ldots,m$,
\[
\operatorname{rank}(\boldsymbol{C}_{0k})=r_k,
\qquad
\inf_n\lambda_{r_k}(\boldsymbol{M}_{0k})>0.
\]
If $\rho_n\to0$ and $n^{-4/7}=o(\rho_n)$, then
\[
\operatorname{Pr}\left(
\bigcap_{k=1}^m\{\hat d_k=r_k\}
\right)\to1.
\]
\end{theorem}

Theorem~\ref{thm:2} establishes the selection consistency
of the RER criterion. The signal eigenvalues remain
bounded away from zero, whereas the noise eigenvalues
are of order $O_p(n^{-4/7})$.
The ridge condition makes these noise eigenvalues
negligible relative to $\rho_n$.
Consequently, the ratio at the true dimension diverges,
while all competing ratios remain stochastically bounded.

The detailed derivation of the above results is provided in Appendix A .

\section{Simulation Studies}

In this section, we conduct a series of simulation studies to evaluate the finite-sample performance of the proposed PMTR method under a variety of data-generating scenarios. In particular, we investigate the effects of the error distribution, coefficient structure, sample size, and noise level on the performance of the proposed estimator. 
The performance of PMTR is mainly evaluated through the recovery of the coefficient tensor and the corresponding estimation error.

We generate the response as:
$
Y_i
=
\alpha_0
+
\gamma_0^{\top} Z_i
+
\langle \mathcal{B}_0, \mathcal{X}_i \rangle
+
\varepsilon_i,
 i=1,\ldots,n,
$
where $\mathcal{X}_i \in \mathbb{R}^{50\times 50} $ denotes the tensor predictor, $Z_i$ is a low-dimensional covariate vector, $\mathcal{B}_0 \in \mathbb{R}^{50\times 50} $ is the true tensor coefficient, and the error term $\varepsilon_i$ is generated such that its modal is zero.
The tensor predictor $\mathcal{X}_i$ is generated from a tensor normal distribution with mode-wise covariance matrices $\Sigma_k$, following the envelope covariance structure adopted in  \citet{Sunetal.2025pqtr, ZhangandLi2017TEPLS}.
The true tensor coefficient $\mathcal{B}_0$ is generated under seven structural patterns: square-shaped, T-shaped, cross-shaped, bi-circle-shaped, frame-shaped, disk-shaped, and triangle-shaped.
We set
$
\alpha_0 = 0,
\gamma_0 = (-1,1)^{\top}.
$
The low-dimensional covariate
$
Z_i = (Z_{i1}, Z_{i2})^{\top}
$
is generated with
$
Z_{i1} \sim N(0,1),
Z_{i2} \sim \operatorname{Bernoulli}(0.5),
$
independently of $\mathcal{X}_i$.
The error term $\varepsilon_i$ is generated from Gaussian, $t_3$, log-normal, and contaminated normal distributions, with appropriate location adjustments to ensure that the modal is zero. 

For each simulation setting, the experiment is independently repeated \(R=100\) times. To assess the recovery of the underlying coefficient structure, we report the empirical mean of the estimated coefficient tensors over the Monte Carlo replications. Specifically, letting \(\widehat{\mathcal B}^{(r)}\) denote the coefficient estimate obtained from the \(r\)th replication, the average estimate is defined as
$ \overline{\mathcal B} = \frac{1}{R}\sum_{r=1}^{R}\widehat{\mathcal B}^{(r)}. $
In addition to the visual comparison of the estimated and true coefficient tensors, estimation accuracy is quantified by the absolute Frobenius error
$ \mathrm{FE}^{(r)} = \left\| \widehat{\mathcal B}^{(r)}-\mathcal B_0 \right\|_F, $
where a smaller value indicates more accurate estimation. The distributions of the Frobenius errors across the Monte Carlo replications are further summarized using boxplots.
Let
$
\eta_i=\alpha_0+\gamma_0^{\top}Z_i+\langle\mathcal{B}_0,\mathcal{X}_i\rangle
$denote the signal component, and let \(s_\eta\) be its sample standard deviation. For a given noise level \(\delta\), the error term is scaled so that its standard deviation is \(\delta s_\eta\). For example, a noise level of \(10\%\) means that the standard deviation of the error term is set to \(0.1s_\eta\).

\begin{figure}[p]
    \centering

  % 第一行：Shape 1 和 Shape 2
  \shapebox{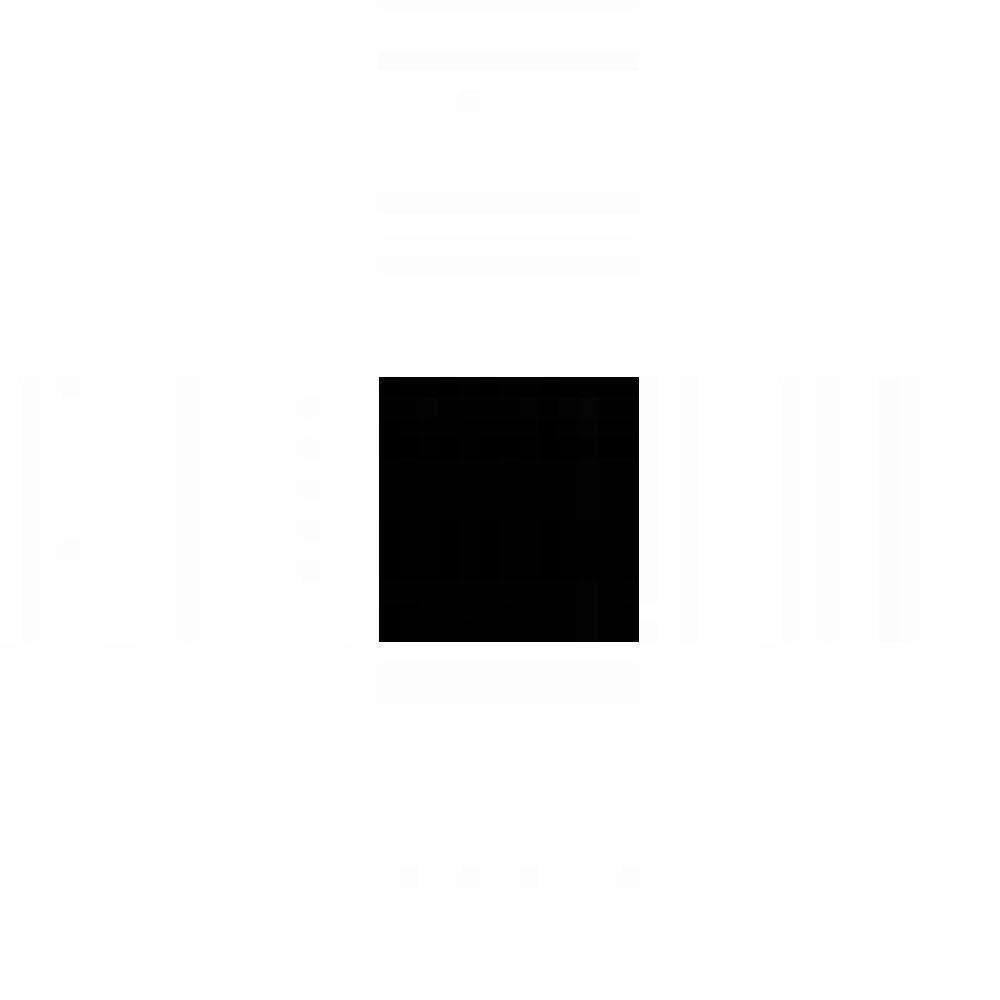}
  \hfill
  \shapebox{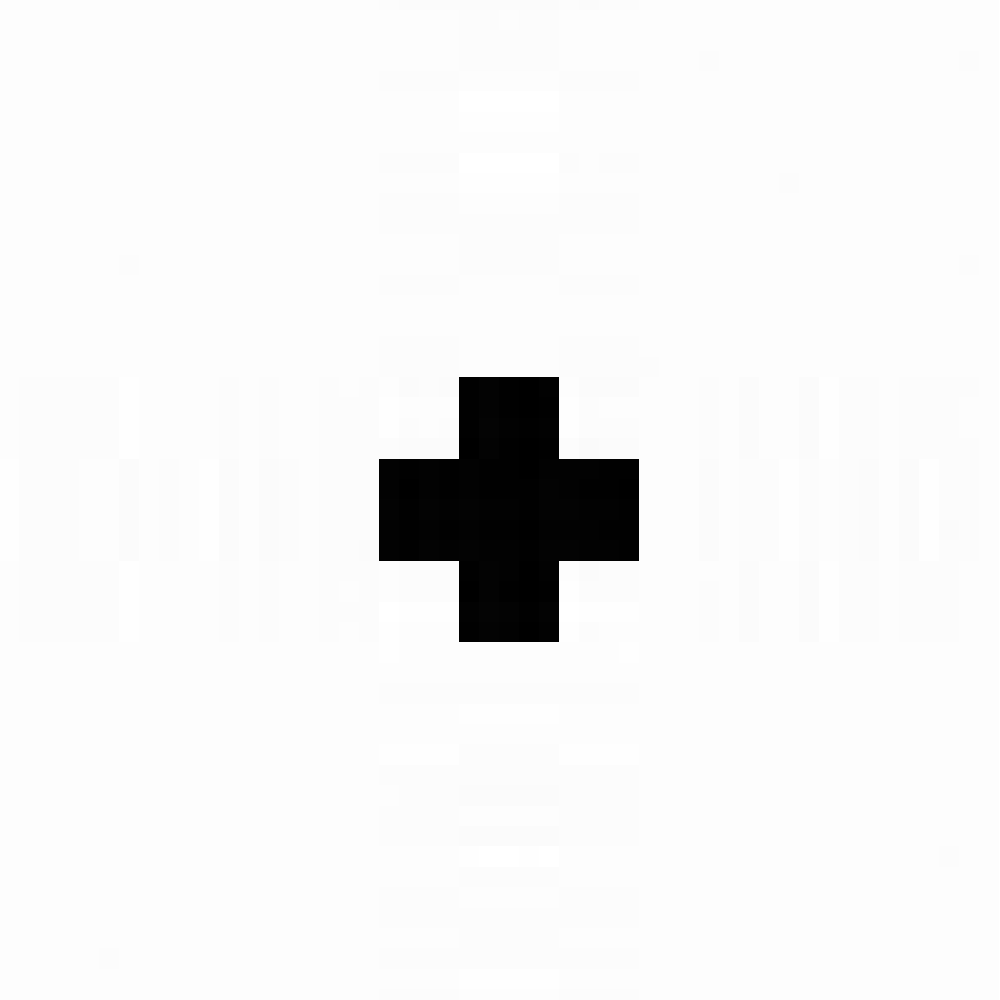}

  \par\vspace{5mm}

  % 第二行：Shape 3 和 Shape 4
  \shapebox{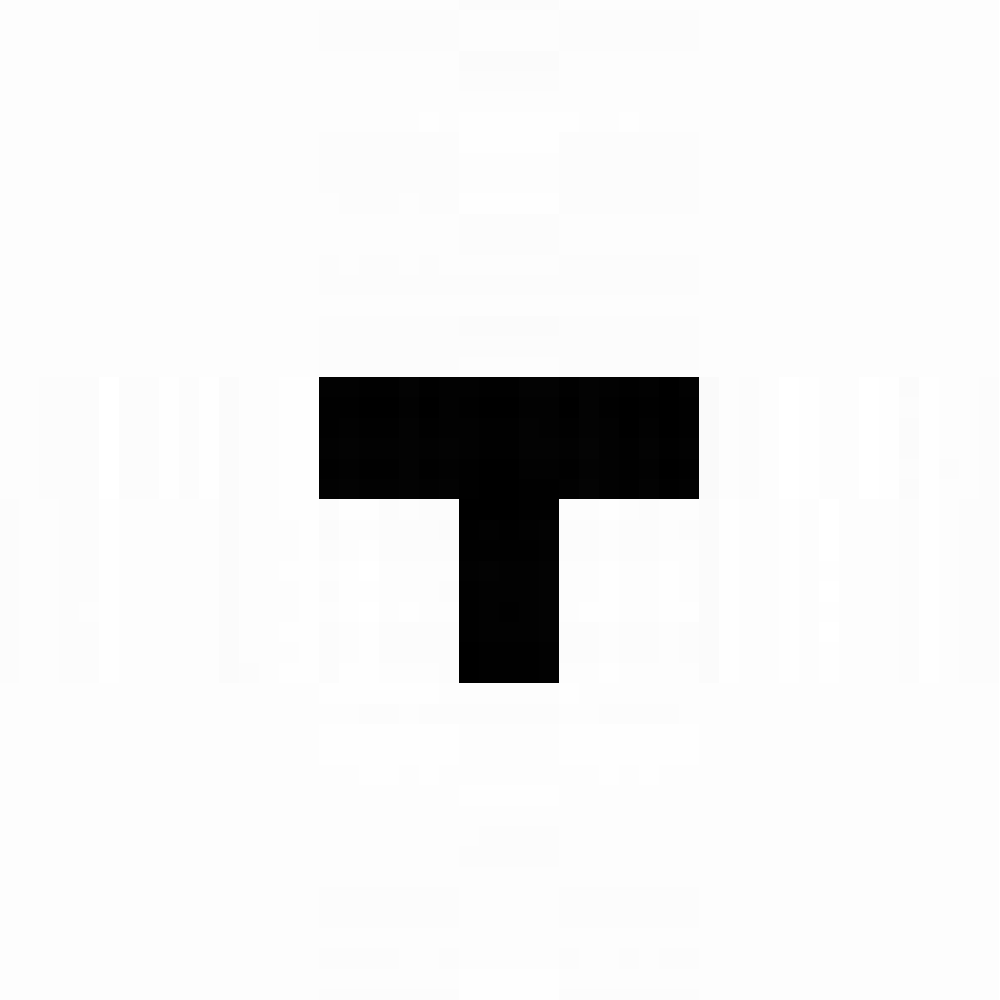}
  \hfill
  \shapebox{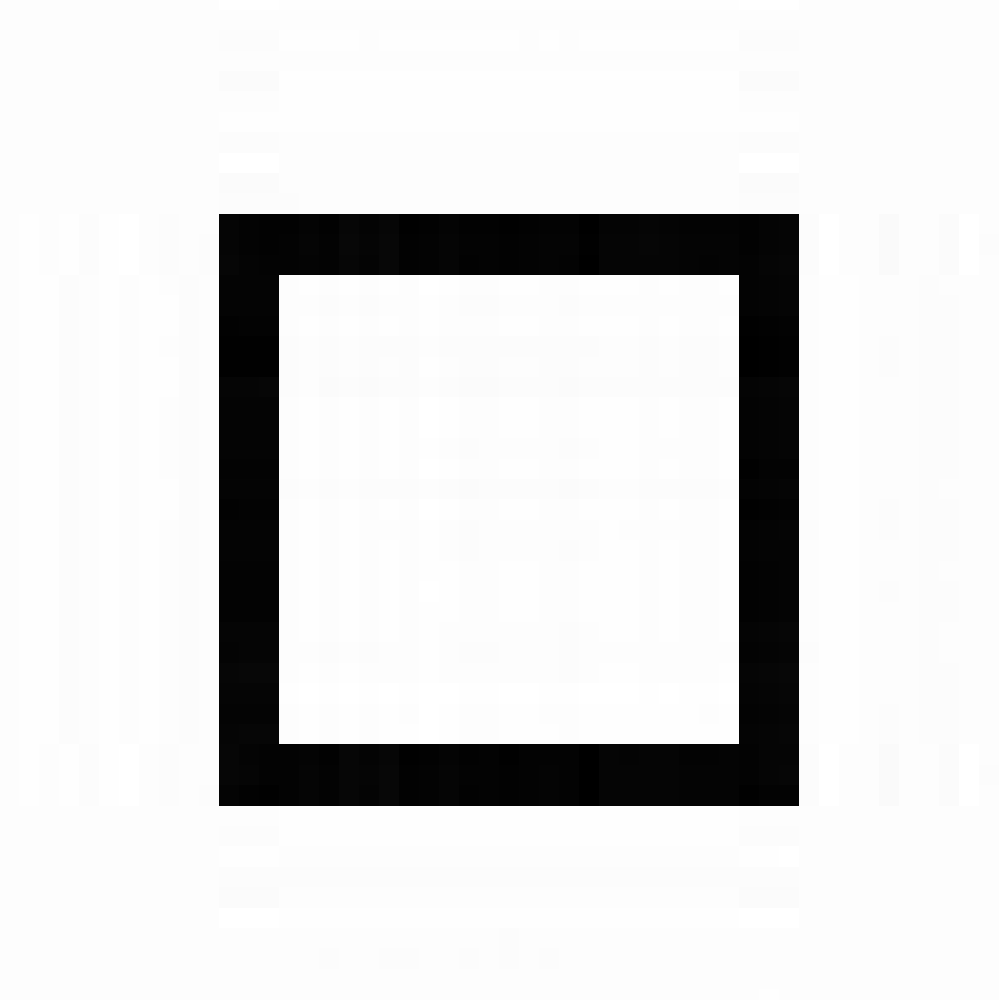}
      \caption{
    Average estimated coefficient tensors for lower-rank coefficient structures
    under different error distributions and sample sizes.
    The four structures are square-shaped (rank 1), cross-shaped (rank 2),
    T-shaped (rank 2), and frame-shaped (rank 2).
    The results are averaged over 100 Monte Carlo replications. LogN and Cont. denote the log-normal and contaminated normal distributions, respectively.
    }
    \label{fig1}
\end{figure}

\begin{figure}[p]
  \centering
  % 第一个区域：Shape 6
  \shapebox{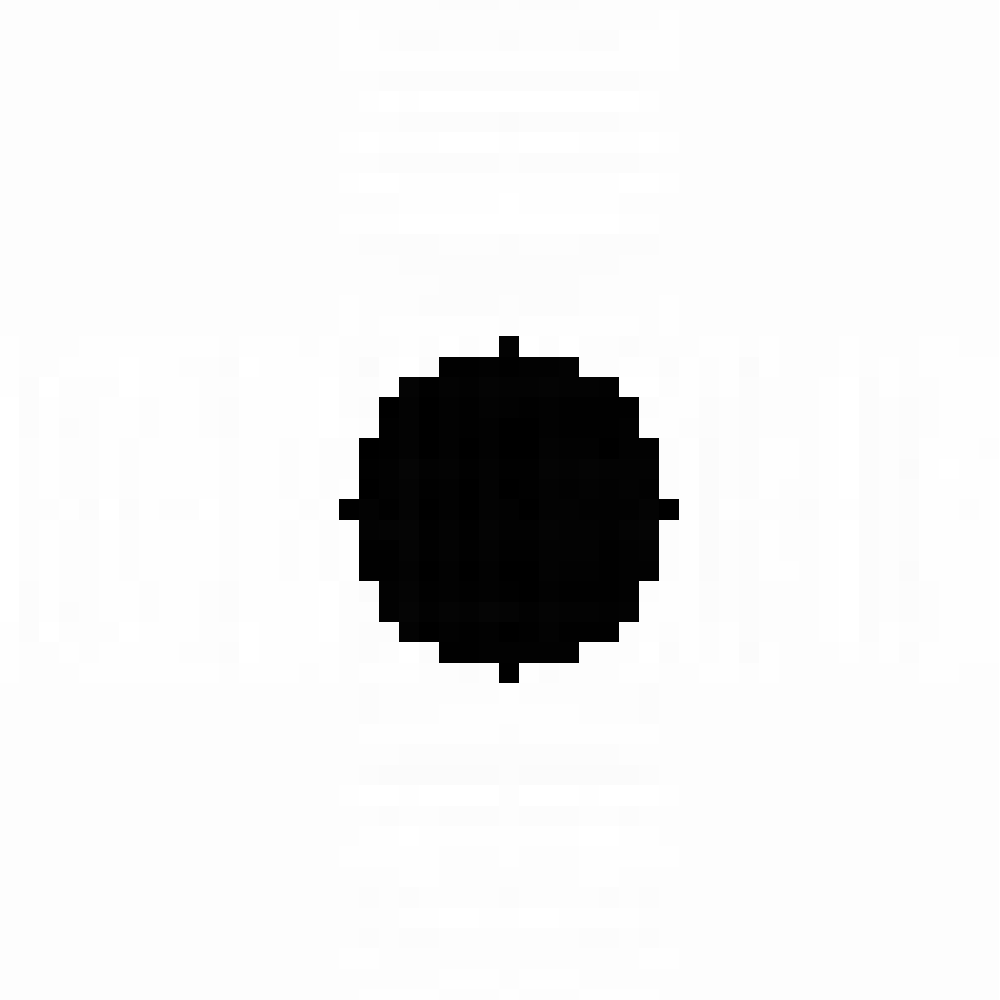}

  \par\vspace{4mm}

  % 第二个区域：Shape 5
  \shapebox{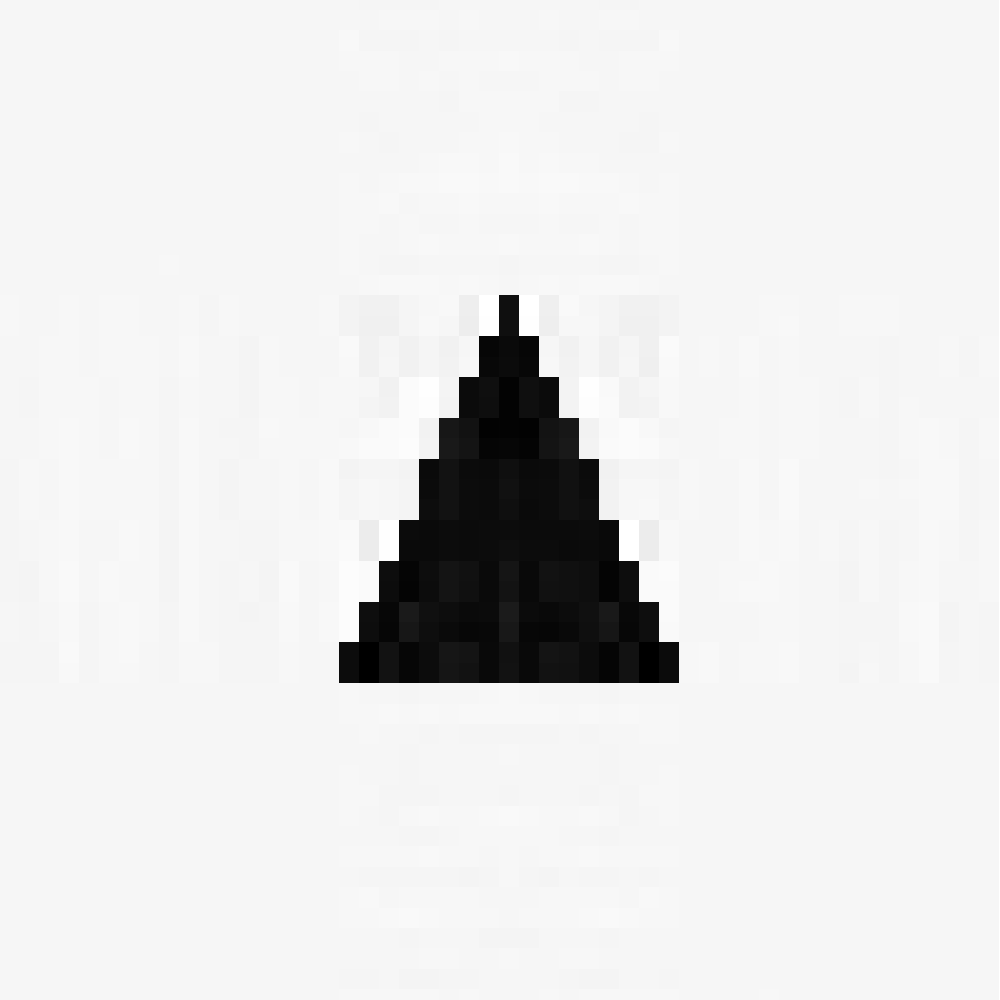}

  \par\vspace{4mm}

  % 第三个区域：Shape 7
  \shapebox{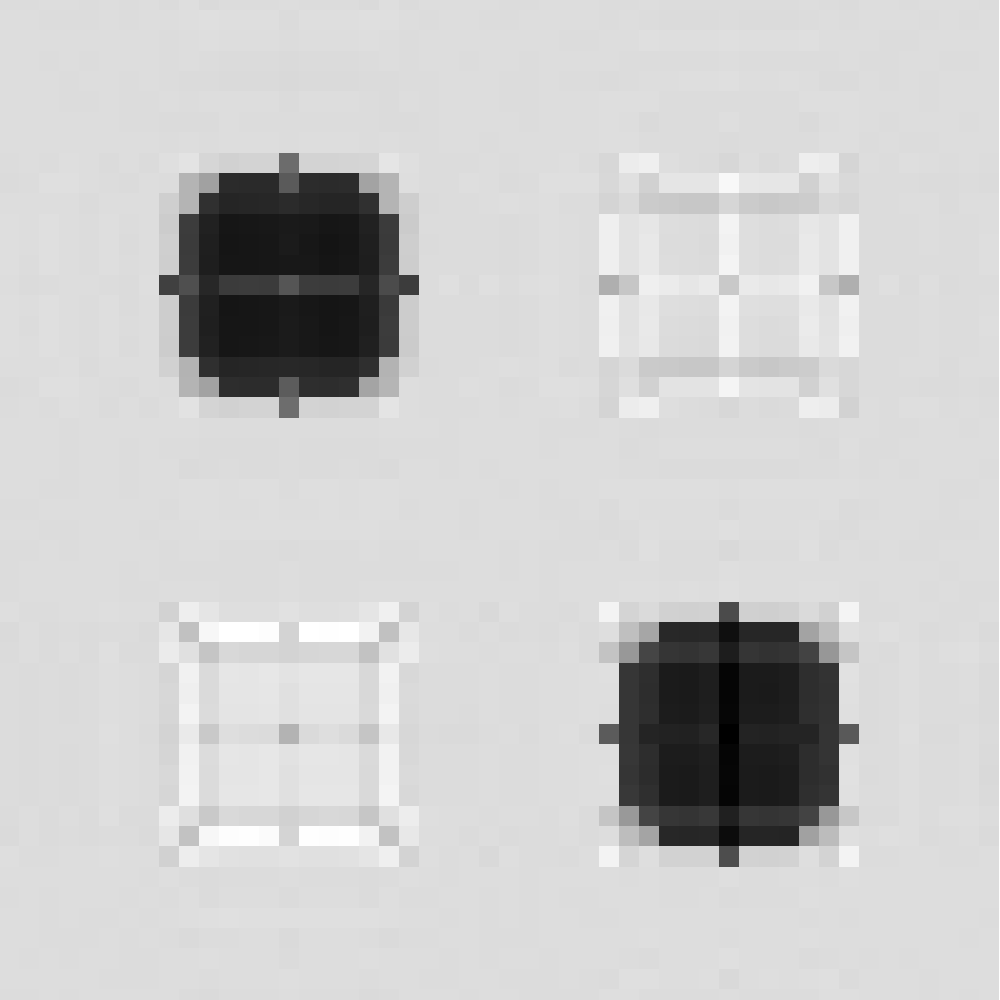}

  \caption{
    Average estimated coefficient tensors for higher-rank coefficient structures
    under different error distributions and sample sizes.
    The three structures are disk-shaped (rank 6), triangle-shaped (rank 9),
    and bi-circle-shaped (rank 10).
    The results are averaged over 100 Monte Carlo replications. 
  }
  \label{fig2}
\end{figure}

Figure\ref{fig1} and Figure\ref{fig2} present the estimated coefficient tensors under different sample sizes and error distributions.
To further investigate the effect of coefficient-structure complexity on estimation performance, we separately report the estimation results for lower-rank (square-shaped (rank 1), cross-shaped (rank 2), T-shaped (rank 2), and frame-shaped (rank 2) )and higher-rank (disk-shaped (rank 6), triangle-shaped (rank 9), and bi-circle-shaped (rank 10)) coefficient structures.
We examine the coefficient estimation performance under different coefficient structures, with sample sizes \(n=50,100,200,\) and \(400\), and four error distributions: Gaussian, \(t_3\), log-normal, and contaminated normal distributions. As shown in Figure\ref{fig1} , for the lower-rank coefficient structures, PMTR accurately recovers the main structural features of the true coefficient tensors across different sample sizes and error distributions, with clearly identified nonzero regions and structural patterns. Estimation becomes more challenging for the higher-rank settings presented in Figure\ref{fig2} . In particular, for the triangle-shaped (rank 9) and bi-circle-shaped (rank 10) structures, some spurious signals are observed when the sample size is relatively small (\(n=50\) or \(100\)). 
This observation is also supported by the absolute Frobenius errors reported in Figure \ref{fig4} (a). As the rank of the coefficient tensor increases, the absolute Frobenius error generally exhibits an increasing trend, indicating that higher-rank coefficient structures pose greater challenges for coefficient estimation.
As the sample size increases, however, these spurious signals are substantially reduced and the underlying coefficient structures are recovered with increasing accuracy. When the sample size reaches \(n=200\) or \(400\), the estimated coefficient tensors clearly reproduce the main structural features of their true counterparts. These results indicate that, although higher-rank coefficient structures pose greater estimation challenges in small samples, the estimation accuracy of PMTR improves markedly as the sample size increases.
The above observations are further supported by the absolute Frobenius estimation errors. As shown in Figure\ref{fig4} (b), the absolute Frobenius errors exhibit a clear decreasing trend as the sample size increases across the simulation settings. For the \(50\times50\) coefficient tensors, when \(n=50\), the absolute Frobenius errors are mainly concentrated between 10 and 20. As the sample size increases, the error distributions shift toward smaller values. In particular, when \(n\) reaches 200, the Frobenius errors are approximately 5 or lower in most settings. These quantitative results are consistent with the coefficient recovery patterns observed in Figure\ref{fig1}  and Figure\ref{fig2} , further demonstrating that the finite-sample estimation accuracy of PMTR steadily improves with increasing sample size.

\begin{figure}[htbp]
\centering
  \setlength{\tabcolsep}{5pt}

  \begin{tabular}{ccccc}
    \toprule
    Error distribution
    & $10\%$
    & $50\%$
    & $100\%$
    & $200\%$
    \\
    \midrule

    Normal
    &
    \panelimage{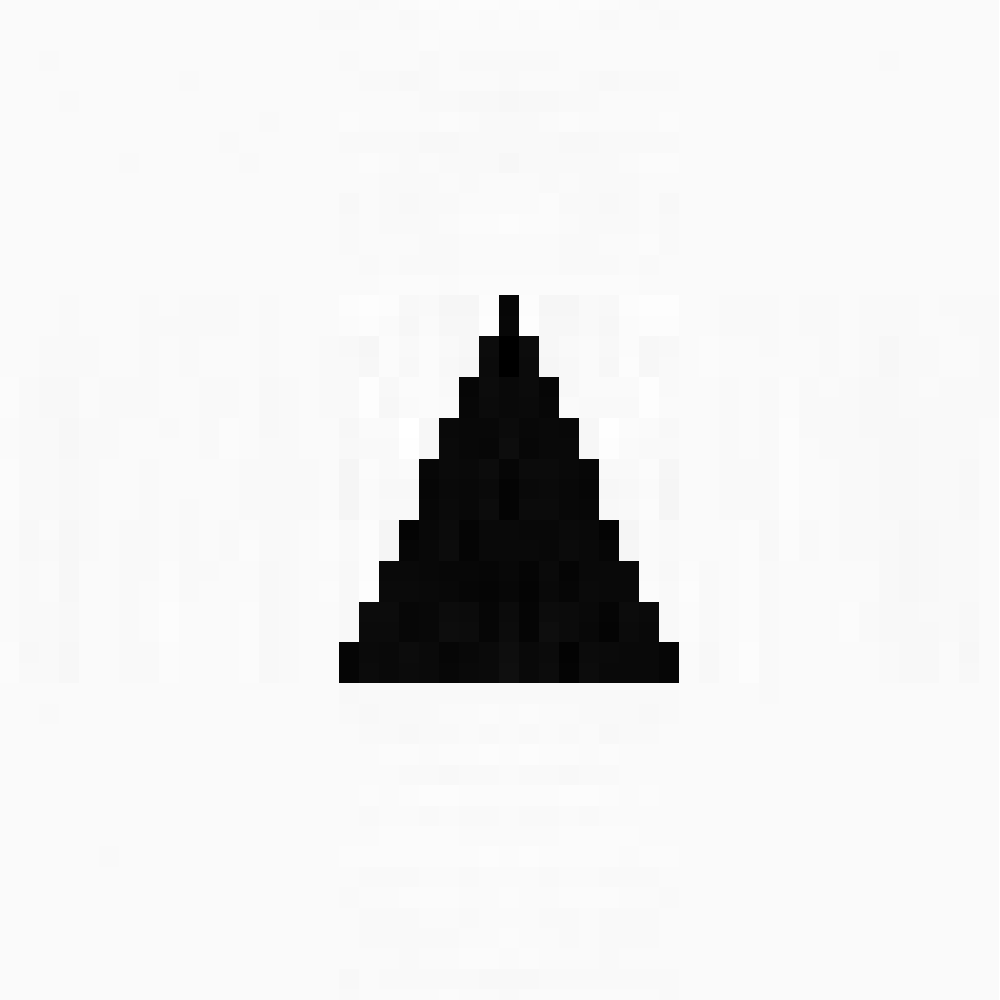}
    &
    \panelimage{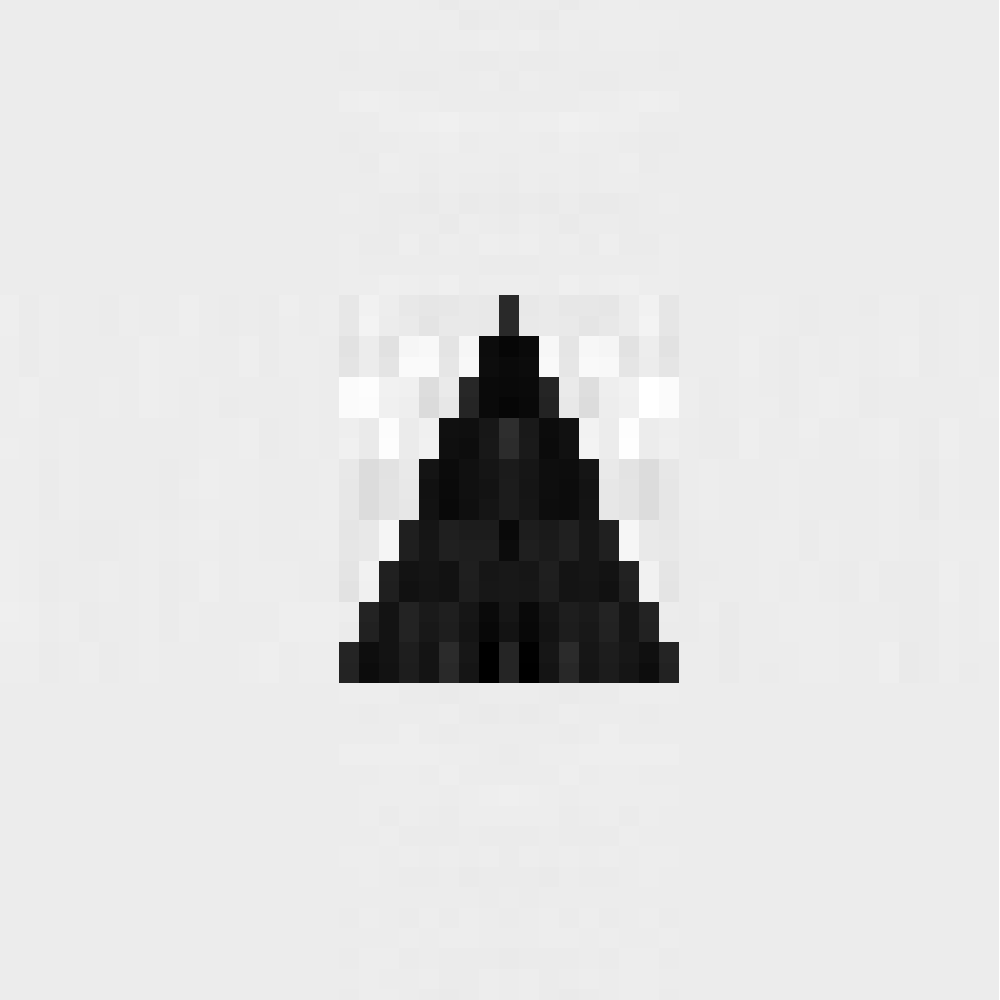}
    &
    \panelimage{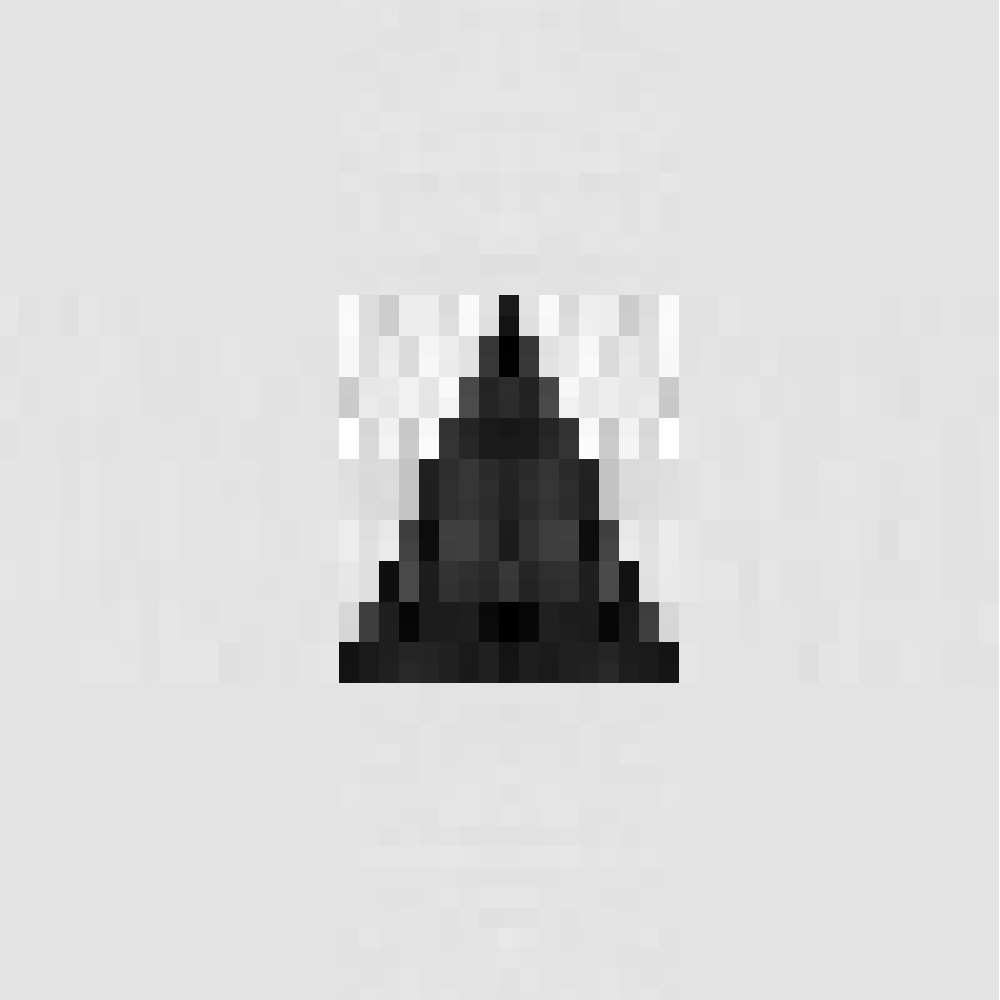}
    &
    \panelimage{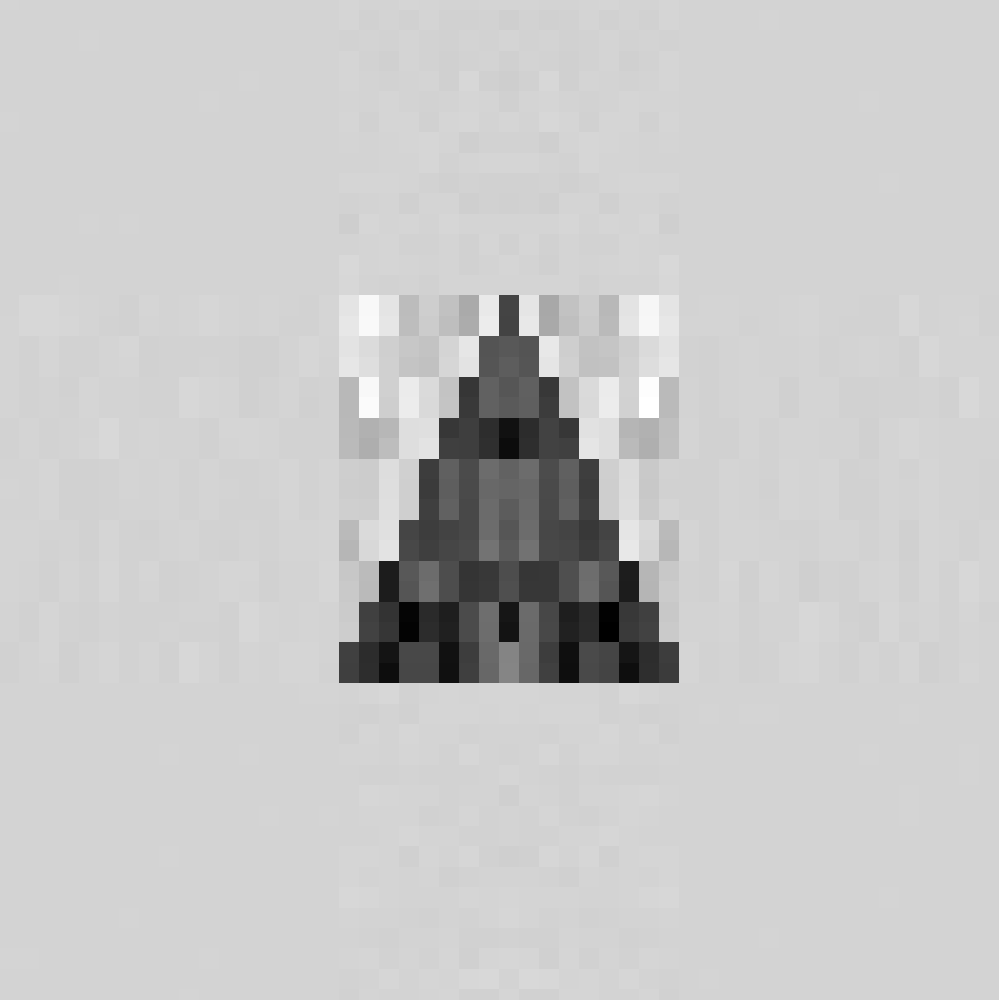}
    \\

    $t_3$
    &
    \panelimage{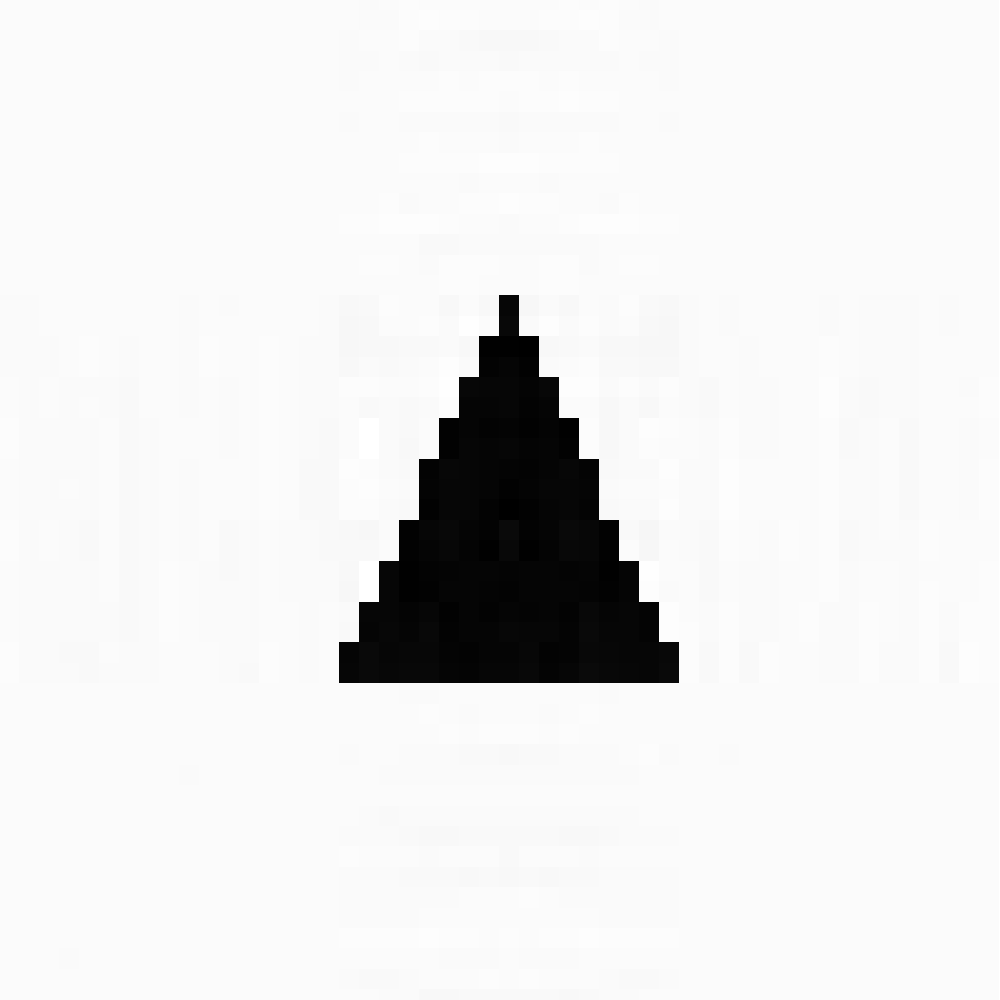}
    &
    \panelimage{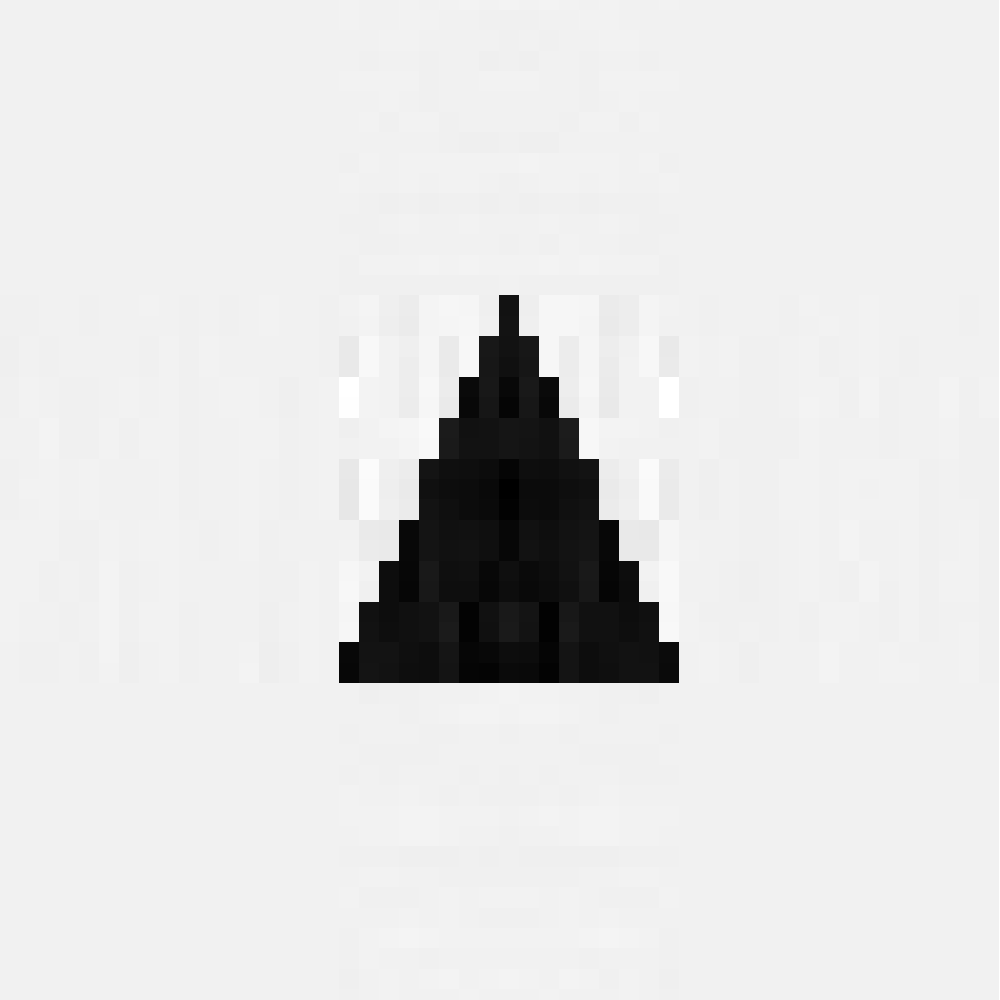}
    &
    \panelimage{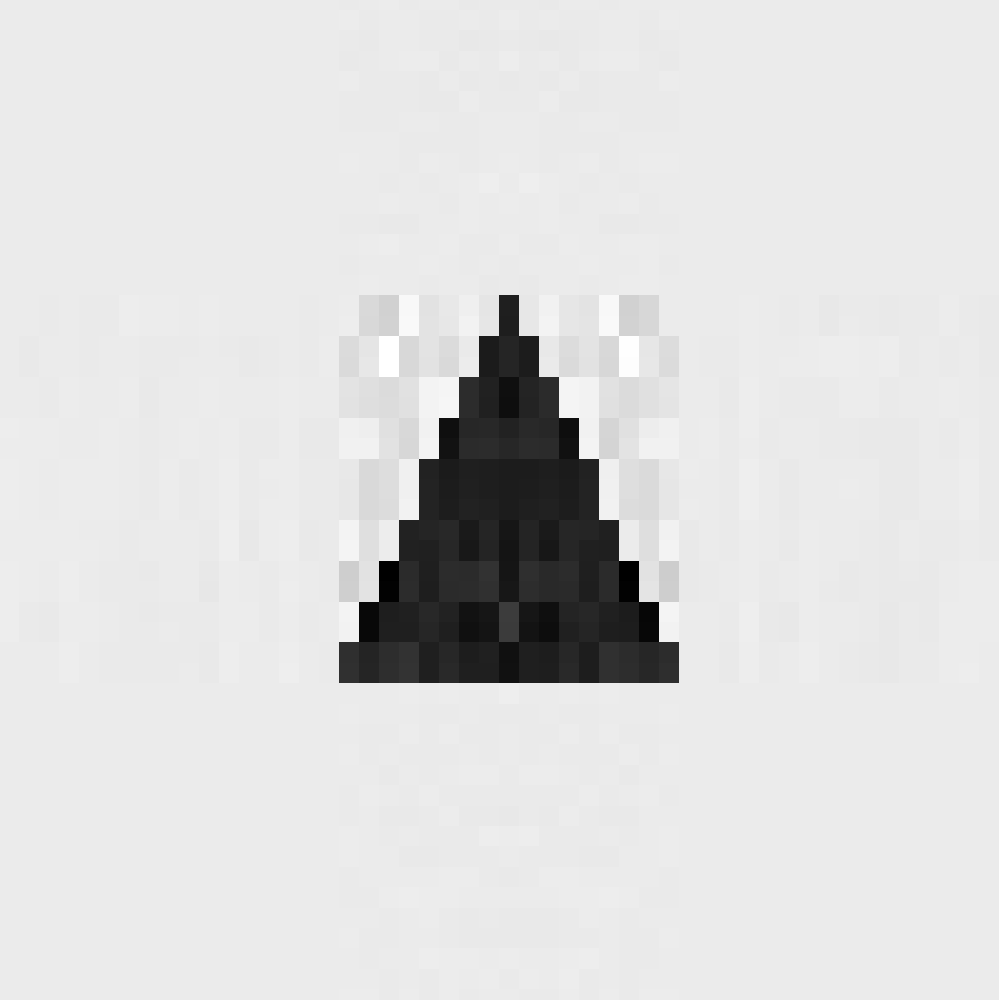}
    &
    \panelimage{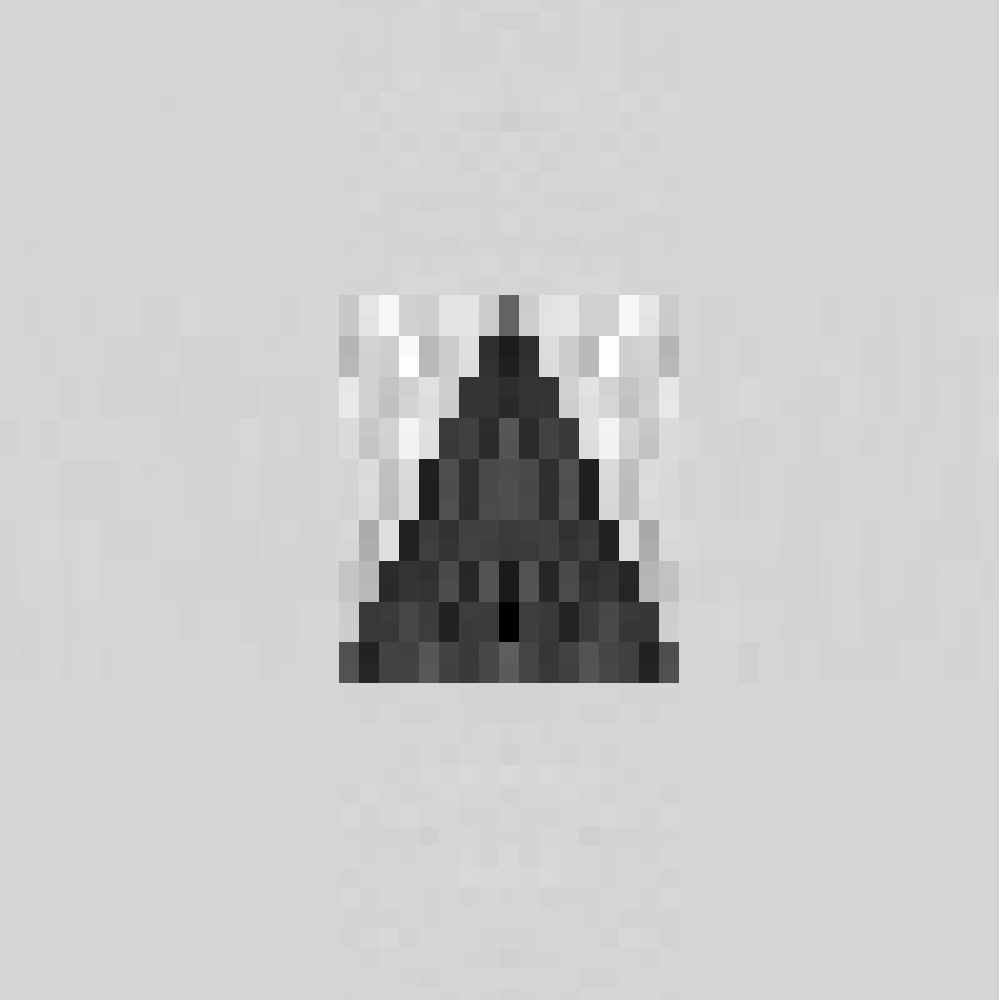}
    \\

    LogN 
    &
    \panelimage{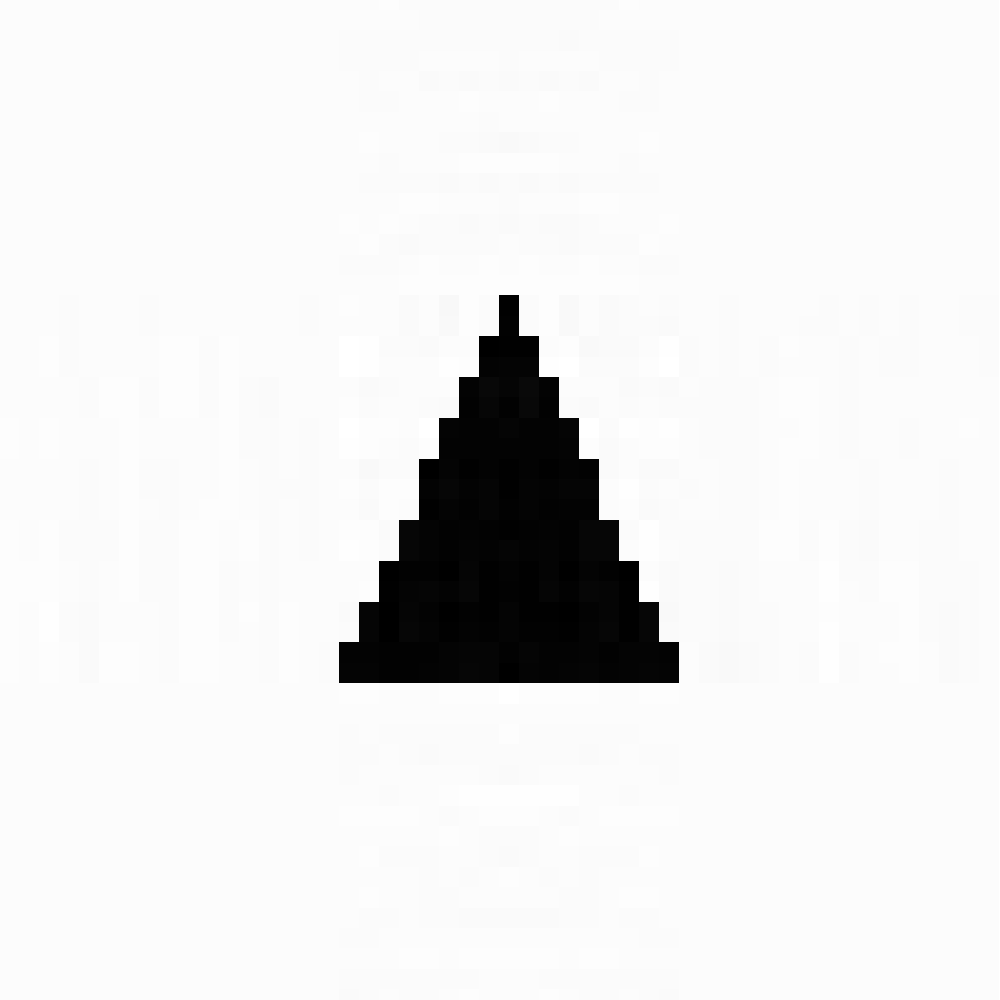}
    &
    \panelimage{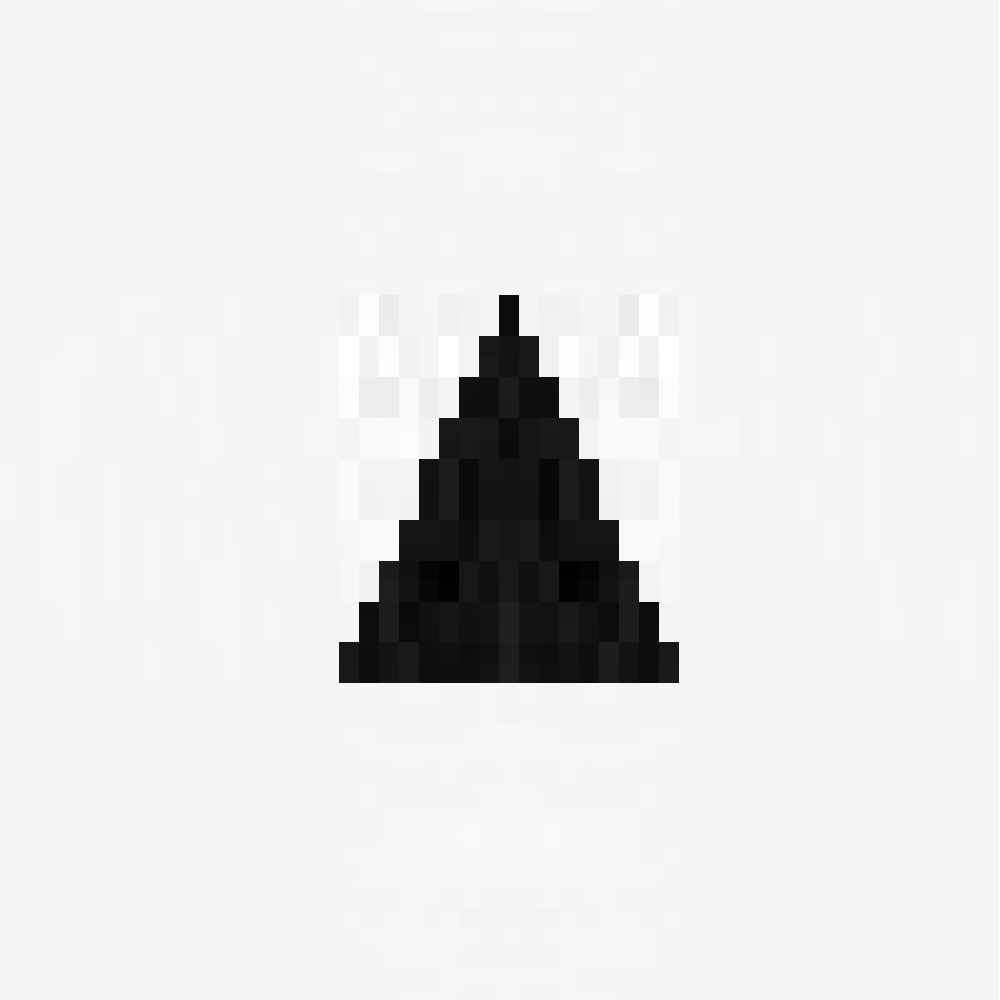}
    &
    \panelimage{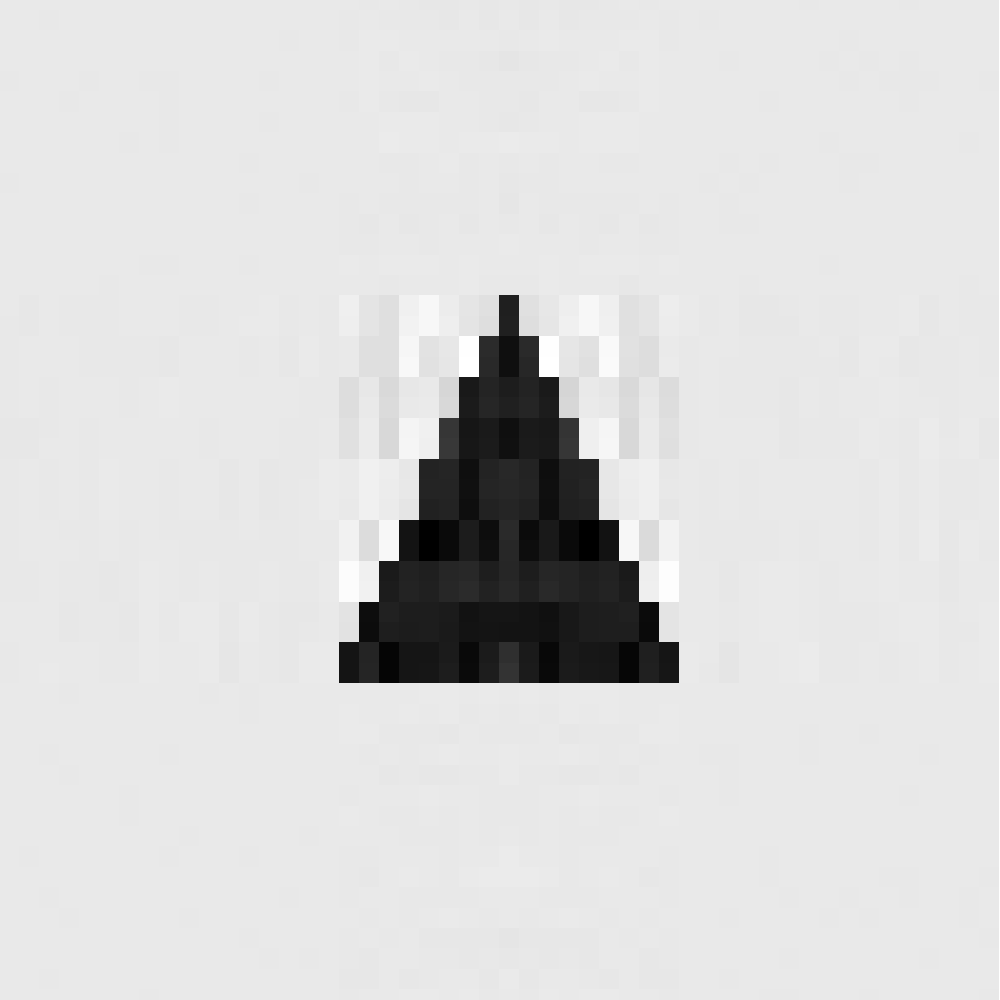}
    &
    \panelimage{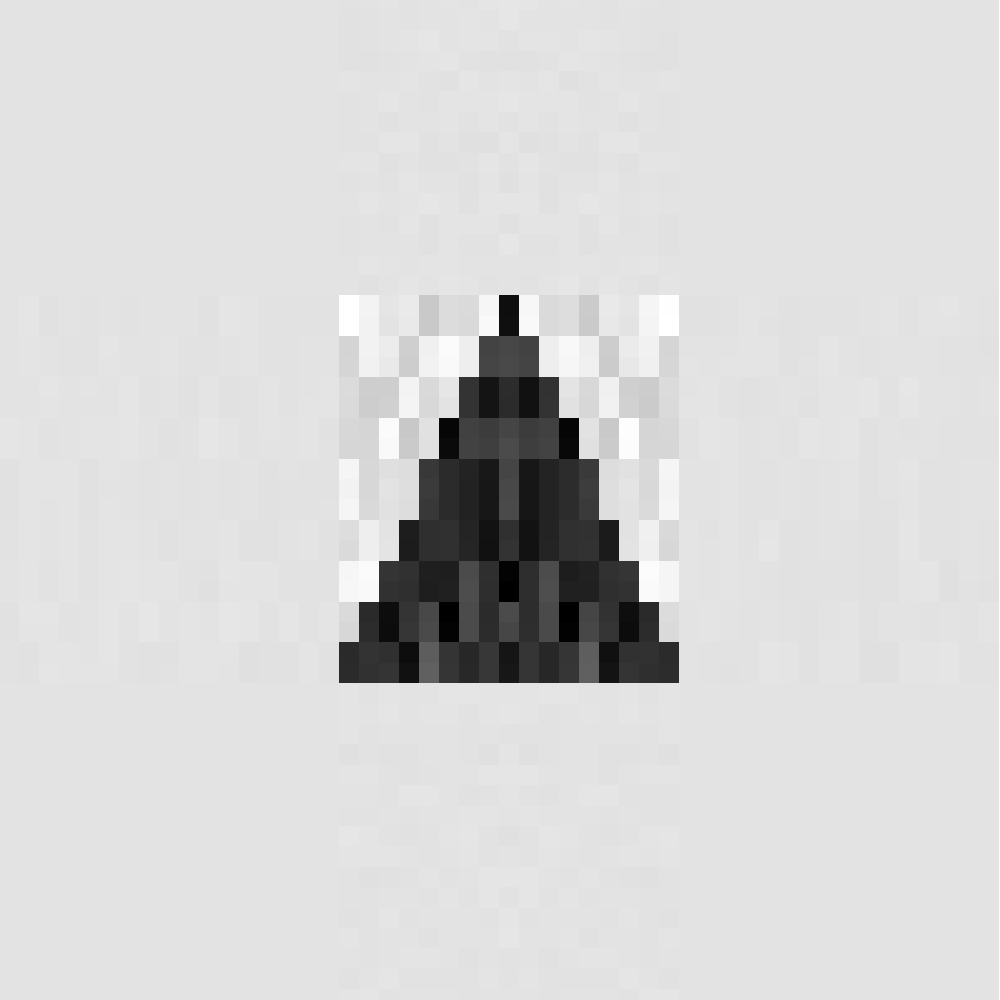}
    \\

    Cont.
    &
    \panelimage{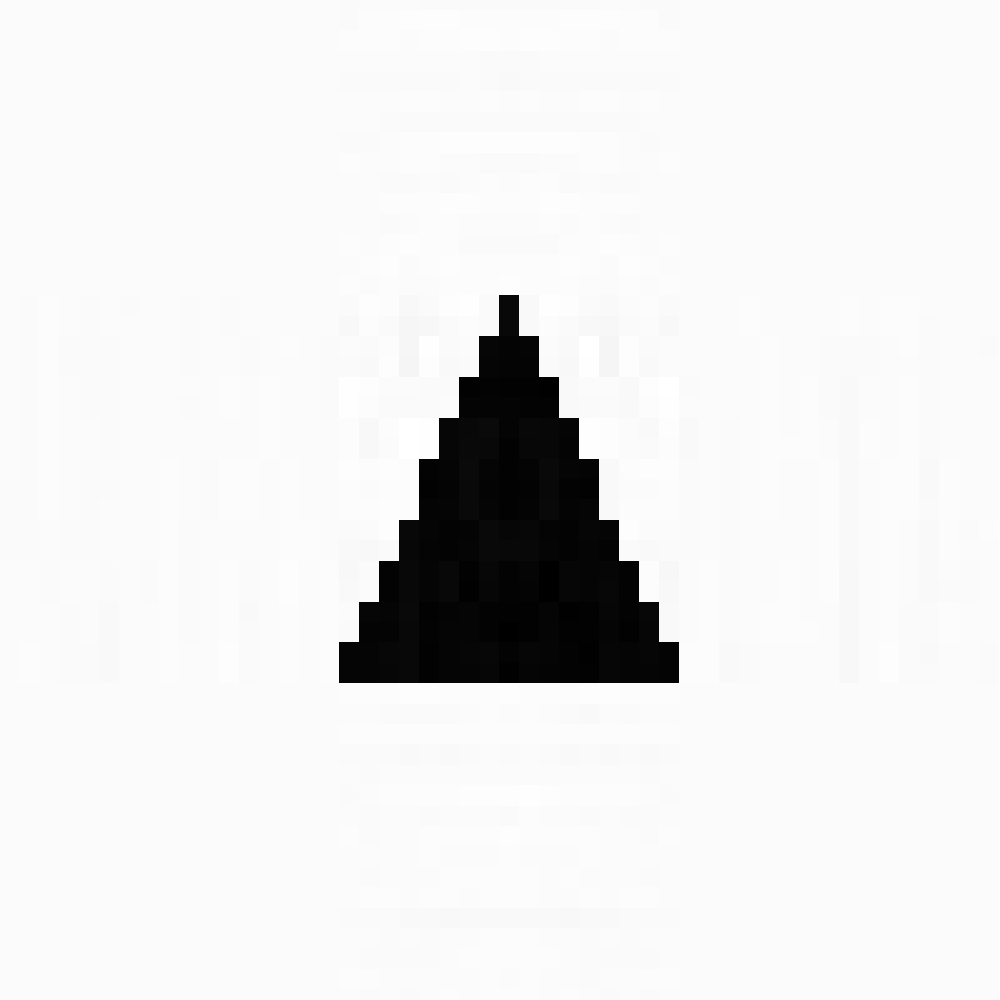}
    &
    \panelimage{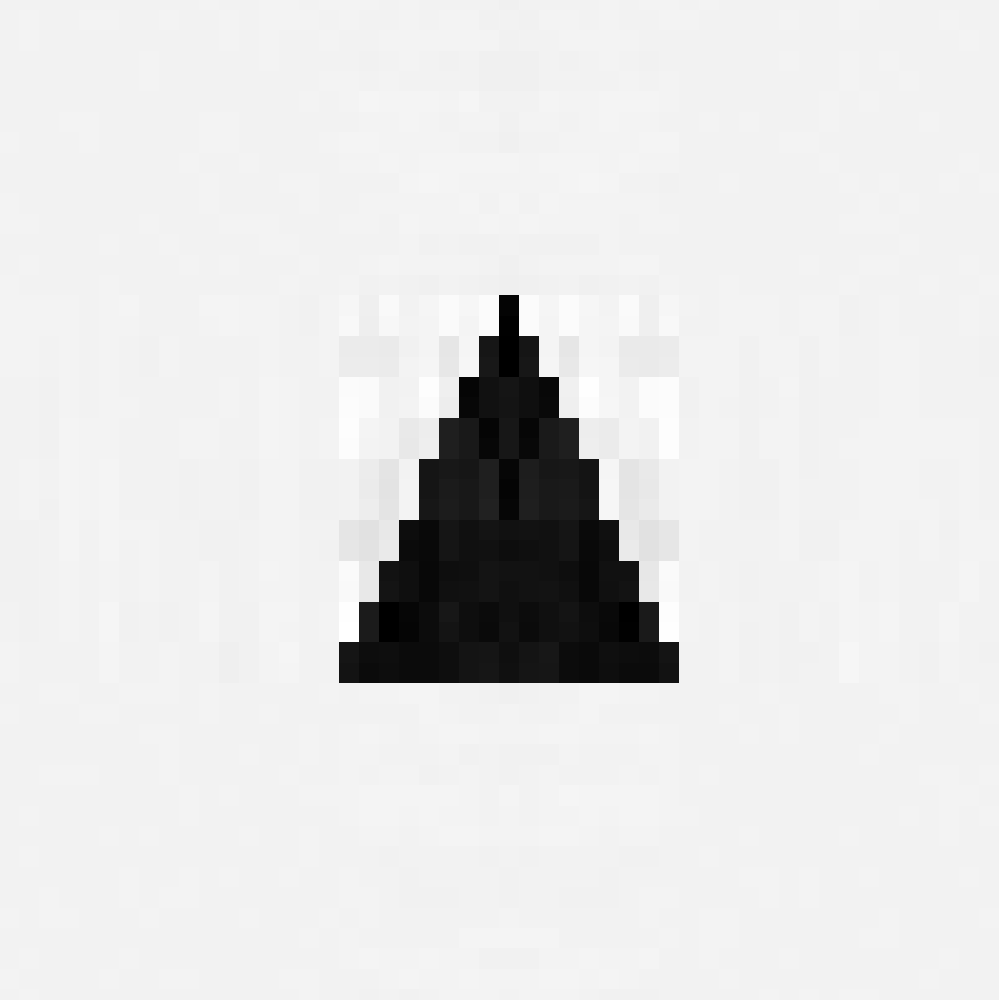}
    &
    \panelimage{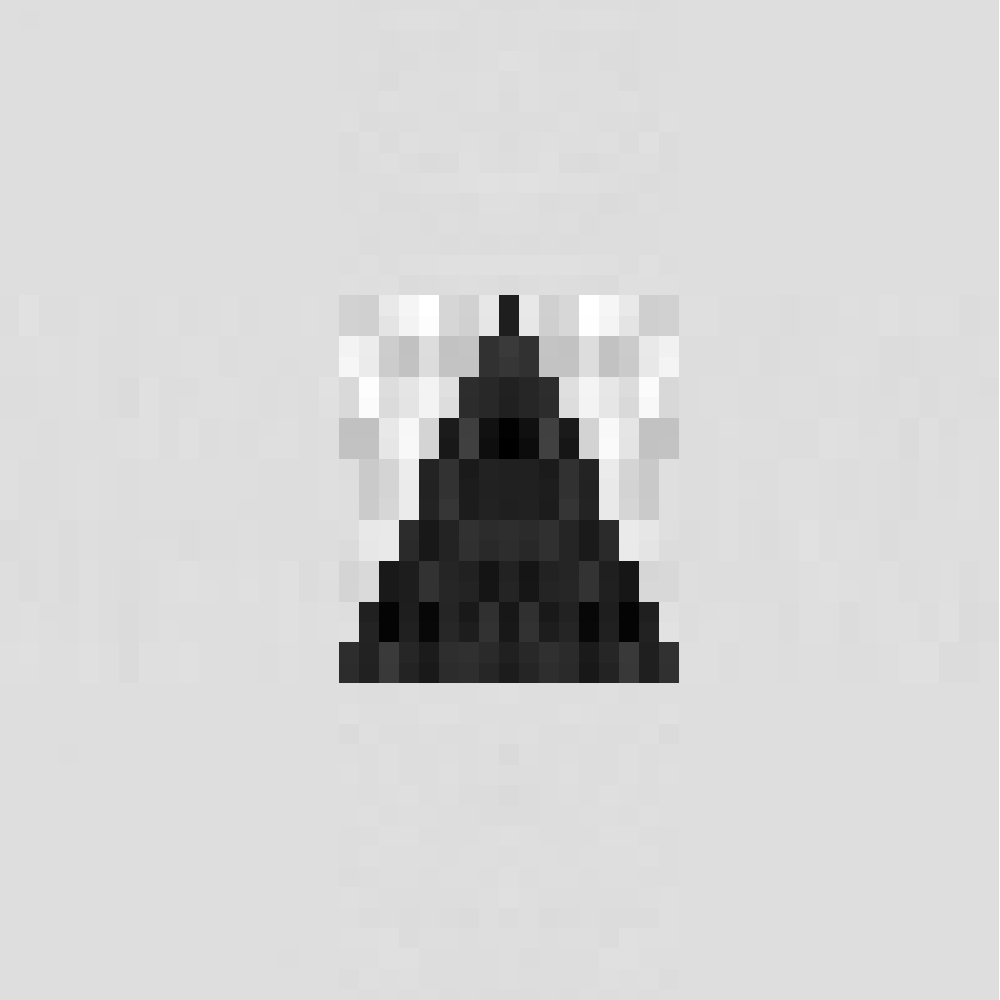}
    &
    \panelimage{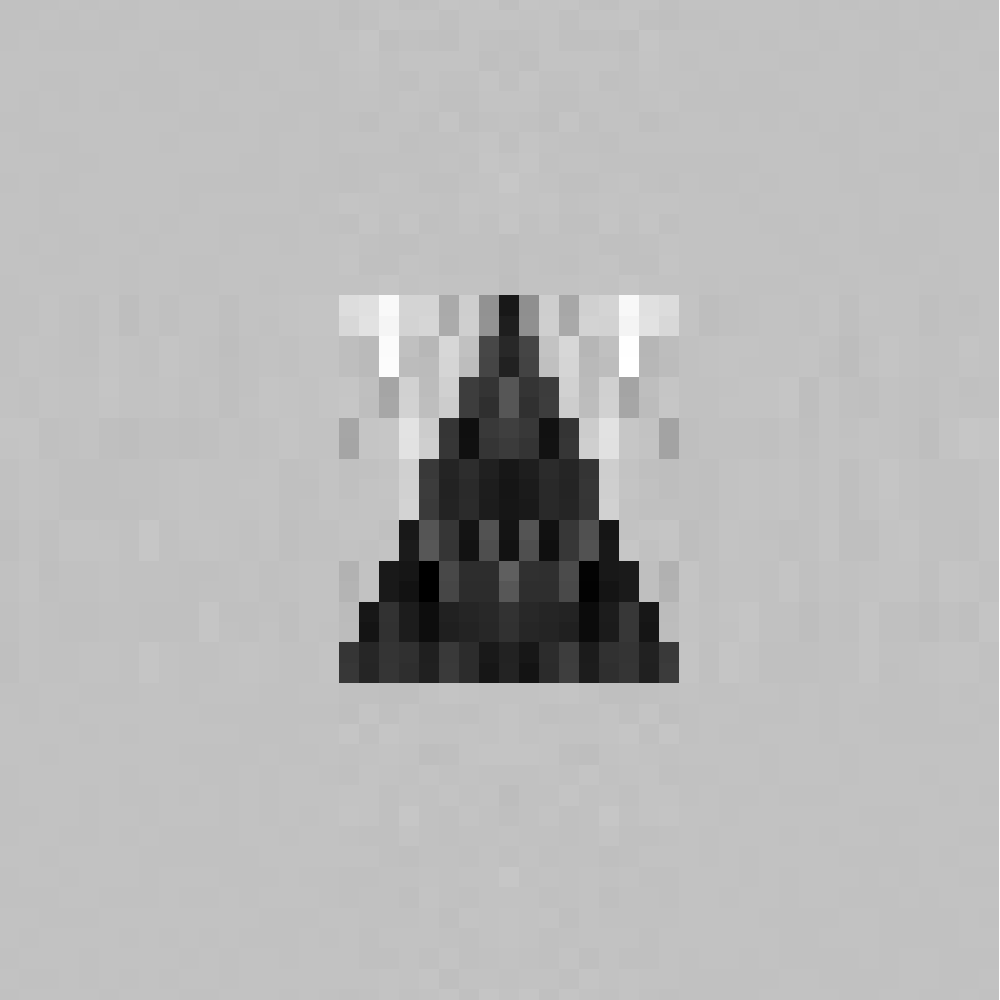}
    \\

    \bottomrule
  \end{tabular}
\caption{
Average estimated coefficient tensors for the triangle-shaped structure
under different error distributions and noise level.
The true coefficient tensor has rank 9 and the sample size is $n=200$.
The results are averaged over 100 Monte Carlo replications.
}
\label{fig3}
\end{figure}

\begin{figure}[htbp]
  \centering
  \includegraphics[width=0.48\textwidth]{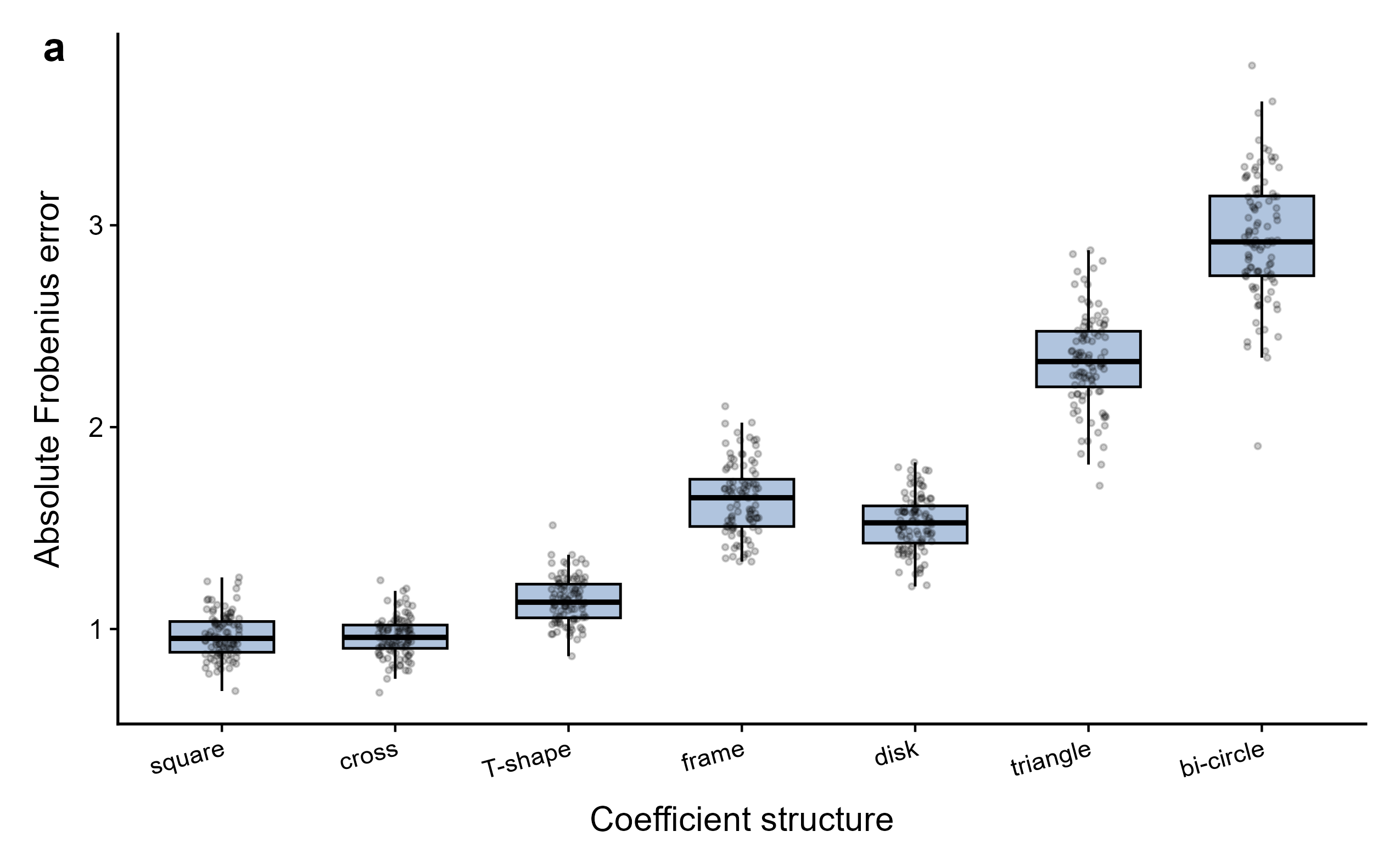}
  \hfill
  \includegraphics[width=0.48\textwidth]{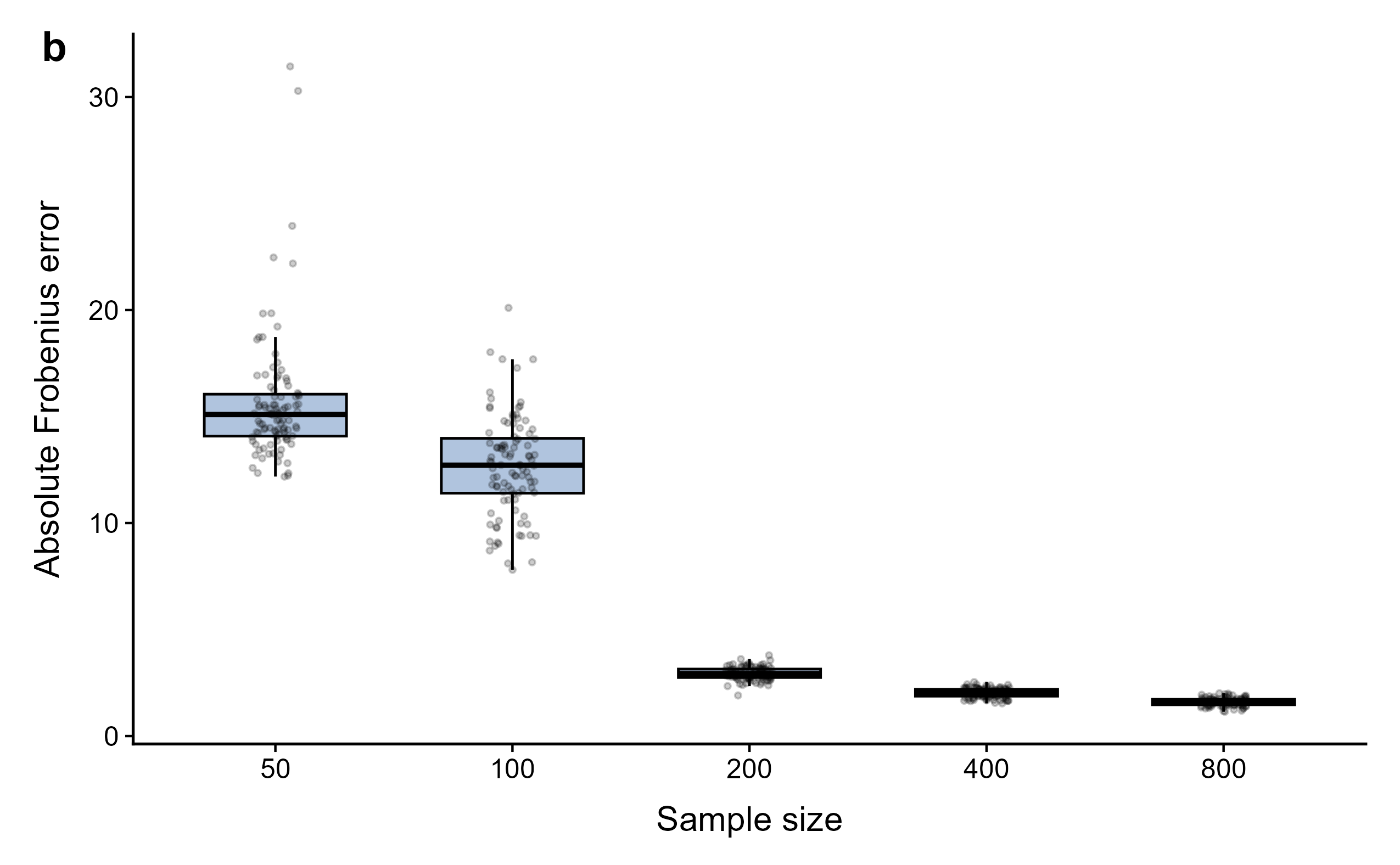}

  \vspace{4mm}

  \includegraphics[width=0.48\textwidth]{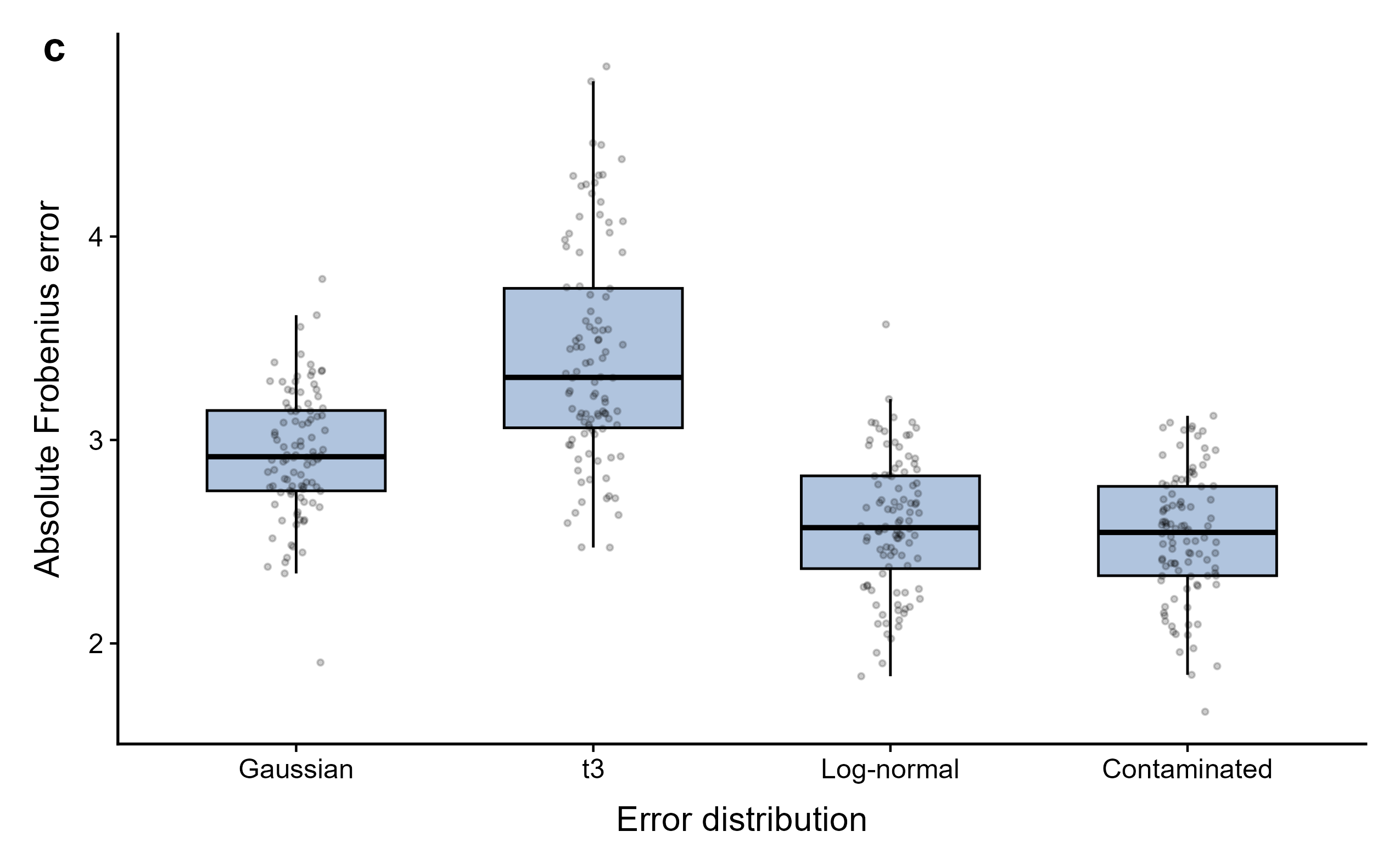}
  \hfill
  \includegraphics[width=0.48\textwidth]{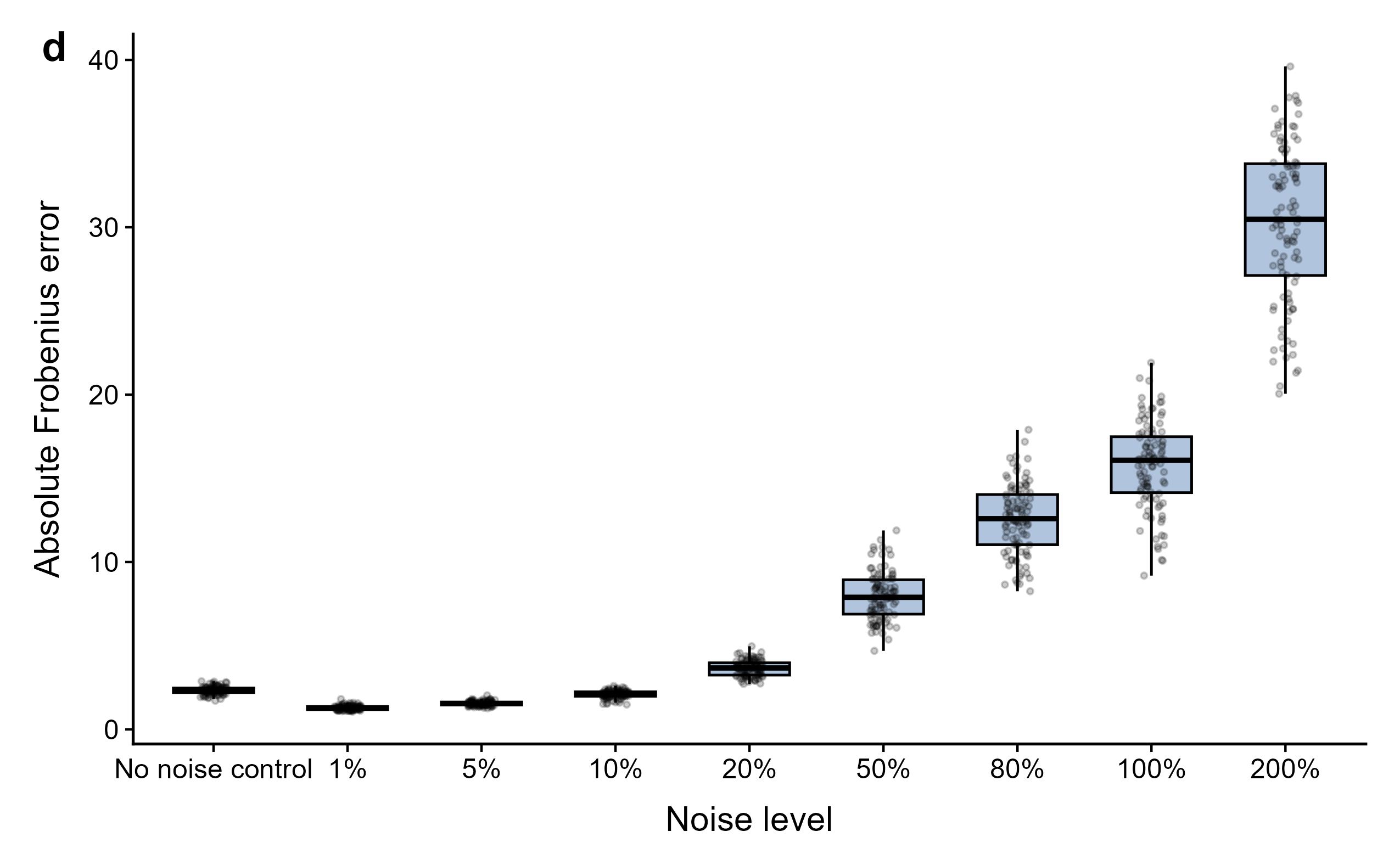}

\caption{
Boxplots of absolute Frobenius estimation errors under different simulation settings.
Panel (a) compares different coefficient structures with sample size $n=200$
and Gaussian errors, without additional noise scaling.
Panel (b) examines the effect of sample size for the bi-circle-shaped coefficient
structure under Gaussian errors, without additional noise scaling.
Panel (c) compares different error distributions with sample size $n=200$
and the bi-circle-shaped coefficient structure, without additional noise scaling.
Panel (d) examines the effect of noise level with sample size $n=200$,
the bi-circle-shaped coefficient structure, and Gaussian errors.
}
  \label{fig4}
\end{figure}

We further investigate the effect of different error distributions on the coefficient estimation performance of PMTR. As shown in Figure \ref{fig1}, for the four lower-rank coefficient structures, PMTR is able to recover the main structural features of the true coefficient tensors under Gaussian, \(t_3\), log-normal, and contaminated normal errors. Similar patterns can be observed for the higher-rank coefficient structures in Figure \ref{fig2}. In particular, the effect of the error distribution appears to be relatively modest compared with that of the sample size. From the visual comparison of the recovered coefficient tensors, no substantial differences are observed among the four error distributions. This observation is further supported by the absolute Frobenius estimation errors presented in Figure \ref{fig4} (c). Specifically, the errors under the \(t_3\) distribution are slightly larger than those under the other three error distributions, possibly reflecting the additional estimation difficulty induced by heavy-tailed errors. Nevertheless, the overall error levels remain comparable across the four distributions. Taken together, the coefficient recovery results and the quantitative comparisons of the Frobenius errors indicate that PMTR exhibits stable estimation performance across the error distributions considered.
To further examine the coefficient recovery performance of PMTR under increasingly severe noise contamination, we gradually increase the noise level and report the corresponding coefficient estimates in Figure \ref{fig3}. As the noise level increases, more spurious signals appear in the estimated coefficient tensors, making accurate coefficient recovery increasingly challenging. Even when the noise level reaches \(200\%\), although a considerable number of spurious signals are present, the main structural features of the true coefficient tensor remain clearly identifiable. The boxplots in Figure \ref{fig4}(d) provide further quantitative evidence for this pattern. The absolute Frobenius estimation errors generally increase with the noise level, indicating a gradual deterioration in estimation accuracy as the noise contamination becomes more severe. Nevertheless, the increase in estimation error remains relatively gradual, and the main structural features of the true coefficient tensor can still be retained under relatively high noise levels. These results suggest that PMTR maintains stable estimation performance across different error distributions and retains the ability to recover the underlying coefficient structure even in the presence of substantial noise contamination.

Overall, the simulation results demonstrate that PMTR exhibits satisfactory finite-sample estimation performance across a wide range of settings. For coefficient tensors with different ranks and structural patterns, PMTR is able to recover the main structural features of the true coefficients, with fewer spurious signals and improved estimation accuracy as the sample size increases. In comparison, the influence of the error distribution is relatively modest, and PMTR exhibits stable estimation performance under Gaussian, \(t_3\), log-normal, and contaminated normal errors. Although higher-rank coefficient structures and increased noise levels make coefficient estimation more challenging and lead to larger Frobenius errors, the estimation of higher-rank structures improves steadily with increasing sample size, while the main structural features of the true coefficient tensor remain identifiable even under substantial noise contamination. Taken together, these findings provide both visual and quantitative evidence for the effectiveness and stability of PMTR in finite-sample settings.

\section{An Application to the Beijing Air-Quality Data}
We apply the proposed PMTR method to data from the Wanliu monitoring station in the Beijing Multi-Site Air Quality Data set\citet{beijing_multi-site_air_quality_501}, which is publicly available from the UCI Machine Learning Repository\footnote{\url{(https://archive.ics.uci.edu/dataset/501/beijing+multi+site+air+quality+data)}}. The Wanliu data cover the period from March 2013 to February 2017 and contain 35,064 hourly observations. In addition to hourly PM\(_{2.5}\) concentrations, the data record a variety of air pollutants and meteorological variables. These measurements can therefore be naturally organized as a matrix-valued predictor with an hour-by-variable structure.
Our analysis focuses on the daily average PM\(_{2.5}\) concentration, with each selected day treated as one observational unit. For the \(i\)th selected day, we construct a matrix-valued predictor
$
X_i\in\mathbb R^{24\times 8},
$where the 24 rows correspond to hourly measurements from 00:00 to 23:00, and the eight columns correspond to
$
\mathrm{SO_2},\ \mathrm{NO_2},\ \mathrm{CO},\ \mathrm{O_3},
\ \mathrm{TEMP},\ \mathrm{PRES},\ \mathrm{DEWP},\ \mathrm{WSPM},
$representing sulfur dioxide, nitrogen dioxide, carbon monoxide, ozone, temperature, atmospheric pressure, dew point temperature, and wind speed, respectively. The response variable is defined as the daily average PM\(_{2.5}\) concentration,
$
Y_i=\frac{1}{24}\sum_{h=1}^{24}\mathrm{PM}_{2.5,ih}.
$ To avoid additional uncertainty introduced by missing-value imputation, we retain only those days for which all 24 hourly measurements of the predictor variables and PM\(_{2.5}\) are completely observed. To further mitigate potential serial dependence between consecutive daily observations, the entire study period is partitioned into consecutive seven-day blocks, and one complete day is randomly selected from each block containing at least one fully observed day. This procedure yields a final sample of \(n=192\) daily observations. Since the pollutant and meteorological variables are measured on substantially different scales, each predictor variable is standardized prior to model fitting.
To account for potential annual seasonality in PM\(_{2.5}\) concentrations, we further include two periodic covariates based on the day of year (DOY). Let \(d_i\) denote the day-of-year index corresponding to the \(i\)th selected date, and define
$
Z_i=
\begin{pmatrix}
\sin\left(2\pi d_i/365.25\right), 
\cos\left(2\pi d_i/365.25\right)
\end{pmatrix} ^ \top.
$Compared with treating calendar month as a continuous covariate, this harmonic representation provides a more natural characterization of annual seasonal variation while preserving continuity between the end and the beginning of a year. Based on the above construction, we model the conditional mode of the daily average PM\(_{2.5}\) concentration as
$
\operatorname{Modal}(Y_i\mid X_i,Z_i)
=
\alpha+\langle B,X_i\rangle+\gamma^\top Z_i,
$where \(B\in\mathbb R^{24\times8}\) is the coefficient matrix of primary interest and \(\gamma\in\mathbb R^2\) captures the annual seasonal effect.

\begin{figure}[ht]
	\centering
	\includegraphics[width=0.5 \linewidth]{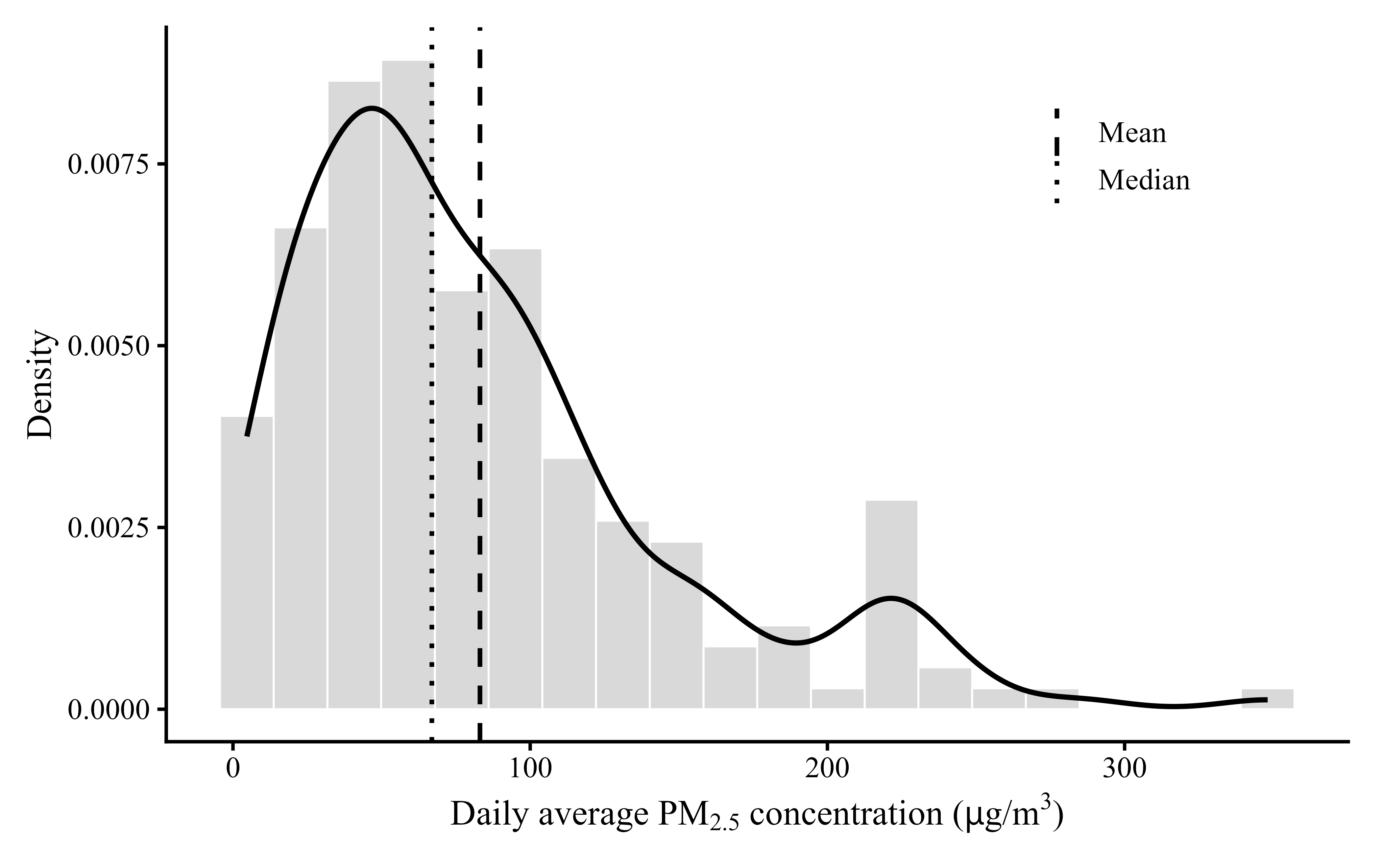}
    \caption{Empirical distribution of the daily average PM$_{2.5}$ concentrations.}
	\label{fig5}
\end{figure}

Figure \ref{fig5} presents the empirical distribution of the daily average PM\(_{2.5}\) concentrations in the selected sample. The distribution is noticeably right-skewed, with most observations concentrated at relatively low concentration levels and a pronounced upper tail containing several relatively large observations. In addition, the sample mean is clearly larger than the median, further indicating substantial distributional asymmetry. These features suggest a clear departure from a symmetric Gaussian distribution and provide practical motivation for modeling the conditional mode rather than focusing solely on the conditional mean.

\begin{figure}[ht]
	\centering
	\includegraphics[width=\linewidth]{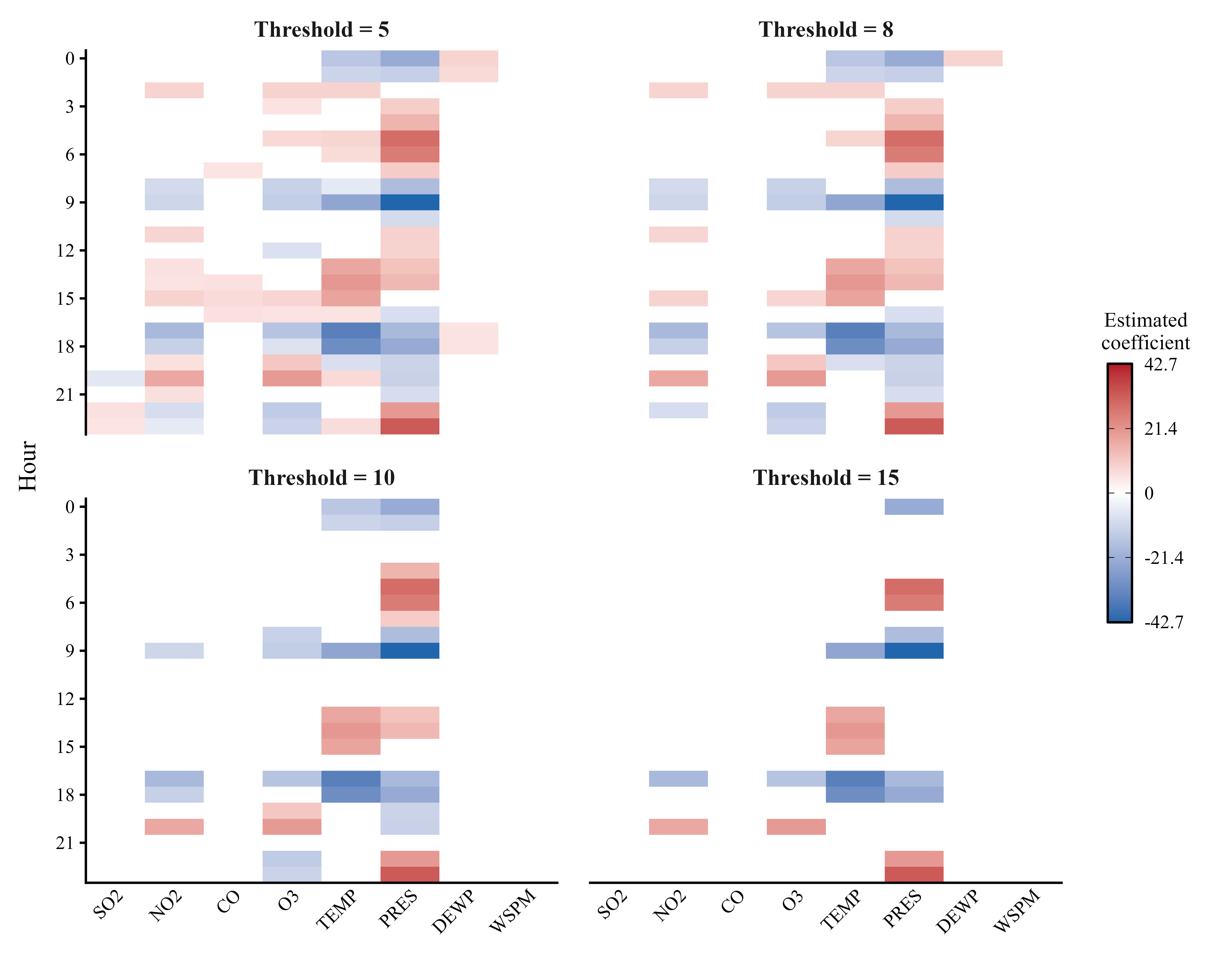}
	\caption{Heat-map visualization of the estimated PMTR coefficient matrix thresholded at the cutoff values 5, 8, 10, and 15. Coefficients with absolute values smaller than the corresponding cutoff are set to zero. The columns correspond to SO\(_2\), NO\(_2\), CO, O\(_3\), TEMP, PRES, DEWP, and WSPM, and the rows correspond to the 24 hours of a day.}
	\label{fig6}
\end{figure}

Using the RER criterion for dimension selection, the estimated dimensions of the two tensor modes are
$
\hat d_1=8,  \hat d_2=8,
$
reducing the original \(24\times8\) matrix-valued predictor to an effective \(8\times8\) representation. This result indicates a substantial reduction along the hourly mode, whereas the full dimension of the environmental-variable mode is retained. Based on the selected dimensions, the PMTR model yields an estimated coefficient matrix
$
\widehat B\in\mathbb R^{24\times8}.
$
The minimum, first quartile, median, mean, third quartile, and maximum of the estimated coefficients are \(-42.7038\), \(-1.6636\), \(1.3391\), \(0.6397\), \(4.6697\), and \(32.6757\), respectively. Most of the estimated coefficients are concentrated around relatively small values, while a limited number exhibit substantially larger magnitudes. To better highlight the dominant hour-by-variable association patterns, we first visualize the estimated coefficient matrix using a threshold of 5, setting \(\hat B_{jk}=0\) whenever \(|\hat B_{jk}|<5\). Thresholds of 8, 10, and 15 are further considered for comparison to assess the stability of the main coefficient patterns with respect to the threshold choice.

As shown in Figure \ref{fig6}, the relatively large and stable coefficients across different threshold levels are mainly concentrated on \(\mathrm{NO_2}\), \(\mathrm{O_3}\), \(\mathrm{TEMP}\), and \(\mathrm{PRES}\). More specifically, \(\mathrm{NO_2}\) exhibits a pronounced positive coefficient around 20:00, whereas negative coefficients are mainly observed around 17:00--18:00. A broadly similar temporal pattern is observed for \(\mathrm{O_3}\). For \(\mathrm{TEMP}\), relatively large negative coefficients appear around 09:00 and 18:00, while positive coefficients are mainly observed during 13:00--15:00. The coefficients associated with \(\mathrm{PRES}\) display a more pronounced time-varying pattern, being predominantly positive during 04:00--08:00, negative during 09:00--11:00 and around 14:00, and positive again around 23:00. Overall, these patterns remain relatively stable across the different threshold choices, suggesting that \(\mathrm{NO_2}\), \(\mathrm{O_3}\), \(\mathrm{TEMP}\), and \(\mathrm{PRES}\) are more strongly associated with the conditional modal of the daily average PM\(_{2.5}\) concentration, with the strength and direction of the associations varying over the course of the day.

Overall, the analysis of the Beijing air-quality data demonstrates that PMTR can effectively extract low-dimensional information associated with the conditional modal of daily average PM\(_{2.5}\) concentrations while preserving the intrinsic matrix structure of the predictors. The dimension selected by the RER criterion suggests that the response-relevant information is primarily concentrated in a lower-dimensional temporal subspace, while the thresholded coefficient analysis further reveals several stable hour-by-variable association patterns, with \(\mathrm{NO_2}\), \(\mathrm{O_3}\), \(\mathrm{TEMP}\), and \(\mathrm{PRES}\) exhibiting the most pronounced effects. Moreover, the direction and magnitude of these associations vary substantially over the course of the day, highlighting the importance of retaining the hour-by-variable structure in modeling complex air-quality data. Together with the noticeable right skewness and pronounced upper tail observed in the distribution of daily average PM\(_{2.5}\), these findings illustrate the practical usefulness of PMTR for analyzing structured predictors under asymmetric response distributions and for providing interpretable modal association patterns.

\section{Conclusion}
In this paper, we proposed a partial modal tensor regression (PMTR) framework for characterizing the conditional modal relationship between a continuous response and high-dimensional tensor predictors. By exploiting the intrinsic multiway structure of tensor-valued covariates, PMTR performs mode-wise dimension reduction to extract response-relevant low-dimensional information, thereby avoiding the dimensional inflation and structural information loss caused by direct vectorization. Unlike tensor regression approaches targeting the conditional mean or conditional quantiles, PMTR directly models the conditional modal of the response and thus provides an interpretable alternative for data with asymmetric, heavy-tailed, or potentially contaminated response distributions.

From a theoretical perspective, we established the consistency and asymptotic properties of the proposed estimators. From a computational perspective, an estimation procedure was developed to integrate tensor dimension reduction with modal regression within a unified framework. 
Extensive simulation studies under various sample sizes, coefficient structures, error distributions, and noise levels demonstrate that PMTR can stably recover the underlying coefficient structure, with estimation accuracy improving as the sample size increases while maintaining satisfactory performance across different distributional and noise settings. The analysis of the Beijing air-quality data further illustrates that PMTR can achieve meaningful tensor dimension reduction and identify interpretable environmental association patterns that vary over the course of the day, demonstrating its practical usefulness for analyzing structured high-dimensional predictors through the conditional modal.

The present study focuses primarily on the PMTR model under independent observations, leaving several directions for future investigation. In settings with higher-dimensional tensor predictors or pronounced sparsity in the underlying coefficient tensor, combining mode-wise dimension reduction with sparse regularization may further improve estimation efficiency and model interpretability. In addition, more systematic uncertainty quantification and statistical inference for the estimated coefficient tensor and the corresponding dimension-reduction subspaces constitute another important direction for future research.

\newpage
\bibliographystyle{plainnat}
\bibliography{references}

\begin{thebibliography}{18}
\providecommand{\natexlab}[1]{#1}
\providecommand{\url}[1]{\texttt{#1}}
\expandafter\ifx\csname urlstyle\endcsname\relax
  \providecommand{\doi}[1]{doi: #1}\else
  \providecommand{\doi}{doi: \begingroup \urlstyle{rm}\Url}\fi

\bibitem[Auddy et~al.(2025)Auddy, Xia, and Yuan]{auddy2025tensors}
Arnab Auddy, Dong Xia, and Ming Yuan.
\newblock Tensors in high-dimensional data analysis: Methodological
  opportunities and theoretical challenges.
\newblock \emph{Annual review of statistics and its application}, 12\penalty0
  (1):\penalty0 527--551, 2025.

\bibitem[Chen(2017)]{beijing_multi-site_air_quality_501}
Song Chen.
\newblock {Beijing Multi-Site Air Quality}.
\newblock UCI Machine Learning Repository, 2017.
\newblock {DOI}: https://doi.org/10.24432/C5RK5G.

\bibitem[Chen et~al.(2016)Chen, Genovese, Tibshirani, and
  Wasserman]{chen2016nonparametric}
Yen-Chi Chen, Christopher~R Genovese, Ryan~J Tibshirani, and Larry Wasserman.
\newblock Nonparametric modal regression.
\newblock \emph{The Annals of Statistics}, 44\penalty0 (2):\penalty0 489--514,
  2016.

\bibitem[Hoff(2011)]{hoff2011separable}
Peter~D. Hoff.
\newblock Separable covariance arrays via the tucker product, with applications
  to multivariate relational data.
\newblock \emph{Bayesian Analysis}, 6\penalty0 (2):\penalty0 179--196, 2011.

\bibitem[Kemp and Silva(2012)]{kemp2012regression}
Gordon~CR Kemp and JMC~Santos Silva.
\newblock Regression towards the mode.
\newblock \emph{Journal of Econometrics}, 170\penalty0 (1):\penalty0 92--101,
  2012.

\bibitem[Kolda and Bader(2009)]{kolda2009tensor}
Tamara~G Kolda and Brett~W Bader.
\newblock Tensor decompositions and applications.
\newblock \emph{SIAM review}, 51\penalty0 (3):\penalty0 455--500, 2009.

\bibitem[Lee(1989)]{lee1989mode}
Myoung-jae Lee.
\newblock Mode regression.
\newblock \emph{Journal of Econometrics}, 42\penalty0 (3):\penalty0 337--349,
  1989.

\bibitem[Lock(2018)]{lock2018tensor}
Eric~F Lock.
\newblock Tensor-on-tensor regression.
\newblock \emph{Journal of Computational and Graphical Statistics}, 27\penalty0
  (3):\penalty0 638--647, 2018.

\bibitem[Sun et~al.(2025)Sun, Peng, Qiu, Guo, and Manatunga]{Sunetal.2025pqtr}
Dayu Sun, Limin Peng, Zhiping Qiu, Ying Guo, and Amita Manatunga.
\newblock {Partial Quantile Tensor Regression}.
\newblock \emph{Journal of the American Statistical Association}, 120\penalty0
  (515):\penalty0 1724--1735, 2025.

\bibitem[Tsiligkaridis and Hero(2013)]{TsiligkaridisandHero2013}
Theodoros Tsiligkaridis and Alfred~O. Hero.
\newblock Covariance estimation in high dimensions via kronecker product
  expansions.
\newblock \emph{IEEE Transactions on Signal Processing}, 61\penalty0
  (21):\penalty0 5347--5360, 2013.

\bibitem[Wang et~al.(2017)Wang, Chen, Cai, Shen, and
  Huang]{wang2017regularized}
Xiaoqian Wang, Hong Chen, Weidong Cai, Dinggang Shen, and Heng Huang.
\newblock Regularized modal regression with applications in cognitive
  impairment prediction.
\newblock \emph{Advances in neural information processing systems}, 30, 2017.

\bibitem[Xia et~al.(2015)Xia, Xu, and Zhu]{xia2015consistently}
Qiang Xia, Wangli Xu, and Lixing Zhu.
\newblock Consistently determining the number of factors in multivariate
  volatility modelling.
\newblock \emph{Statistica Sinica}, pages 1025--1044, 2015.

\bibitem[Yao and Li(2014)]{yaoandli2014modal}
Weixin Yao and Longhai Li.
\newblock A new regression model: modal linear regression.
\newblock \emph{Scandinavian Journal of Statistics}, 41\penalty0 (3):\penalty0
  656--671, 2014.

\bibitem[Yu et~al.(2015)Yu, Wang, and Samworth]{yu2015useful}
Y.~Yu, T.~Wang, and R.~J. Samworth.
\newblock A useful variant of the davis--kahan theorem for statisticians.
\newblock \emph{Biometrika}, 102\penalty0 (2):\penalty0 315--323, 2015.

\bibitem[Zhang and Li(2017)]{ZhangandLi2017TEPLS}
Xin Zhang and Lexin Li.
\newblock Tensor envelope partial least-squares regression.
\newblock \emph{Technometrics}, 59\penalty0 (4):\penalty0 426--436, 2017.

\bibitem[Zhong et~al.(2015)Zhong, Xing, and Suslick]{zhong2015tensor}
Wenxuan Zhong, Xin Xing, and Kenneth Suslick.
\newblock Tensor sufficient dimension reduction.
\newblock \emph{Wiley Interdisciplinary Reviews: Computational Statistics},
  7\penalty0 (3):\penalty0 178--184, 2015.

\bibitem[Zhou et~al.(2013)Zhou, Li, and Zhu]{zhou2013tensor}
Hua Zhou, Lexin Li, and Hongtu Zhu.
\newblock Tensor regression with applications in neuroimaging data analysis.
\newblock \emph{Journal of the American Statistical Association}, 108\penalty0
  (502):\penalty0 540--552, 2013.

\bibitem[Zhu et~al.(2017)Zhu, Guo, and Zhu]{zhu2017adaptive}
Xuehu Zhu, Xu~Guo, and Lixing Zhu.
\newblock An adaptive-to-model test for partially parametric single-index
  models.
\newblock \emph{Statistics and Computing}, 27\penalty0 (5):\penalty0
  1193--1204, 2017.

\end{thebibliography}

\newpage
\appendix
% 附录部分
\section{ Proofs of Lemmas and Theorems }

\subsection{Proof of Lemma 1}

\begin{proof}
For fixed
\((\mathcal{X},\boldsymbol{Z})\), define
\[
H(s)
=
\mathbb{E}\!\left[
\mathrm{R}(Y-s)
\mid
\mathcal{X},\boldsymbol{Z}
\right].
\]
Differentiation with respect to \(s\) gives
\[
H'(s)
=
-\mathbb{E}\!\left[
\mathrm{R}'(Y-s)
\mid
\mathcal{X},\boldsymbol{Z}
\right].
\]
Therefore, by the fundamental theorem of calculus and the change of
variables \(s=a+u(b-a)\),
\[
\begin{aligned}
H(a)-H(b)
&=
\int_b^a H'(s)\,\mathrm{d}s \\
&=
(b-a)
\int_0^1
\mathbb{E}\!\left[
\mathrm{R}'\!\left\{
Y-a-u(b-a)
\right\}
\mid
\mathcal{X},\boldsymbol{Z}
\right]
\,\mathrm{d}u \\
&=
-v(\boldsymbol{Z},\mathcal{X})(b-a).
\end{aligned}
\]
Because \(Y-b=\varepsilon\),
\[
H(b)
=
\mathbb{E}\!\left[
\mathrm{R}(\varepsilon)
\mid
\mathcal{X},\boldsymbol{Z}
\right]
=
d_h(\mathcal{X},\boldsymbol{Z}).
\]
It follows that
\[
\begin{aligned}
H(a)
=d_h(\mathcal{X},\boldsymbol{Z}) -
v(\boldsymbol{Z},\mathcal{X})
\left[
\left(
\widetilde{\boldsymbol{\gamma}}_0
-
\widetilde{\boldsymbol{\gamma}}_{Y\mid Z}^{0}
\right)^\top
\widetilde{\boldsymbol{Z}}
+
\operatorname{vec}(\mathcal{X})^\top\beta_0
\right].
\end{aligned}
\]

Since
\(\widetilde{\boldsymbol{\gamma}}_{Y\mid Z}^{0}\)
is an interior maximizer of the population marginal objective, its
first-order condition is
\[
\mathbb{E}\!\left[
\widetilde{\boldsymbol{Z}}\,
\mathrm{R}\!\left\{
Y-
\left(
\widetilde{\boldsymbol{\gamma}}_{Y\mid Z}^{0}
\right)^\top
\widetilde{\boldsymbol{Z}}
\right\}
\right]
=\boldsymbol{0}.
\]
Multiplying the preceding expression for \(H(a)\) by
\(\widetilde{\boldsymbol{Z}}\), taking expectations, and applying this
first-order condition yield
\[
\begin{aligned}
\mathbb{E}\!\left[
v(\boldsymbol{Z},\mathcal{X})
\widetilde{\boldsymbol{Z}}
\widetilde{\boldsymbol{Z}}^\top
\right]
\left(
\widetilde{\boldsymbol{\gamma}}_0
-
\widetilde{\boldsymbol{\gamma}}_{Y\mid Z}^{0}
\right) 
+
\mathbb{E}\!\left[
v(\boldsymbol{Z},\mathcal{X})
\widetilde{\boldsymbol{Z}}
\operatorname{vec}(\mathcal{X})^\top
\right]\beta_0 
=
\mathbb{E}\!\left[
\widetilde{\boldsymbol{Z}}\,
d_h(\mathcal{X},\boldsymbol{Z})
\right].
\end{aligned}
\]
Hence,
\[
\begin{aligned}
\widetilde{\boldsymbol{\gamma}}_0
-
\widetilde{\boldsymbol{\gamma}}_{Y\mid Z}^{0}
=
\left\{
\mathbb{E}\!\left[
v(\boldsymbol{Z},\mathcal{X})
\widetilde{\boldsymbol{Z}}
\widetilde{\boldsymbol{Z}}^\top
\right]
\right\}^{-1}
\mathbb{E}\!\left[
\widetilde{\boldsymbol{Z}}\,
d_h(\mathcal{X},\boldsymbol{Z})
\right] -
A\beta_0.
\end{aligned}
\]

By the definition of the partial modal tensor covariance,
\[
\operatorname{vec}\!\left\{
\operatorname{pmtcov}
(Y,\mathcal{X}\mid\boldsymbol{Z})
\right\}
=
\mathbb{E}\!\left[
\operatorname{vec}(\mathcal{X})H(a)
\right].
\]
Substituting the preceding expression for
\(\widetilde{\boldsymbol{\gamma}}_0
-
\widetilde{\boldsymbol{\gamma}}_{Y\mid Z}^{0}\)
into this identity gives
\[
\begin{aligned}
\operatorname{vec}\!\left\{
\operatorname{pmtcov}
(Y,\mathcal{X}\mid\boldsymbol{Z})
\right\} 
&=
\mathbb{E}\!\left[
\operatorname{vec}(\mathcal{X})
d_h(\mathcal{X},\boldsymbol{Z})
\right] \\
&\quad-
\mathbb{E}\!\left[
v(\boldsymbol{Z},\mathcal{X})
\operatorname{vec}(\mathcal{X})
\widetilde{\boldsymbol{Z}}^\top
\right]
\left\{
\mathbb{E}\!\left[
v(\boldsymbol{Z},\mathcal{X})
\widetilde{\boldsymbol{Z}}
\widetilde{\boldsymbol{Z}}^\top
\right]
\right\}^{-1} 
\mathbb{E}\!\left[
\widetilde{\boldsymbol{Z}}
d_h(\mathcal{X},\boldsymbol{Z})
\right] \\
&\quad+
\mathbb{E}\!\left[
v(\boldsymbol{Z},\mathcal{X})
\operatorname{vec}(\mathcal{X})
\widetilde{\boldsymbol{Z}}^\top
\right]A\beta_0 
-
\mathbb{E}\!\left[
v(\boldsymbol{Z},\mathcal{X})
\operatorname{vec}(\mathcal{X})
\operatorname{vec}(\mathcal{X})^\top
\right]\beta_0.
\end{aligned}
\]

By the definition of \(A^\top\), the first two terms involving
\(d_h(\mathcal{X},\boldsymbol{Z})\) satisfy
\[
\begin{aligned}
&\mathbb{E}\!\left[
\operatorname{vec}(\mathcal{X})
d_h(\mathcal{X},\boldsymbol{Z})
\right] \\
&\quad-
\mathbb{E}\!\left[
v(\boldsymbol{Z},\mathcal{X})
\operatorname{vec}(\mathcal{X})
\widetilde{\boldsymbol{Z}}^\top
\right]
\left\{
\mathbb{E}\!\left[
v(\boldsymbol{Z},\mathcal{X})
\widetilde{\boldsymbol{Z}}
\widetilde{\boldsymbol{Z}}^\top
\right]
\right\}^{-1}
\mathbb{E}\!\left[
\widetilde{\boldsymbol{Z}}
d_h(\mathcal{X},\boldsymbol{Z})
\right] \\
&=
\mathbb{E}\!\left[
\operatorname{vec}(\mathcal{X})
d_h(\mathcal{X},\boldsymbol{Z})
\right]
-A^\top
\mathbb{E}\!\left[
\widetilde{\boldsymbol{Z}}
d_h(\mathcal{X},\boldsymbol{Z})
\right] \\
&=
\mathbb{E}\!\left[
\big\{
\operatorname{vec}(\mathcal{X})
-A^\top\widetilde{\boldsymbol{Z}}
\big\}
d_h(\mathcal{X},\boldsymbol{Z})
\right]\\
&=\boldsymbol{r}_h.
\end{aligned}
\]
The two terms involving \(\beta_0\) can be combined as
\[
\begin{aligned}
&\mathbb{E}\!\left[
v(\boldsymbol{Z},\mathcal{X})
\operatorname{vec}(\mathcal{X})
\widetilde{\boldsymbol{Z}}^\top
\right]A\beta_0 -
\mathbb{E}\!\left[
v(\boldsymbol{Z},\mathcal{X})
\operatorname{vec}(\mathcal{X})
\operatorname{vec}(\mathcal{X})^\top
\right]\beta_0 \\
&=-
\Bigg\{
\mathbb{E}\!\left[
v(\boldsymbol{Z},\mathcal{X})
\operatorname{vec}(\mathcal{X})
\operatorname{vec}(\mathcal{X})^\top
\right] \\
&\qquad-
\mathbb{E}\!\left[
v(\boldsymbol{Z},\mathcal{X})
\operatorname{vec}(\mathcal{X})
\widetilde{\boldsymbol{Z}}^\top
\right]
\left\{
\mathbb{E}\!\left[
v(\boldsymbol{Z},\mathcal{X})
\widetilde{\boldsymbol{Z}}
\widetilde{\boldsymbol{Z}}^\top
\right]
\right\}^{-1} \\
&\qquad\quad\times
\mathbb{E}\!\left[
v(\boldsymbol{Z},\mathcal{X})
\widetilde{\boldsymbol{Z}}
\operatorname{vec}(\mathcal{X})^\top
\right]
\Bigg\}\beta_0,
\end{aligned}
\]
where we used the transpose of the definition of \(A^\top\), namely,
\[
A=
\left\{
\mathbb{E}\!\left[
v(\boldsymbol{Z},\mathcal{X})
\widetilde{\boldsymbol{Z}}
\widetilde{\boldsymbol{Z}}^\top
\right]
\right\}^{-1}
\mathbb{E}\!\left[
v(\boldsymbol{Z},\mathcal{X})
\widetilde{\boldsymbol{Z}}
\operatorname{vec}(\mathcal{X})^\top
\right].
\]
Here the matrix inside the inverse is symmetric. It remains to show that
the matrix inside braces is \(V\). Directly expanding the definition of
\(V\) gives
\[
\begin{aligned}
V
&=
\mathbb{E}\!\left[
v(\boldsymbol{Z},\mathcal{X})
\big\{
\operatorname{vec}(\mathcal{X})
-A^\top\widetilde{\boldsymbol{Z}}
\big\}
\big\{
\operatorname{vec}(\mathcal{X})
-A^\top\widetilde{\boldsymbol{Z}}
\big\}^\top
\right] \\
&=
\mathbb{E}\!\left[
v(\boldsymbol{Z},\mathcal{X})
\operatorname{vec}(\mathcal{X})
\operatorname{vec}(\mathcal{X})^\top
\right] \\
&\quad-
\mathbb{E}\!\left[
v(\boldsymbol{Z},\mathcal{X})
\operatorname{vec}(\mathcal{X})
\widetilde{\boldsymbol{Z}}^\top
\right]A .
\end{aligned}
\]
Therefore,
\[
\operatorname{vec}\!\left\{
\operatorname{pmtcov}
(Y,\mathcal{X}\mid\boldsymbol{Z})
\right\}
=
-V\operatorname{vec}(\mathcal{B}_0)
+\boldsymbol{r}_h.
\]

If
$
\mathbb{E}\!\left[
\mathrm{R}(\varepsilon)
\mid
\mathcal{X},\boldsymbol{Z}
\right]
=0
\quad\text{almost surely},
$
then
\(\boldsymbol{r}_h=\boldsymbol{0}\).
If \(V\) is also nonsingular, the exact identity implies
\[
\operatorname{pmtcov}
(Y,\mathcal{X}\mid\boldsymbol{Z})
=0
\quad\Longleftrightarrow\quad
\mathcal{B}_0=0.
\]

It remains to establish the order of the smoothing term for asymmetric
conditional errors. Integration by parts and the substitution \(e=ht\)
give
\[
\begin{aligned}
d_h(\mathcal{X},\boldsymbol{Z})
&=
\int_{\mathbb{R}}
\phi_h'(e)
g(e\mid\mathcal{X},\boldsymbol{Z})
\,\mathrm{d}e \\
&=
-\int_{\mathbb{R}}
\phi(t)
g^{(1)}(ht\mid\mathcal{X},\boldsymbol{Z})
\,\mathrm{d}t.
\end{aligned}
\]
Using
\[
g^{(1)}(0\mid\mathcal{X},\boldsymbol{Z})=0,
\qquad
\int_{\mathbb{R}}t\phi(t)\,\mathrm{d}t=0,
\qquad
\int_{\mathbb{R}}t^2\phi(t)\,\mathrm{d}t=1,
\]
a Taylor expansion around zero yields
\[
d_h(\mathcal{X},\boldsymbol{Z})
=
-\frac{h^2}{2}
g^{(3)}(0\mid\mathcal{X},\boldsymbol{Z})
+o(h^2).
\]
Under the stated dominated-expansion and integrability conditions,
this expansion may be integrated after multiplication by
\(\operatorname{vec}(\mathcal{X})
-A^\top\widetilde{\boldsymbol{Z}}\).
Consequently,
\[
\boldsymbol{r}_h=O(h^2),
\]
which proves the final assertion.

\end{proof}

\subsection{Proof of Lemma 2}

\begin{proof}
Let
$
\delta_n
=
\widehat{\widetilde{\boldsymbol{\gamma}}}_{Y\mid Z}
-
\widetilde{\boldsymbol{\gamma}}_{Y\mid Z}^{0}.
$
By Assumption~\ref{assum:A4} and $h_Z\asymp n^{-1/7}$,
$$
\|\delta_n\|_2=O_p(a_n),
\qquad
a_n=h_Z^2+(nh_Z^3)^{-1/2}\asymp n^{-2/7}.
$$
Moreover,
$
\widehat e_i
=
e_i^0-\widetilde{\boldsymbol Z}_i^\top\delta_n.
$

For notational simplicity, set
$
P=\prod_{k=1}^m p_k
,
d=p_z+1
,
A_i=\operatorname{vec}(\mathcal X_i)\in\mathbb R^P
,
z_i=\widetilde{\boldsymbol Z}_i\in\mathbb R^d,
$
and write $A=\operatorname{vec}(\mathcal X)$ and
$z=\widetilde{\boldsymbol Z}$ for their population counterparts.
Throughout the proof, $C$ denotes a finite deterministic
constant independent of $n$, whose value may change from
line to line.

Since vectorization preserves the Frobenius norm, it suffices
to bound
\[
\operatorname{vec}\!\left\{
\widehat{\mathcal C}_{\mathrm M}
-
\mathcal C_{\mathrm M}
\right\}
=
A_{1n}+A_{2n},
\]
where
\[
A_{1n}
=
\frac1n\sum_{i=1}^n
\left[
\mathrm R(e_i^0)A_i
-
\mathbb E\{\mathrm R(e^0)A\}
\right]
\]
and
\[
A_{2n}
=
\frac1n\sum_{i=1}^n
\left[
\mathrm R(e_i^0-z_i^\top\delta_n)
-
\mathrm R(e_i^0)
\right]A_i.
\]

\medskip
\noindent
\emph{Step 1: Control of the oracle empirical term.}

Since $\mathrm R(e)=h_Z^{-2}\phi'(e/h_Z)$,
a change of variables and Assumption~\ref{assum:A12} give
\[
\begin{aligned}
\mathbb E\!\left[
\mathrm R(e^0)^2
\mid\mathcal X,\boldsymbol Z
\right]
&=
h_Z^{-3}
\int_{\mathbb R}
\{\phi'(u)\}^2
f_n(h_Zu\mid\mathcal X,\boldsymbol Z)\,du
\\
&\leq Ch_Z^{-3}.
\end{aligned}
\]
Consequently, by Assumption~\ref{assum:A6},
\[
\mathbb E\|\mathrm R(e^0)A\|_2^2
\leq
Ch_Z^{-3}\mathbb E\|A\|_2^2
\leq
Ch_Z^{-3}
.
\]
Independence of the observations therefore implies
\[
\mathbb E\|A_{1n}\|_2^2
\leq
\frac{C}{nh_Z^3}.
\]
By Markov's inequality, for every $M>0$,
\[
\begin{aligned}
\Pr\left\{
\|A_{1n}\|_2>M(nh_Z^3)^{-1/2}
\right\}
&\leq
\frac{nh_Z^3}{M^2}\,
\mathbb E\|A_{1n}\|_2^2
\\
&\leq \frac{C}{M^2}.
\end{aligned}
\]
Therefore,
\[
\|A_{1n}\|_2
=
O_p\!\left((nh_Z^3)^{-1/2}\right).
\]

\medskip
\noindent
\emph{Step 2: Expansion of the plug-in term.}

Put $s_i=z_i^\top\delta_n$.
Taylor's theorem gives
\[
\mathrm R(e_i^0-s_i)
=
\mathrm R(e_i^0)
-
\mathrm R'(e_i^0)s_i
+
\frac12\mathrm R''(e_i^0)s_i^2
+
r_{i,n},
\]
where
\[
|r_{i,n}|
\leq
\frac{|s_i|^3}{6}
\sup_{|v|\leq|s_i|}
\left|\mathrm R'''(e_i^0-v)\right|.
\]
Thus,
\[
A_{2n}
=
-B_{1n}\delta_n
+
\frac12B_{2n}[\delta_n,\delta_n]
+
\mathcal R_n,
\]
where
\[
B_{1n}
=
\frac1n\sum_{i=1}^n
\mathrm R'(e_i^0)A_i z_i^\top,
\]
\[
B_{2n}[u,v]
=
\frac1n\sum_{i=1}^n
\mathrm R''(e_i^0)A_i
(z_i^\top u)(z_i^\top v),
\qquad u,v\in\mathbb R^d,
\]
and
\[
\mathcal R_n
=
\frac1n\sum_{i=1}^n r_{i,n}A_i.
\]

\medskip
\noindent
\emph{Step 3: Control of the first-order plug-in term.}

Since $\mathrm R'=\phi_{h_Z}''$,
two integrations by parts yield
\[
\mathbb E\!\left[
\mathrm R'(e^0)
\mid\mathcal X,\boldsymbol Z
\right]
=
\int_{\mathbb R}
\phi(u)
f_n^{(2)}(h_Zu\mid\mathcal X,\boldsymbol Z)\,du.
\]
The boundary terms vanish by
Assumption~\ref{assum:A12} and the decay of the Gaussian
kernel derivatives at infinity.
Hence,
\[
\left|
\mathbb E\!\left[
\mathrm R'(e^0)
\mid\mathcal X,\boldsymbol Z
\right]
\right|
\leq C.
\]
Together with Assumption~\ref{assum:A6}, this implies
\[
\|\mathbb E B_{1n}\|_F
\leq
C\mathbb E(\|A\|_2\|z\|_2)
=
O(1).
\]

Furthermore,
\[
\begin{aligned}
\mathbb E\!\left[
\{\mathrm R'(e^0)\}^2
\mid\mathcal X,\boldsymbol Z
\right]
&=
h_Z^{-5}
\int_{\mathbb R}
\{\phi''(u)\}^2
f_n(h_Zu\mid\mathcal X,\boldsymbol Z)\,du
\\
&\leq Ch_Z^{-5}.
\end{aligned}
\]
Since
$\mathbb E(\|A\|_2^2\|z\|_2^2)<\infty$,
we obtain
\[
\mathbb E
\|B_{1n}-\mathbb E B_{1n}\|_F^2
\leq
\frac{C}{nh_Z^5}.
\]
Therefore,
\[
\|B_{1n}\|_F
=
O_p\!\left\{1+(nh_Z^5)^{-1/2}\right\},
\]
and
\[
\|B_{1n}\delta_n\|_2
=
O_p\!\left[
a_n\{1+(nh_Z^5)^{-1/2}\}
\right].
\]
Because $nh_Z^5\asymp n^{2/7}\to\infty$,
\[
\|B_{1n}\delta_n\|_2
=
O_p(n^{-2/7}).
\]

\medskip
\noindent
\emph{Step 4: Control of the second-order plug-in term.}

Represent the bilinear map $B_{2n}$ by the finite-dimensional
array $H_n=(H_{n,j\ell t})$, where
\[
H_{n,j\ell t}
=
\frac1n\sum_{i=1}^n
\mathrm R''(e_i^0)A_{ij}z_{i\ell}z_{it},
\]
for $1\leq j\leq P$ and $1\leq\ell,t\leq d$.
Define
\[
\|H_n\|_F^2
=
\sum_{j=1}^P
\sum_{\ell=1}^d
\sum_{t=1}^d
H_{n,j\ell t}^2.
\]
The Cauchy--Schwarz inequality gives
\[
\|B_{2n}[u,v]\|_2
\leq
\|H_n\|_F\|u\|_2\|v\|_2.
\]

Since $\mathrm R''=\phi_{h_Z}'''$,
three integrations by parts yield
\[
\mathbb E\!\left[
\mathrm R''(e^0)
\mid\mathcal X,\boldsymbol Z
\right]
=
-
\int_{\mathbb R}
\phi(u)
f_n^{(3)}(h_Zu\mid\mathcal X,\boldsymbol Z)\,du.
\]
By Assumptions~\ref{assum:A6} and~\ref{assum:A12},
\[
\|\mathbb E H_n\|_F
\leq
C\mathbb E(\|A\|_2\|z\|_2^2)
=
O(1).
\]
Moreover,
\[
\begin{aligned}
\mathbb E\!\left[
\{\mathrm R''(e^0)\}^2
\mid\mathcal X,\boldsymbol Z
\right]
&=
h_Z^{-7}
\int_{\mathbb R}
\{\phi'''(u)\}^2
f_n(h_Zu\mid\mathcal X,\boldsymbol Z)\,du
\\
&\leq Ch_Z^{-7}.
\end{aligned}
\]
Using
$\mathbb E(\|A\|_2^2\|z\|_2^4)<\infty$,
we obtain
\[
\mathbb E\|H_n-\mathbb E H_n\|_F^2
\leq
\frac{C}{nh_Z^7}.
\]
Hence,
\[
\|H_n\|_F
=
O_p\!\left\{1+(nh_Z^7)^{-1/2}\right\},
\]
and
\[
\|B_{2n}[\delta_n,\delta_n]\|_2
=
O_p\!\left[
a_n^2\{1+(nh_Z^7)^{-1/2}\}
\right].
\]
Since $nh_Z^7\asymp1$, it follows that
\[
\|B_{2n}[\delta_n,\delta_n]\|_2
=
O_p(n^{-4/7}).
\]

\medskip
\noindent
\emph{Step 5: Control of the Taylor remainder.}

For a fixed $K>0$, define
\[
\mathcal E_{n,K}
=
\{\|\delta_n\|_2\leq Ka_n\}.
\]
Since $\|\delta_n\|_2=O_p(a_n)$, the probability of
$\mathcal E_{n,K}^{\,c}$ can be made arbitrarily small,
for all sufficiently large $n$, by choosing $K$
sufficiently large.

On $\mathcal E_{n,K}$,
\[
|s_i|\leq Ka_n\|z_i\|_2.
\]
Consequently,
\[
\begin{aligned}
\|\mathcal R_n\|_2
&\leq
C_Ka_n^3
\frac1n\sum_{i=1}^n
\|A_i\|_2\|z_i\|_2^3
\\
&\qquad\qquad\times
\sup_{|v|\leq Ka_n\|z_i\|_2}
\left|\mathrm R'''(e_i^0-v)\right|,
\end{aligned}
\]
where $C_K$ is a finite constant depending on $K$.

Let $F=\phi^{(4)}$.
Both $F$ and $F'=\phi^{(5)}$ are integrable.
For every $L\geq0$, the fundamental theorem of calculus gives
\[
\sup_{|v|\leq L}|F(u-v)|
\leq
|F(u)|
+
\int_{-L}^{L}|F'(u-t)|\,dt.
\]
Integrating over $u$ and applying Tonelli's theorem yields
\[
\int_{\mathbb R}
\sup_{|v|\leq L}|F(u-v)|\,du
\leq
\|F\|_{L_1}
+
2L\|F'\|_{L_1}.
\]

Since
\[
\mathrm R'''(e)
=
h_Z^{-5}\phi^{(4)}(e/h_Z),
\]
Assumption~\ref{assum:A12} implies
\[
\begin{aligned}
&\mathbb E\!\left[
\sup_{|v|\leq Ka_n\|z\|_2}
|\mathrm R'''(e^0-v)|
\;\middle|\;
\mathcal X,\boldsymbol Z
\right]
\\
&\quad\leq
Mh_Z^{-5}
\int_{\mathbb R}
\sup_{|v|\leq Ka_n\|z\|_2}
\left|
\phi^{(4)}\!\left(\frac{e-v}{h_Z}\right)
\right|\,de
\\
&\quad\leq
Ch_Z^{-4}
\left(
1+\frac{Ka_n\|z\|_2}{h_Z}
\right).
\end{aligned}
\]

By Assumption~\ref{assum:A6} and the Cauchy--Schwarz
inequality,
\[
\begin{aligned}
\mathbb E(\|A\|_2\|z\|_2^3)
&\leq
\left\{
\mathbb E(\|A\|_2^2\|z\|_2^4)
\right\}^{1/2}
\left\{
\mathbb E\|z\|_2^2
\right\}^{1/2}
\\
&<\infty,
\end{aligned}
\]
and
\[
\begin{aligned}
\mathbb E(\|A\|_2\|z\|_2^4)
&\leq
\left\{
\mathbb E(\|A\|_2^2\|z\|_2^4)
\right\}^{1/2}
\left\{
\mathbb E\|z\|_2^4
\right\}^{1/2}
\\
&<\infty.
\end{aligned}
\]
Therefore, for each fixed $K$,
\[
\begin{aligned}
&\mathbb E\!\left[
\|A\|_2\|z\|_2^3
\sup_{|v|\leq Ka_n\|z\|_2}
|\mathrm R'''(e^0-v)|
\right]
\\
&\qquad\leq
C_K\left(h_Z^{-4}+a_nh_Z^{-5}\right).
\end{aligned}
\]
By Markov's inequality, the corresponding nonnegative
empirical average is
\[
O_p\!\left(h_Z^{-4}+a_nh_Z^{-5}\right).
\]
Combining this bound with the preceding localization argument
gives
\[
\|\mathcal R_n\|_2
=
O_p\!\left(
a_n^3h_Z^{-4}+a_n^4h_Z^{-5}
\right).
\]
Since $a_n\asymp n^{-2/7}$ and $h_Z\asymp n^{-1/7}$,
\[
a_n^3h_Z^{-4}\asymp n^{-2/7},
\qquad
a_n^4h_Z^{-5}\asymp n^{-3/7}.
\]
Thus,
\[
\|\mathcal R_n\|_2=O_p(n^{-2/7}).
\]

\medskip
\noindent
\emph{Step 6: Conclusion.}

Combining the preceding bounds, we obtain
\[
\begin{aligned}
&
\left\|
\widehat{\mathcal C}_{\mathrm M}
-
\mathcal C_{\mathrm M}
\right\|_F
\\
&\quad=
O_p\Big[
(nh_Z^3)^{-1/2}
+
a_n\{1+(nh_Z^5)^{-1/2}\}
\\
&\qquad\qquad
+
a_n^2\{1+(nh_Z^7)^{-1/2}\}
+
a_n^3h_Z^{-4}
+
a_n^4h_Z^{-5}
\Big]
\\
&\quad=
O_p(n^{-2/7}).
\end{aligned}
\]
Because $(nh_Z^3)^{-1/2}\asymp n^{-2/7}$, this is
equivalently
\[
\left\|
\widehat{\mathcal C}_{\mathrm M}
-
\mathcal C_{\mathrm M}
\right\|_F
=
O_p\!\left((nh_Z^3)^{-1/2}\right).
\]
This completes the proof.
\end{proof}

\subsection{Proof of Lemma 3}
\begin{proof}
By Lemma~\ref{lem:2},
\begin{equation}
\label{eq:1.45}
\sqrt{nh_Z^3}
\left\|
\widehat{\mathcal{C}}_{\mathrm{M}(k)}
-
\mathcal{C}_{\mathrm{M}(k)}
\right\|_{\mathrm{op}}
=
O_p(1).
\end{equation}

By Lemma S6 of \citet{Sunetal.2025pqtr},
the empirical mode-$k$ covariance matrix satisfies
\begin{equation}
\label{eq:1.22}
\sqrt n
\left\|
\widehat{\boldsymbol{\Sigma}}_k
-
c_k\boldsymbol{\Sigma}_k
\right\|_{\mathrm{op}}
=
O_p(1).
\end{equation}

Since the inverse square-root map is continuously
differentiable on the cone of positive definite matrices,
Assumption~\ref{assum:A7} and the finite-dimensional
delta method yield
\[
\sqrt n
\left\|
\widehat{\boldsymbol{\Sigma}}_\ell^{-1/2}
-
c_\ell^{-1/2}\boldsymbol{\Sigma}_\ell^{-1/2}
\right\|_{\mathrm{op}}
=
O_p(1).
\]

Using the multiplicative property of the operator norm
for Kronecker products, it follows that
\begin{equation}
\label{eq:1.47}
\sqrt n
\left\|
\widehat{\boldsymbol{\Omega}}_{-k}
-
c_{0k}\boldsymbol{\Omega}_{-k}
\right\|_{\mathrm{op}}
=
O_p(1).
\end{equation}
In particular,
\[
\|\widehat{\boldsymbol{\Omega}}_{-k}\|_{\mathrm{op}}
=
O_p(1).
\]

Moreover, integration by parts and
Assumption~\ref{assum:A12} give
\[
\mathbb E\!\left[
\mathrm R(e^0)\mid\mathcal X,\boldsymbol Z
\right]
=
-\int_{\mathbb R}
\phi(u)
f_n^{(1)}(h_Zu\mid\mathcal X,\boldsymbol Z)\,du,
\]
whose absolute value is uniformly bounded.
Consequently, by Assumption~\ref{assum:A6},
\[
\|\mathcal C_{\mathrm{M}(k)}\|_{\mathrm{op}}
\leq
\|\mathcal C_{\mathrm M}\|_F
\leq
C\mathbb E\|\mathcal X\|_F
=
O(1).
\]

By definition,
\[
\widehat{\mathcal C}_{0k}
=
\widehat{\mathcal C}_{\mathrm{M}(k)}
\widehat{\boldsymbol{\Omega}}_{-k}.
\]
Hence,
\[
\begin{aligned}
\widehat{\mathcal C}_{0k}
-
c_{0k}\mathcal C_{0k}
&=
\left(
\widehat{\mathcal C}_{\mathrm{M}(k)}
-
\mathcal C_{\mathrm{M}(k)}
\right)
\widehat{\boldsymbol{\Omega}}_{-k}
\\
&\quad+
\mathcal C_{\mathrm{M}(k)}
\left(
\widehat{\boldsymbol{\Omega}}_{-k}
-
c_{0k}\boldsymbol{\Omega}_{-k}
\right).
\end{aligned}
\]

Combining \eqref{eq:1.45} and \eqref{eq:1.47}
with the preceding bounds and the submultiplicativity
of the operator norm yields
\[
\begin{aligned}
\left\|
\widehat{\mathcal C}_{0k}
-
c_{0k}\mathcal C_{0k}
\right\|_{\mathrm{op}}
&=
O_p\!\left((nh_Z^3)^{-1/2}\right)
+
O_p(n^{-1/2})
\\
&=
O_p\!\left((nh_Z^3)^{-1/2}\right),
\end{aligned}
\]
where the last equality follows from
$n^{-1/2}=o((nh_Z^3)^{-1/2})$.
This completes the proof.
\end{proof}

\subsection{Proof of Lemma 4}
\begin{proof}
Fix a mode $k$.
Since eigenvectors are invariant to positive scalar
multiplication, the constant $c_{0k}$ in
Lemma~\ref{lem:3} does not affect the corresponding
population directions.
We proceed by mathematical induction.

\medskip
\noindent
\emph{Step 1: Consistency of the initial positive
semidefinite matrix.}

By definition,
\[
\begin{aligned}
\widehat{\boldsymbol M}_{0k}
-
c_{0k}^2\boldsymbol M_{0k}
&=
\left(
\widehat{\mathcal C}_{0k}
-
c_{0k}\mathcal C_{0k}
\right)
\widehat{\mathcal C}_{0k}^{\top}
\\
&\quad+
c_{0k}\mathcal C_{0k}
\left(
\widehat{\mathcal C}_{0k}
-
c_{0k}\mathcal C_{0k}
\right)^{\top}.
\end{aligned}
\]
The proof of Lemma~\ref{lem:3} gives
$\|\mathcal C_{0k}\|_{\mathrm{op}}=O(1)$.
Together with Lemma~\ref{lem:3}, this implies
$\|\widehat{\mathcal C}_{0k}\|_{\mathrm{op}}=O_p(1)$.
Hence, by the submultiplicativity of the operator norm,
\[
\sqrt{nh_Z^3}
\left\|
\widehat{\boldsymbol M}_{0k}
-
c_{0k}^2\boldsymbol M_{0k}
\right\|_{\mathrm{op}}
=
O_p(1).
\]

\medskip
\noindent
\emph{Step 2: Consistency of the first direction.}

By Assumptions~\ref{assum:A8}--\ref{assum:A9}
and Corollary 1 of \citet{yu2015useful},
\[
\left\|
\widehat{\boldsymbol w}_{1k}
-
\boldsymbol w_{1k}^0
\right\|_2
\leq
\frac{
2\sqrt{2}
\left\|
\widehat{\boldsymbol M}_{0k}
-
c_{0k}^2\boldsymbol M_{0k}
\right\|_{\mathrm{op}}
}{
c_{0k}^2
\left[
\lambda_1(\boldsymbol M_{0k})
-
\lambda_2(\boldsymbol M_{0k})
\right]
}.
\]
The denominator is bounded away from zero uniformly in $n$.
Therefore,
\begin{equation}
\label{eq:1.54}
\sqrt{nh_Z^3}
\left\|
\widehat{\boldsymbol w}_{1k}
-
\boldsymbol w_{1k}^0
\right\|_2
=
O_p(1).
\end{equation}

\medskip
\noindent
\emph{Step 3: Consistency of the first deflation matrix.}

By definition, $\boldsymbol Q_{1k}^0$ is a continuously
differentiable function of
$(\boldsymbol\Sigma_k,\boldsymbol w_{1k}^0)$.
The multiplicative constant $c_k$ cancels in
\[
\boldsymbol\Sigma_k\boldsymbol W
\left(
\boldsymbol W^\top\boldsymbol\Sigma_k\boldsymbol W
\right)^{-1}
\boldsymbol W^\top,
\]
and hence does not affect the population deflation matrix.

Under Assumptions~\ref{assum:A7} and~\ref{assum:A9},
the derivatives of the defining map are uniformly bounded
in a neighborhood of the population quantities.
Combining \eqref{eq:1.22} and \eqref{eq:1.54} therefore gives
\[
\begin{aligned}
\left\|
\widehat{\boldsymbol Q}_{1k}
-
\boldsymbol Q_{1k}^0
\right\|_{\mathrm{op}}
&=
O_p\!\left(
n^{-1/2}+(nh_Z^3)^{-1/2}
\right)
\\
&=
O_p\!\left((nh_Z^3)^{-1/2}\right),
\end{aligned}
\]
where $n^{-1/2}=o((nh_Z^3)^{-1/2})$.
Thus,
\[
\sqrt{nh_Z^3}
\left\|
\widehat{\boldsymbol Q}_{1k}
-
\boldsymbol Q_{1k}^0
\right\|_{\mathrm{op}}
=
O_p(1).
\]

\medskip
\noindent
\emph{Step 4: Induction step.}

Suppose that, for some $1\leq s<r_k$,
\[
\max_{1\leq j\leq s}
\sqrt{nh_Z^3}
\left\|
\widehat{\boldsymbol w}_{jk}
-
\boldsymbol w_{jk}^0
\right\|_2
=
O_p(1)
\]
and
\[
\sqrt{nh_Z^3}
\left\|
\widehat{\boldsymbol Q}_{sk}
-
\boldsymbol Q_{sk}^0
\right\|_{\mathrm{op}}
=
O_p(1).
\]

Since
\[
\widehat{\mathcal C}_{sk}
=
\widehat{\boldsymbol Q}_{sk}
\widehat{\mathcal C}_{0k},
\]
we have
\[
\begin{aligned}
\widehat{\mathcal C}_{sk}
-
c_{0k}\mathcal C_{sk}
&=
\left(
\widehat{\boldsymbol Q}_{sk}
-
\boldsymbol Q_{sk}^0
\right)
\widehat{\mathcal C}_{0k}
\\
&\quad+
\boldsymbol Q_{sk}^0
\left(
\widehat{\mathcal C}_{0k}
-
c_{0k}\mathcal C_{0k}
\right).
\end{aligned}
\]
The unit norms of the population directions and
Assumptions~\ref{assum:A7} and~\ref{assum:A9} imply
$\|\boldsymbol Q_{sk}^0\|_{\mathrm{op}}=O(1)$.
It follows that
\[
\sqrt{nh_Z^3}
\left\|
\widehat{\mathcal C}_{sk}
-
c_{0k}\mathcal C_{sk}
\right\|_{\mathrm{op}}
=
O_p(1).
\]
Moreover,
$\|\mathcal C_{sk}\|_{\mathrm{op}}=O(1)$ and
$\|\widehat{\mathcal C}_{sk}\|_{\mathrm{op}}=O_p(1)$.
Applying the same product expansion as in Step 1 gives
\[
\sqrt{nh_Z^3}
\left\|
\widehat{\boldsymbol M}_{sk}
-
c_{0k}^2\boldsymbol M_{sk}
\right\|_{\mathrm{op}}
=
O_p(1).
\]

Another application of Corollary 1 of
\citet{yu2015useful}, together with
Assumptions~\ref{assum:A8}--\ref{assum:A9}, yields
\begin{equation}
\label{eq:1.62}
\sqrt{nh_Z^3}
\left\|
\widehat{\boldsymbol w}_{s+1,k}
-
\boldsymbol w_{s+1,k}^0
\right\|_2
=
O_p(1).
\end{equation}

Applying the same smoothness argument as in Step 3
to the defining map of $\boldsymbol Q_{s+1,k}^0$,
we obtain
\[
\sqrt{nh_Z^3}
\left\|
\widehat{\boldsymbol Q}_{s+1,k}
-
\boldsymbol Q_{s+1,k}^0
\right\|_{\mathrm{op}}
=
O_p(1).
\]
Together with \eqref{eq:1.54}, induction therefore establishes
\[
\sqrt{nh_Z^3}
\left\|
\widehat{\boldsymbol w}_{sk}
-
\boldsymbol w_{sk}^0
\right\|_2
=
O_p(1),
\qquad s=1,\ldots,r_k.
\]

Finally, since $r_k$ is fixed,
\[
\begin{aligned}
\sqrt{nh_Z^3}
\left\|
\widehat{\boldsymbol W}_k
-
\boldsymbol W_{0k}
\right\|_F
&=
\sqrt{nh_Z^3}
\left[
\sum_{s=1}^{r_k}
\left\|
\widehat{\boldsymbol w}_{sk}
-
\boldsymbol w_{sk}^0
\right\|_2^2
\right]^{1/2}
\\
&=
O_p(1).
\end{aligned}
\]
This completes the proof.
\end{proof}

\subsection{Proof of Lemma 5}

\begin{proof}
Write $\tau_n=n^{-2/7}$ and let
\[
A_i=\operatorname{vec}(\mathcal X_i),
\qquad
P=\prod_{k=1}^m p_k.
\]
As in the original decomposition, define
\[
\Delta_i
=
\widehat{\boldsymbol U}_i-\boldsymbol U_i^0,
\]
\[
r_i^0(\boldsymbol\eta)
=
Y_i-(\boldsymbol U_i^0)^\top\boldsymbol\eta,
\qquad
\widehat r_i(\boldsymbol\eta)
=
r_i^0(\boldsymbol\eta)-\Delta_i^\top\boldsymbol\eta.
\]

\medskip
\noindent
\emph{Step 1: Control of the estimated design.}

Define
\[
D_n
=
\begin{pmatrix}
\boldsymbol 0_{(p_z+1)\times P}
\\[2mm]
\widehat{\boldsymbol W}_m^\top
\otimes\cdots\otimes
\widehat{\boldsymbol W}_1^\top
-
\boldsymbol W_{0m}^\top
\otimes\cdots\otimes
\boldsymbol W_{01}^\top
\end{pmatrix}.
\]
Then
\[
\Delta_i=D_nA_i.
\]
By Lemma~\ref{lem:4} and expansion of the finite
Kronecker product,
\[
\|D_n\|_{\mathrm{op}}=O_p(\tau_n).
\]
Consequently,
\[
\|\Delta_i\|_2
\leq
\|D_n\|_{\mathrm{op}}\|\mathcal X_i\|_F.
\]

\medskip
\noindent
\emph{Step 2: Local uniform bounds for kernel derivatives.}

Introduce the auxiliary vector
\[
\boldsymbol V_i
=
\begin{pmatrix}
\widetilde{\boldsymbol Z}_i\\
A_i
\end{pmatrix}.
\]
Its dimension is fixed, and Assumption~\ref{assum:A13}
implies
\[
\mathbb E\|\boldsymbol V_i\|_2^6<\infty.
\]
For a vector $v$ of the same dimension, define
\[
\boldsymbol H_n^*(v)
=
\frac1n\sum_{i=1}^n
\phi_h''(\varepsilon_i-\boldsymbol V_i^\top v)
\boldsymbol V_i\boldsymbol V_i^\top
\]
and
\[
\boldsymbol S_n^*(v)
=
\frac1n\sum_{i=1}^n
\phi_h'(\varepsilon_i-\boldsymbol V_i^\top v)
\boldsymbol V_i.
\]

Conditional on $(\mathcal X,\boldsymbol Z)$, the residual
$e^0$ differs from $\varepsilon$ only by a deterministic
location shift.
Thus, Assumption~\ref{assum:A12} also provides uniform
bounds on the conditional density of $\varepsilon$
and its first three derivatives.

Two integrations by parts and the kernel second-moment
calculation give
\[
\|\mathbb E\boldsymbol H_n^*(0)\|_{\mathrm{op}}=O(1)
\]
and
\[
\mathbb E
\left\|
\boldsymbol H_n^*(0)
-
\mathbb E\boldsymbol H_n^*(0)
\right\|_F^2
=
O\!\left((nh^5)^{-1}\right).
\]
Therefore,
\[
\|\boldsymbol H_n^*(0)\|_{\mathrm{op}}=O_p(1).
\]

To obtain a uniform bound, let $\mathcal G_n$ denote
the coefficient array with entries
\[
(\mathcal G_n)_{abc}
=
\frac1n\sum_{i=1}^n
\phi_h'''(\varepsilon_i)
V_{ia}V_{ib}V_{ic}.
\]
Three integrations by parts show that
\[
\|\mathbb E\mathcal G_n\|_F=O(1).
\]
Moreover,
\[
\mathbb E\!\left[
\{\phi_h'''(\varepsilon)\}^2
\mid\mathcal X,\boldsymbol Z
\right]
\leq Ch^{-7}.
\]
By Assumption~\ref{assum:A13},
\[
\mathbb E
\|\mathcal G_n-\mathbb E\mathcal G_n\|_F^2
=
O\!\left((nh^7)^{-1}\right).
\]
Since $nh^7\asymp1$,
\[
\|\mathcal G_n\|_F=O_p(1).
\]

Fix $K<\infty$.
Taylor's theorem gives
\[
\begin{aligned}
&
\sup_{\|v\|_2\leq K\tau_n}
\left\|
\boldsymbol H_n^*(v)-\boldsymbol H_n^*(0)
\right\|_{\mathrm{op}}
\\
&\quad\leq
K\tau_n\|\mathcal G_n\|_F
+
\frac{K^2\tau_n^2}{2n}
\sum_{i=1}^n
\|\boldsymbol V_i\|_2^4
\sup_{|a|\leq K\tau_n\|\boldsymbol V_i\|_2}
|\phi_h^{(4)}(\varepsilon_i-a)|.
\end{aligned}
\]

For the Gaussian kernel,
\[
\int_{\mathbb R}
\sup_{|a|\leq L}
|\phi^{(4)}(u-a)|\,du
\leq
\|\phi^{(4)}\|_{L_1}
+
2L\|\phi^{(5)}\|_{L_1},
\qquad L\geq0.
\]
This follows from the fundamental theorem of calculus
and Tonelli's theorem.
Using the conditional density bound, we therefore obtain
\[
\begin{aligned}
&
\mathbb E\!\left[
\sup_{|a|\leq K\tau_n\|\boldsymbol V\|_2}
|\phi_h^{(4)}(\varepsilon-a)|
\;\middle|\;
\mathcal X,\boldsymbol Z
\right]
\\
&\qquad\leq
Ch^{-4}
\left(
1+\frac{K\tau_n\|\boldsymbol V\|_2}{h}
\right).
\end{aligned}
\]
Assumption~\ref{assum:A13} and Markov's inequality imply
\[
\begin{aligned}
&
\sup_{\|v\|_2\leq K\tau_n}
\left\|
\boldsymbol H_n^*(v)-\boldsymbol H_n^*(0)
\right\|_{\mathrm{op}}
\\
&\qquad=
O_p\!\left(
\tau_n+\tau_n^2h^{-4}+\tau_n^3h^{-5}
\right)
=
O_p(1).
\end{aligned}
\]
Hence, for every fixed $K$,
\[
\sup_{\|v\|_2\leq K\tau_n}
\|\boldsymbol H_n^*(v)\|_{\mathrm{op}}
=
O_p(1).
\]

Next, the condition
$g^{(1)}(0\mid\mathcal X,\boldsymbol Z)=0$,
the symmetry of the Gaussian kernel, and the bounded
third derivative imply
\[
\|\mathbb E\boldsymbol S_n^*(0)\|_2=O(h^2).
\]
Also,
\[
\mathbb E
\left\|
\boldsymbol S_n^*(0)
-
\mathbb E\boldsymbol S_n^*(0)
\right\|_2^2
=
O\!\left((nh^3)^{-1}\right).
\]
It follows that
\[
\|\boldsymbol S_n^*(0)\|_2
=
O_p\!\left(h^2+(nh^3)^{-1/2}\right)
=
O_p(\tau_n).
\]
The integral identity
\[
\boldsymbol S_n^*(v)
=
\boldsymbol S_n^*(0)
-
\int_0^1\boldsymbol H_n^*(tv)v\,dt
\]
then yields
\[
\sup_{\|v\|_2\leq K\tau_n}
\|\boldsymbol S_n^*(v)\|_2
=
O_p(\tau_n).
\]

\medskip
\noindent
\emph{Step 3: Control of the score difference.}

By the fundamental theorem of calculus,
\[
\begin{aligned}
\phi_h'(\widehat r_i(\boldsymbol\eta))
-
\phi_h'(r_i^0(\boldsymbol\eta))
=
-
\int_0^1
\phi_h''\!\left(
r_i^0(\boldsymbol\eta)
-
t\Delta_i^\top\boldsymbol\eta
\right)
\Delta_i^\top\boldsymbol\eta\,dt.
\end{aligned}
\]
Consequently, the score difference admits the exact
decomposition
\[
\nabla Q_{n,h}(\boldsymbol\eta)
-
\nabla Q_{n,h}^0(\boldsymbol\eta)
=
T_{1n}(\boldsymbol\eta)
+
T_{2n}(\boldsymbol\eta),
\]
where
\[
T_{1n}(\boldsymbol\eta)
=
-
D_n
\left[
\frac1n\sum_{i=1}^n
\phi_h'(r_i^0(\boldsymbol\eta))A_i
\right]
\]
and
\[
\begin{aligned}
T_{2n}(\boldsymbol\eta)
&=
\int_0^1
\left[
\frac1n\sum_{i=1}^n
\phi_h''\!\left(
r_i^0(\boldsymbol\eta)
-
t\Delta_i^\top\boldsymbol\eta
\right)
\widehat{\boldsymbol U}_i A_i^\top
\right]
D_n^\top\boldsymbol\eta\,dt.
\end{aligned}
\]

The vectors $\boldsymbol U_i^0$ and
$\widehat{\boldsymbol U}_i$ are linear transformations
of $\boldsymbol V_i$ with bounded operator norms:
the dimensions are fixed and all extracted directions
have unit Euclidean norm.
Using
$\varepsilon_i=Y_i-(\boldsymbol U_i^0)^\top
\boldsymbol\eta_0$,
we may write
\[
r_i^0(\boldsymbol\eta)
-
t\Delta_i^\top\boldsymbol\eta
=
\varepsilon_i-\boldsymbol V_i^\top v_{t,\boldsymbol\eta},
\]
where
\[
\sup_{\substack{
\|\boldsymbol\eta-\boldsymbol\eta_0\|_2\leq\delta_n\\
0\leq t\leq1
}}
\|v_{t,\boldsymbol\eta}\|_2
\leq
C\delta_n
+
\|D_n\|_{\mathrm{op}}
\sup_{\|\boldsymbol\eta-\boldsymbol\eta_0\|_2\leq\delta_n}
\|\boldsymbol\eta\|_2
=
O_p(\tau_n).
\]
Here we used Assumption~\ref{assum:A2},
$\delta_n=O(\tau_n)$, and Step 1.

Applying the local uniform bounds from Step 2 on events
where all these vectors lie in a ball of radius $K\tau_n$
gives
\[
\sup_{\|\boldsymbol\eta-\boldsymbol\eta_0\|_2\leq\delta_n}
\left\|
\frac1n\sum_{i=1}^n
\phi_h'(r_i^0(\boldsymbol\eta))A_i
\right\|_2
=
O_p(\tau_n)
\]
and
\[
\begin{aligned}
&
\sup_{\substack{
\|\boldsymbol\eta-\boldsymbol\eta_0\|_2\leq\delta_n\\
0\leq t\leq1
}}
\left\|
\frac1n\sum_{i=1}^n
\phi_h''\!\left(
r_i^0(\boldsymbol\eta)
-
t\Delta_i^\top\boldsymbol\eta
\right)
\widehat{\boldsymbol U}_i A_i^\top
\right\|_{\mathrm{op}}
\\
&\qquad=O_p(1).
\end{aligned}
\]
For the second bound, the displayed matrix is obtained
from $\boldsymbol H_n^*(v_{t,\boldsymbol\eta})$
by left and right multiplication by bounded matrices.
These uniform bounds do not require independence
between $D_n$ and the observations.

It follows that
\[
\sup_{\|\boldsymbol\eta-\boldsymbol\eta_0\|_2\leq\delta_n}
\|T_{1n}(\boldsymbol\eta)\|_2
=
O_p(\tau_n^2)
\]
and
\[
\sup_{\|\boldsymbol\eta-\boldsymbol\eta_0\|_2\leq\delta_n}
\|T_{2n}(\boldsymbol\eta)\|_2
=
O_p(\tau_n).
\]
Combining these bounds proves
\[
\sup_{\|\boldsymbol\eta-\boldsymbol\eta_0\|_2\leq\delta_n}
\left\|
\nabla Q_{n,h}(\boldsymbol\eta)
-
\nabla Q_{n,h}^0(\boldsymbol\eta)
\right\|_2
=
O_p(n^{-2/7}).
\]
This completes the proof.
\end{proof}

\subsection{Proof of Theorem 1}

\begin{proof}
Write $\tau_n=n^{-2/7}$.
Throughout the proof, all dimensions are fixed.

First, conditional integration by parts, together with
$g^{(1)}(0\mid\mathcal{X},\boldsymbol{Z})=0$ and the
symmetry of the Gaussian kernel, gives
\[
\left\|
\mathbb{E}\{
\nabla Q_{n,h}^{0}(\boldsymbol{\eta}_0)
\}
\right\|_2
=O(h^2).
\]
Moreover,
\[
\mathbb{E}\left\|
\nabla Q_{n,h}^{0}(\boldsymbol{\eta}_0)
-
\mathbb{E}\{
\nabla Q_{n,h}^{0}(\boldsymbol{\eta}_0)
\}
\right\|_2^2
=O\bigl((nh^3)^{-1}\bigr).
\]
Consequently, Lemma~\ref{lem:5} implies
\[
\left\|
\nabla Q_{n,h}(\boldsymbol{\eta}_0)
\right\|_2
=O_p(\tau_n).
\]

We next establish that, for every fixed $M<\infty$,
\begin{equation}\label{eq:thm1-local-hessian}
\sup_{\|\boldsymbol{\eta}-\boldsymbol{\eta}_0\|_2
      \leq M\tau_n}
\left\|
\nabla^2 Q_{n,h}(\boldsymbol{\eta})
-\boldsymbol{J}
\right\|_{\mathrm{op}}
=o_p(1).
\end{equation}
Let
\[
\boldsymbol{V}_i
=
\begin{pmatrix}
\widetilde{\boldsymbol{Z}}_i\\
\operatorname{vec}(\mathcal{X}_i)
\end{pmatrix},
\qquad
\boldsymbol{H}_n(\boldsymbol{v})
=
\frac1n\sum_{i=1}^n
\phi_h''(\varepsilon_i-\boldsymbol{V}_i^\top
\boldsymbol{v})
\boldsymbol{V}_i\boldsymbol{V}_i^\top .
\]
Assumption~\ref{assum:A13} gives
$\mathbb{E}\|\boldsymbol{V}_i\|_2^6<\infty$.
Since $e^0$ and $\varepsilon$ differ by a measurable
function of the covariates, Assumption~\ref{assum:A12}
also bounds the conditional density of $\varepsilon$
and its first three derivatives.

Conditional integration by parts and $nh^5\to\infty$
therefore yield
\[
\boldsymbol{H}_n(\boldsymbol{0})
-
\mathbb{E}\!\left[
g^{(2)}(0\mid\mathcal{X},\boldsymbol{Z})
\boldsymbol{V}\boldsymbol{V}^\top
\right]
=o_p(1).
\]
We claim that, for every fixed $K<\infty$,
\begin{equation}\label{eq:thm1-hessian-equicontinuity}
\sup_{\|\boldsymbol{v}\|_2\leq K\tau_n}
\left\|
\boldsymbol{H}_n(\boldsymbol{v})
-\boldsymbol{H}_n(\boldsymbol{0})
\right\|_{\mathrm{op}}
=o_p(1).
\end{equation}

To verify this claim, the coefficient array of the
linear Taylor term is
\[
\frac1n\sum_{i=1}^n
\phi_h'''(\varepsilon_i)
\boldsymbol{V}_i^{\otimes 3}.
\]
Its expectation is bounded, and its centered second
moment is $O((nh^7)^{-1})=O(1)$.
Thus, the linear term is $O_p(\tau_n)$ uniformly over
$\|\boldsymbol{v}\|_2\leq K\tau_n$.

For the remainder, first restrict to
$\|\boldsymbol{V}_i\|_2\leq B$, where $B$ is fixed.
A further Taylor expansion shows that its quadratic
coefficient is $O_p(h^{-1})$ and that the remaining
term is $O_p(\tau_n^3h^{-5})$.
Indeed, three integrations by parts bound the
expectation of the quadratic coefficient by $O(h^{-1})$,
and its centered second moment is $O((nh^9)^{-1})$.
The bound for the remaining term follows from the
integrability of the fifth and sixth Gaussian
derivatives.
Hence the contribution on
$\{\|\boldsymbol{V}_i\|_2\leq B\}$ is
\[
O_p(\tau_n^2h^{-1}+\tau_n^3h^{-5})
=o_p(1).
\]

On $\{\|\boldsymbol{V}_i\|_2>B\}$, the expected
supremum of the remainder is bounded by
\[
C_K\tau_n^2h^{-4}
\mathbb{E}\!\left[
\|\boldsymbol{V}\|_2^4
\mathbf{1}_{\{\|\boldsymbol{V}\|_2>B\}}
\right]
+
C_K\tau_n^3h^{-5}
\mathbb{E}\!\left[
\|\boldsymbol{V}\|_2^5
\mathbf{1}_{\{\|\boldsymbol{V}\|_2>B\}}
\right].
\]
Here we used the Gaussian derivative envelope bound
\[
\int_{\mathbb{R}}
\sup_{|a|\leq L}
|\phi^{(j)}(u-a)|\,du
\leq
\|\phi^{(j)}\|_{L_1}
+
2L\|\phi^{(j+1)}\|_{L_1}.
\]
Since $\tau_n^2h^{-4}=O(1)$ and
$\tau_n^3h^{-5}=o(1)$, Markov's inequality, followed
by $B\to\infty$, proves
\eqref{eq:thm1-hessian-equicontinuity}.

Both $\boldsymbol{U}_i^0$ and
$\widehat{\boldsymbol{U}}_i$ are linear transformations
of $\boldsymbol{V}_i$ with bounded coefficient matrices.
By Lemma~\ref{lem:4}, these coefficient matrices differ
by $O_p(\tau_n)$.
Furthermore, uniformly over
$\|\boldsymbol{\eta}-\boldsymbol{\eta}_0\|_2
\leq M\tau_n$, we can write
\[
Y_i-\widehat{\boldsymbol{U}}_i^\top\boldsymbol{\eta}
=
\varepsilon_i-\boldsymbol{V}_i^\top
\boldsymbol{v}_{\boldsymbol{\eta}},
\qquad
\sup_{\|\boldsymbol{\eta}-\boldsymbol{\eta}_0\|_2
      \leq M\tau_n}
\|\boldsymbol{v}_{\boldsymbol{\eta}}\|_2
=O_p(\tau_n).
\]
Applying \eqref{eq:thm1-hessian-equicontinuity} on
deterministic balls and then localizing the random
vectors $\boldsymbol{v}_{\boldsymbol{\eta}}$
establishes \eqref{eq:thm1-local-hessian}.

Let $c>0$ satisfy
$\lambda_{\min}(-\boldsymbol{J})\geq c$.
For $\|\boldsymbol{u}\|_2=M\tau_n$, Taylor's formula
with an integral remainder gives
\[
\begin{aligned}
&Q_{n,h}(\boldsymbol{\eta}_0+\boldsymbol{u})
-Q_{n,h}(\boldsymbol{\eta}_0)\\
&\quad=
\boldsymbol{u}^\top
\nabla Q_{n,h}(\boldsymbol{\eta}_0)
+
\int_0^1(1-t)\,
\boldsymbol{u}^\top
\nabla^2Q_{n,h}(\boldsymbol{\eta}_0+t\boldsymbol{u})
\boldsymbol{u}\,dt .
\end{aligned}
\]
By \eqref{eq:thm1-local-hessian}, the right-hand side
is at most
\[
M\tau_n
\left\|\nabla Q_{n,h}(\boldsymbol{\eta}_0)\right\|_2
-\frac{c}{4}M^2\tau_n^2
\]
with probability tending to one.
Since the score is $O_p(\tau_n)$, choosing $M$
sufficiently large makes the boundary objective
smaller than its value at $\boldsymbol{\eta}_0$
with arbitrarily high probability.
Continuity and the interior condition therefore yield
a local maximizer satisfying
\[
\left\|
\widehat{\boldsymbol{\eta}}-\boldsymbol{\eta}_0
\right\|_2
=O_p(\tau_n).
\]

Finally, the Tucker multilinear mapping is locally
Lipschitz on bounded sets. Hence,
\[
\left\|
\widehat{\mathcal{B}}-\mathcal{B}_0
\right\|_F
\leq
C\left\|
\widehat{\mathcal{D}}-\mathcal{D}_0
\right\|_F
+
C\sum_{k=1}^m
\left\|
\widehat{\boldsymbol{W}}_k-\boldsymbol{W}_{0k}
\right\|_F
=O_p(\tau_n).
\]
Combining this bound with the corresponding bounds
for $\widehat{\alpha}$ and
$\widehat{\boldsymbol{\gamma}}$ proves
\[
d(\widehat{\boldsymbol{\theta}},\boldsymbol{\theta}_0)
=O_p(n^{-2/7}).
\]
\end{proof}

\subsection{Proof of Theorem 2}

\begin{proof}
Fix a mode $k$ and define, as before,
\[
\boldsymbol{C}_{0k}^*
=c_{0k}\boldsymbol{C}_{0k},
\qquad
\boldsymbol{M}_{0k}^*
=\boldsymbol{C}_{0k}^*
(\boldsymbol{C}_{0k}^*)^\top
=c_{0k}^2\boldsymbol{M}_{0k},
\]
and
\[
\boldsymbol{E}_{nk}
=\widehat{\boldsymbol{C}}_{0k}
-\boldsymbol{C}_{0k}^*.
\]
By Lemma~\ref{lem:3},
\[
\|\boldsymbol{E}_{nk}\|_{\mathrm{op}}
=O_p(n^{-2/7}).
\]
The boundedness established in that lemma gives
$\|\boldsymbol{C}_{0k}^*\|_{\mathrm{op}}=O(1)$.
Consequently,
\begin{equation}\label{eq:t2.1}
\begin{aligned}
\left\|
\widehat{\boldsymbol{M}}_{0k}
-\boldsymbol{M}_{0k}^*
\right\|_{\mathrm{op}}
&\leq
2\|\boldsymbol{C}_{0k}^*\|_{\mathrm{op}}
\|\boldsymbol{E}_{nk}\|_{\mathrm{op}}
+\|\boldsymbol{E}_{nk}\|_{\mathrm{op}}^2\\
&=O_p(n^{-2/7}).
\end{aligned}
\end{equation}

Let $\lambda_{kj}^*$ denote the ordered eigenvalues
of $\boldsymbol{M}_{0k}^*$.
Weyl's inequality yields
\begin{equation}\label{eq:t2.2}
\widehat{\lambda}_{kj}
=\lambda_{kj}^*+O_p(n^{-2/7}),
\qquad j=1,\ldots,r_k.
\end{equation}
Since
$\lambda_{kj}^*=c_{0k}^2\lambda_j(\boldsymbol{M}_{0k})$,
the signal eigenvalues are uniformly bounded above
and away from zero.

For the remaining eigenvalues, the rank condition
and the singular-value perturbation inequality imply
\[
s_j(\widehat{\boldsymbol{C}}_{0k})
\leq\|\boldsymbol{E}_{nk}\|_{\mathrm{op}},
\qquad j>r_k,
\]
where singular values are padded with zeros when
necessary. Therefore,
\begin{equation}\label{eq:t2.3}
0\leq\widehat{\lambda}_{kj}
=s_j^2(\widehat{\boldsymbol{C}}_{0k})
\leq\|\boldsymbol{E}_{nk}\|_{\mathrm{op}}^2
=O_p(n^{-4/7}),
\qquad j>r_k.
\end{equation}

For $l<r_k$, both adjacent eigenvalues are signal
eigenvalues. Thus,
\begin{equation}\label{eq:t2.4}
\mathrm{RER}_{kl}
=
\frac{\lambda_{kl}^*}{\lambda_{k,l+1}^*}
+o_p(1)
=O_p(1).
\end{equation}
For $l>r_k$, \eqref{eq:t2.3} and
$n^{-4/7}=o(\rho_n)$ give
\begin{equation}\label{eq:t2.5}
\mathrm{RER}_{kl}
=
\frac{\rho_n\{1+o_p(1)\}}
     {\rho_n\{1+o_p(1)\}}
=1+o_p(1).
\end{equation}
At the true dimension,
\[
\rho_n\mathrm{RER}_{k,r_k}
=
\frac{\widehat{\lambda}_{k,r_k}+\rho_n}
     {1+\widehat{\lambda}_{k,r_k+1}/\rho_n}
=
\lambda_{k,r_k}^*+o_p(1).
\]
Since $\inf_n\lambda_{k,r_k}^*>0$ and
$\rho_n\to0$, it follows that
\begin{equation}\label{eq:t2.6}
\mathrm{RER}_{k,r_k}\overset{p}{\longrightarrow}\infty.
\end{equation}

The candidate set is finite. Equations
\eqref{eq:t2.4}--\eqref{eq:t2.6} therefore imply
\[
\operatorname{Pr}(\hat d_k=r_k)\to1.
\]
Finally, since $m$ is fixed,
\[
\operatorname{Pr}\left(
\bigcup_{k=1}^m\{\hat d_k\neq r_k\}
\right)
\leq
\sum_{k=1}^m
\operatorname{Pr}(\hat d_k\neq r_k)
\to0,
\]
which proves the joint selection consistency.
\end{proof}

\end{document}